\pdfoutput=1
\newif\ifreviewversion
\reviewversionfalse
\newif\ifaclstyle
\IfFileExists{acl.sty}{\aclstyletrue}{\aclstylefalse}

\ifaclstyle
  \documentclass[11pt]{article}
  \ifreviewversion\usepackage[review]{acl}\else\usepackage[preprint]{acl}\fi
  \usepackage{times}
  \usepackage{latexsym}
\else
  \documentclass[11pt,twocolumn]{article}
  \usepackage[a4paper,margin=2.5cm]{geometry}
  \usepackage{natbib}
  \usepackage{times}
\fi

\usepackage[T1]{fontenc}
\usepackage[utf8]{inputenc}
\usepackage{microtype}
\ifaclstyle\usepackage{inconsolata}\fi
\usepackage{amsmath}
\usepackage{booktabs}
\usepackage{tabularx}
\usepackage{enumitem}
\usepackage{xcolor}
\usepackage{graphicx}
\usepackage{float}
\usepackage{listings}
\usepackage[most]{tcolorbox}

\newtcolorbox{claimboxenv}[1]{%
  enhanced, breakable, colback=black!3, colframe=black!45,
  boxrule=0.4pt, arc=1pt, left=4pt, right=4pt, top=3pt, bottom=3pt,
  fontupper=\small,
  title={#1}, fonttitle=\scriptsize\bfseries,
  coltitle=black, colbacktitle=black!12}
\newtcolorbox{claimboxplain}{%
  enhanced, breakable, colback=black!3, colframe=black!45,
  boxrule=0.4pt, arc=1pt, left=4pt, right=4pt, top=3pt, bottom=3pt,
  fontupper=\small}
\newcommand{\ClaimBox}[2]{%
  \if\relax\detokenize{#1}\relax
    \begin{claimboxplain}#2\end{claimboxplain}%
  \else
    \begin{claimboxenv}{#1}#2\end{claimboxenv}%
  \fi}

\newcommand{\field}[1]{\texttt{#1}}
\ifreviewversion\newcommand{\system}{AnonAgent}\else\newcommand{\system}{ClaimSpy}\fi
\ifreviewversion\fi %

\newcommand{\ci}[2]{{\scriptsize[#1, #2]}}

\title{Same Claim, Different Interpretations: How LLM Agents Resolve Ambiguity in Scientific Claims}

\author{Delip Rao\thanks{Corresponding author.} \quad Ahmed Muharram \quad Chris Callison-Burch \\
  University of Pennsylvania \\
  \texttt{delip@upenn.edu}, \texttt{muharram@seas.upenn.edu}, \texttt{ccb@upenn.edu}}

\begin{document}
\maketitle

\begin{abstract}
Co-scientists and agentic scientific claim verifiers are increasingly used
to assess claims across the sciences. Natural-language scientific claims,
written for human experts, are often underspecified and so admit several
defensible interpretations. Such verifiers usually pick one, often
tacitly, and benchmarks score their verdicts (feasible, cannot determine,
or infeasible) against expert labels without recording the expert's
interpretation or the verifier's in a comparable form. This leads to
disagreements with expert labels that cannot be traced to the evidence or
the interpretation, and to compute spent on unchecked interpretations. We
introduce \emph{Claim Cards} to make such interpretations explicit in the
reasoning itself. A Claim Card records a verifier's interpretation as nine
fields before evidence gathering, each marked \emph{stated},
\emph{inferred}, or \emph{default}. To test whether an agent's
interpretation can change its verdict, we ran a GPT-5.4 agent five times
on each of 60 expert-labeled materials-science and biomedical claims with
a card, without one, and with one card frozen across runs (900 runs). Over
40\% of card fields were not marked stated, and a card added about a tenth
to a run's GPT-5.4 tokens and time without measurably changing agreement
with expert labels. The agent's own interpretations varied between runs,
although within a claim this variation did not measurably track the
verdict. A deliberate change to one field of the frozen card, however,
making the quantifier universal or the threshold for support stricter,
changed the verdict on about a third of the claims, against about one in
seven expected by chance ($p < 0.001$ for each edit). On 24
materials-science claims run with computational tools, the stricter
threshold also changed what the agent asked those tools to compute. We
therefore argue that verdicts should be reported with their
interpretations and that benchmarks could publish the interpretation
behind each label.
\end{abstract}

\section{Introduction}
\label{sec:intro}

\begin{figure*}[t]
\centering
\includegraphics[width=0.83\textwidth]{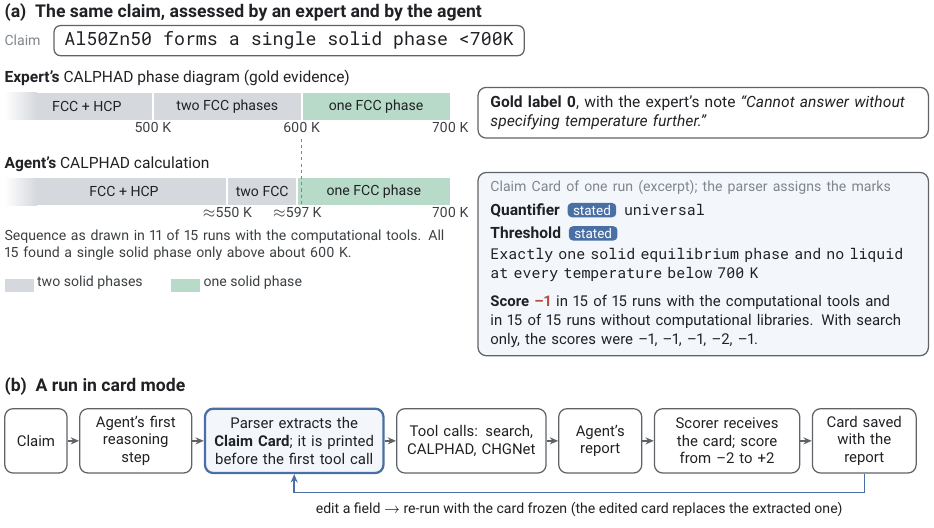}
\caption{(a) The opening claim, assessed by its labeling expert and by
the agent, whose cards required a single solid phase at every
temperature below 700\,K. The expert's boundaries are rounded
(Appendix~\ref{app:gold}). In this card, from a run with the computational
tools, the parser marked the quantifier and threshold \emph{stated},
though the expert judged the temperature range unspecified. (b) A run
in card mode and a re-run with an edited, frozen card.}
\label{fig:overview}
\end{figure*}

Does the alloy in the following claim form a single solid phase at some
temperature below 700\,K, or at every temperature below it?

\ClaimBox{SFBench, materials science}{Al50Zn50 forms a single
solid phase <700K}

\noindent The expert who labeled the claim computed the Al--Zn phase
diagram. It shows two solid phases up to about 600\,K and a single one
above, and the expert gave the label 0, ``cannot determine,'' with the
note ``Cannot answer without specifying temperature further.'' The large
language model (LLM) agent we study chose an interpretation instead. Each
time it was asked to write down its interpretation before gathering
evidence, that interpretation required a single solid phase at every
temperature below 700\,K. When it could run the same kind of
calculation, all 15 of its runs found a single solid phase only above
about 600\,K, as the expert's diagram does, and each
scored the claim $-1$ (Figure~\ref{fig:overview}a). In 20 runs without
computational libraries or with search only, it scored the claim $-1$ or
$-2$.

A claim like this is underspecified. It omits a
detail needed to assess it, here the temperature range over which a single
phase must hold, so several interpretations remain defensible.
\emph{Ambiguity}, as we use the term, is the open choice among
them. Underspecification is a property of the claim, and the ambiguity is
a choice left to whoever assesses it.\footnote{We use ambiguity broadly,
to include the vagueness of an unstated threshold
(Appendix~\ref{app:related-more} relates both terms to prior work).} Even a claim that looks complete can leave such details open,
and Appendix~\ref{app:gan} lists thirteen in one materials-science claim.

We study ambiguity in claim feasibility assessment. In this task, as posed
by SFBench \citep{costello-etal-2026-sfbench}, a system judges how likely a
scientific claim is to be feasible, on a scale from $-2$ (infeasible)
to $+2$ (feasible), and we read the middle label, 0, as ``cannot
determine.'' We also use
biomedical claims from SciFact \citep{wadden-etal-2020-fact}, whose
support and contradict labels we map to the ends of the feasibility
scale. Its claims, which experts rewrote from citation sentences in
biomedical papers, can be far less specific than the opening claim. A verifier testing ``Autophagy declines
in aged organisms'' must decide whether the claim covers every aged
organism or the typical one, which assay measures autophagy and in what
units, and how large a decline counts as support. An automated verifier
answers these questions by choosing a probable interpretation, whose
defaults stay implicit in its reasoning and may differ on the next run.

Feasibility benchmarks score a verdict against one gold label, and
neither the expert's interpretation nor the system's is recorded in a
form that can be compared field by field. When a verdict disagrees with
the label, the benchmark cannot show whether the system and the expert
weighed the evidence differently or interpreted the claim differently. On
the opening claim, the agent's calculations agreed with the expert's
diagram whenever it could run them, but its verdict still disagreed with
the label, which the expert based on the unstated temperature range. An agent that assesses
benchmark claims in a batch has no one to ask about a gap, so it proceeds
on assumptions it may never state, and its reasoning trace need not show
which of them drove the answer. A system that later uses the
verdict, for validation or hypothesis generation, inherits an interpretation that was never written down, and so
does the person who reads it. Where a claim calls for calculations that
take minutes each, the interpretation can also affect which calculations
are run, before anyone has checked it.

We introduce the \emph{Claim Card}. A Claim Card is a structured
interpretation of a claim, recorded before any evidence is gathered. Its
nine fields range from the system the claim is about to the quantifier and
the threshold that determines whether the claim is supported, and each
value is marked as stated in the claim, inferred, or assumed by default
(Section~\ref{sec:claim-card}). A second model extracts the card from the
agent's first reasoning step, and the card is printed before the agent's
first tool call runs. The agent's user can therefore see the
interpretation it stated, with the fields the second model filled in, and a
user who disputes a field can edit it and re-run the agent with the edited
card frozen. Like a preregistration, a card states an interpretation
before the evidence is in, although models write it, not a person.

A natural question to ask is what Claim Cards provide for the overhead
they add. We study Claim Cards in the context of
\system{},\ifreviewversion\footnote{The system's name is anonymized for
blind review.}\fi{} an agent that
assesses claims under domain-specific verification guidelines. It uses a
search agent and a coding agent as tools and outputs a feasibility score.
Using a sample of 30 materials-science claims from SFBench and 30
biomedical claims from SciFact, we
tested \system{} under three conditions: (1) with a Claim Card, (2)
without a card, i.e., the agent is not asked to write one, and (3) with
one card frozen across runs. Each claim was run five times in each condition, for 900 runs. On materials-science claims the agent can call two computational tools, CALPHAD for
phase equilibria and CHGNet for energies and forces. The 60 claims were run without computational libraries, which these tools need, so on those claims the agent relied on
search. We repeated the three conditions with the computational tools on
24 materials-science claims, six of them among the 60, and
also ran their frozen cards with search only. Each edited card
is a copy of a frozen card with the quantifier set to universal or with a
stricter threshold for support. Edited cards were run once on the 60 claims and, on the 24, once with the
computational tools and once with search only. We ask five
questions:
(1) How much of a claim's interpretation is left unstated, and in which
fields? (2) How does the agent resolve the ambiguity, and does it resolve
it the same way in every run? (3) Can a different interpretation change
the verdict? (4) Does a card match its claim at little cost, and does
recording the interpretation change accuracy or stability? (5) How does a
card affect the agent's use of the computational tools?

Claims leave much of their interpretation unstated. On the 60 claims,
with GPT-5.4 as the backbone, more than 40\% of card fields were not
marked as stated in the claim, and the gaps fell
in fields such as the units, the comparator and the threshold rather than
in the entity the claim is about (Section~\ref{sec:underspec}). With or without a card, the agent's first
reasoning step seldom called any value an assumption or a default, and its
interpretation of a claim varied from run to run, although within a claim
this variation did not measurably track the verdict
(Section~\ref{sec:consistency}). When we edited one field of a frozen
card, however, the verdict could change. Without computational libraries, the edited
run's verdict took a different sign from the claim's usual one on 19 of 56
claims with a universal quantifier and on 19 of 58 with a stricter
threshold, nearly always a lower one. Both counts are more than twice
those expected had the edits no effect ($p < 0.001$ for each), and most of
the changed verdicts no longer agreed with the expert's label. With the
computational tools, both edits lowered the feasibility score by about half a point, and the
edited value reached what the agent itself wrote, not only the scoring step
(Section~\ref{sec:control}).\footnote{With a
threshold requiring the single phase to persist over a window of
compositions, the agent calculated three compositions of the opening
claim's alloy, where all five runs with the unedited card had calculated
one.} A
judge model checking one card per claim flagged no
field as unsupported by or contradicting the claim on 50 of the 60 cards,
and in the runs without computational libraries a card added about a tenth to a run's tokens and time. Runs with and
without a card did not differ measurably in accuracy, stability or, with
the computational tools, how often the agent calculated
(Sections~\ref{sec:faithful}--\ref{sec:tooluse}). Both the verdict and,
on claims that need calculations, what the agent calculates can depend on
the interpretation it works from, and we argue that a verdict should be
reported together with that interpretation.\footnote{The cards, the run
records and the code, including all prompts, will be released upon
publication.}

\section{Interpretations and Claim Cards}
\label{sec:claim-card}
\label{sec:interpretations}

\subsection{The Fields of an Interpretation}
\label{sec:fields}
\label{sec:anatomy}

What must an automated system decide before it can assess a claim? An
\emph{interpretation} of a claim is the set of values adopted for nine
fields, whether the claim states them or not. They are the details that
\system{}'s verification guideline asks the agent to record before it uses
any tool (Appendix~\ref{app:prompts}), and each is a dimension along which two defensible
interpretations of the same claim can differ. Table~\ref{tab:card-example}
defines them in three groups. The \emph{system specification} defines the
object of study and its baseline before any intervention, the
\emph{intervention} is the independent variable, and the \emph{evaluation
criteria} define the dependent variable and what counts as support.

The quantifier records whether the claim is \emph{existential} (``can''),
\emph{typical} (``generally'') or \emph{universal} (``always''), and it
sets the guideline's stopping rule. One condition-matched instance that
clears the threshold suffices for an existential claim, whereas one
condition-matched counterexample is decisive negative evidence against a
typical or universal one (Appendix~\ref{app:card-more}).

A \emph{Claim Card} gives a value for each field, and a tenth field,
\field{Provenance}, marks each value \emph{stated} (given in the claim),
\emph{inferred} (following from the text with domain knowledge) or
\emph{default} (assumed where the text is silent). In
Table~\ref{tab:card-example}, the units are a default because a five-word
claim names no assay, and the manipulation is inferred because ``aged''
implies aging as the treatment. Whether the opening claim requires a
single solid phase at every temperature below 700\,K or only at some
temperature is a choice of \field{Quantifier} and \field{Threshold}.
Because every claim is described by the same fields, we can count which
of them a claim leaves unstated (Section~\ref{sec:underspec}), compare the
interpretations of different runs field by field
(Section~\ref{sec:consistency}), and change one field while holding the
others constant (Section~\ref{sec:control}).

\begin{table*}[t]
\centering
\footnotesize
\setlength{\tabcolsep}{4pt}
\begin{tabularx}{\textwidth}{@{}l>{\raggedright\arraybackslash}p{5.6cm}>{\raggedright\arraybackslash}Xl@{}}
\toprule
Field & What it records & Value on this card & Prov. \\
\midrule
\multicolumn{4}{@{}l}{\emph{System specification}} \\
\field{SystemEntity} & The object of study & Organisms undergoing aging & stated \\
\field{StatePhase} & Its phase or state before any intervention & Aged versus young/adult & stated \\
\field{Conditions} & The conditions in which the claim is purported to hold & Matched measurement conditions; taxa and tissues must be identified when evaluating generality & stated \\
\multicolumn{4}{@{}l}{\emph{Intervention}} \\
\field{Manipulation} & How the system is modified, processed, or intervened upon & Aging; no experimental intervention specified & inferred \\
\multicolumn{4}{@{}l}{\emph{Evaluation criteria}} \\
\field{ObservableFoM} & The observable or figure of merit & Autophagy activity or flux & stated \\
\field{Units} & The units in which the observable is measured & Assay-dependent; no units specified & default \\
\field{Quantifier} & Existential, typical or universal & typical & stated \\
\field{Threshold} & The rule that turns a verb like ``enhances'' into a testable criterion & A documented age-associated reduction in autophagy activity or flux under matched conditions & stated \\
\field{Comparator} & The baseline for the observable, and the AND/OR logic across sub-claims & Aged organisms versus young/adult controls; single claim & stated \\
\bottomrule
\end{tabularx}
\caption{The nine fields of an interpretation, and the Claim Card that
\system{} produced for ``Autophagy declines in aged organisms'' in the
first card run, with the provenance mark of each value. It was also the
claim's frozen card.}
\label{tab:card-example}
\end{table*}

\subsection{How a Card Is Produced and Used}
\label{sec:generation}
\label{sec:visibility}

The agent writes its interpretation as free text in its first reasoning
step, a short paragraph (median about 90 words) that names the system and
the evidence the agent will seek. The guideline tells it to
``prefer the weakest faithful interpretation of a claim'' where the claim
leaves a gap. A second model, the \emph{parser} (GPT-6 Sol), converts that
step into a card conforming to a JSON schema (Appendix~\ref{app:schema}).
We separate writing the interpretation from structuring it so that every
decision the agent states is made by the model that has the guideline and
the full problem context. The parser sees only the first
reasoning step, not the claim. It fills fields the step leaves implicit
and assigns every provenance mark, so the marks are the parser's judgment
(Appendix~\ref{app:card-production}).

In card mode, \system{} extracts the card as soon as the agent has written
its first reasoning step, before the tool call that the step selects is
executed, prints it, and saves it with the report when the run ends
(Figure~\ref{fig:overview}b). The
agent gathers evidence under its own written interpretation and is not shown the parsed card. The scorer,
which assigns the feasibility score, receives the claim, the parsed card,
the agent's reasoning and its final answer, and scores the claim \emph{as
interpreted}. In \emph{frozen} mode, which we used for the frozen and
edited cards of Section~\ref{sec:conditions}, \system{} instead injects a
stored card into the agent's prompt, with an instruction to use it exactly
as given and not to change the claim's interpretation, quantifier,
threshold or defaults, and passes the same card to the scorer. A user who disputes a field could re-run the agent in this mode with an
edited card.

\section{Experimental Setup}
\label{sec:setup}

\paragraph{System:} \system{} is a ReAct agent \citep{yao-etal-2023-react}
for claim feasibility assessment. The claim's domain selects its tools, a
verification guideline and a scoring rubric. Under the guideline, the
agent first tries to falsify the claim and then looks for one
condition-matched measurement, with stopping rules set by the quantifier.
Its tools are a search agent (Claude Sonnet 4.6 with web search) and a
coding agent (Claude Sonnet 5.5). GPT-5.4 with high reasoning effort is
the LLM backbone of the agent, the scorer and the report step.\footnote{The
provider sets its sampling temperature, so run-to-run variation is a
property of the model that we measure rather than control
\citep{atil-etal-2025-non, yuan-etal-2025-understanding}.} Appendix~\ref{app:setup-more} describes the
tool-use policies and the other components.

\paragraph{Tool settings:}
\label{sec:tools-setup}
On materials-science claims, the agent reaches the coding agent only
through two computational tools, which compute phase equilibria with
pycalphad \citep{otis-liu-2017-pycalphad} and the COST~507 database
\citep{ansara-etal-1998-cost507} (CALPHAD), or energies and forces with
the machine-learned interatomic potential CHGNet
\citep{deng-etal-2023-chgnet}. We ran
60 claims \emph{without computational libraries}, so the agent relied
mainly on search.\footnote{The agent called a computational tool in 86 of the 450
materials-science runs on these claims, but 89 of the 90 calls returned no
calculation.} A set of
24 materials-science claims was run \emph{with the computational tools},
and their frozen and edited cards also with \emph{search only}, the two
tools removed. The two sets share six claims but differ in tool setting, in
the source of the frozen card and in when they were run, so we compare
conditions within each set and draw no inference from one to the other.

\paragraph{Claims:}
\label{sec:data}
The 60 claims run without computational libraries are 30
materials-science claims from SFBench
\citep{costello-etal-2026-sfbench}, each
with a five-point label and an expert's rationale, and 30 biomedical claims
from the SciFact development set, labeled \textsc{support} or
\textsc{contradict}, which we compare with a run's score by sign. Each half is stratified by label (Appendix~\ref{app:ids}).\footnote{No
stored search result came from SciFact's release pages or named SFBench or
SciFy, but the stored records are condensed and cannot rule
out exposure (Appendix~\ref{app:setup-more}).} The 24 claims run with the
computational tools were chosen at random, before any run, from the 37
SFBench claims tagged for computational tools. Their gold
labels lean positive
(Section~\ref{sec:limitations}).

\paragraph{Conditions and edits:}
\label{sec:conditions}
On the 60 claims without computational libraries and on the 24 with the
computational tools, we ran each claim five times in each of three
conditions.
In \textbf{card} runs, the agent records its interpretation and the parsed
card goes to the scorer. In \textbf{no-card} runs, the guideline loses its subsections on the card,
verb mapping, defaults and comparators but keeps its plausibility checks
and quantifier-based stopping rules, so the agent must still decide how
broad the claim is but is not asked to record it, and no step sees a card.
In \textbf{frozen} runs, all five runs receive the same stored card, so
that any variation that remains can be attributed to later steps. For
the 60 claims it is the card of the claim's first card run, and for the
24 the card of a separate preliminary card run. Two further conditions
edit one field of the frozen card to make the claim harder to satisfy, so
that the score should not rise if the agent follows the card. The
quantifier edit sets the quantifier to \emph{universal} (56 of the 60
claims and 20 of the 24; the other four frozen cards in each set were
already universal). The threshold edit (58 and 24 claims) replaces the
threshold with a stricter criterion, which Claude Opus 5.5 wrote under an
instruction to keep it plausible to a domain expert. We did not check that each edited value
is a defensible interpretation of its claim. Some threshold edits also
demand a kind of evidence, such as replication
(Appendix~\ref{app:edits-more}). Each edited card was run once, and those
of the 24 claims once in each setting. An edit's \emph{shift} is the edited
run's score minus the mean of the five frozen runs with the unedited card
in the same setting, whose lowest and highest scores bound the
\emph{unedited range}. The 60 claims received 900 repeated and 114
edited-card runs and the 24 claims 480 and 88, for 1,582 GPT-5.4 runs on
78 claims, not counting a 20-claim pilot (Table~\ref{tab:design} in
Appendix~\ref{app:setup-more}).

A search outage made searches fail in 17 frozen and 10 edited-card runs
without computational libraries, and we report the affected results with
and without them. Frozen runs were made at a different time from card and no-card runs, so
we treat comparisons of frozen runs with either as descriptive
(Section~\ref{sec:limitations}).

\paragraph{Measures and statistics:}
\label{sec:measures}
\label{sec:stats}
In a field audit, a judge model (Gemini 3.1 Pro) labeled each field of the
first-run card of each of the 60 claims against the claim as faithful, a
reasonable default, unsupported or contradicting.
\emph{Consistency} is the share of a claim's run pairs whose values of a
field express the same interpretation, as grouped by a second judge
(Claude Opus 5.5), averaged over the nine fields. For no-card runs, the parser recovered an implicit interpretation from
each run's first reasoning step and final report. Both judges come
from model families other than the backbone's
\citep{zheng-etal-2023-judging}. We call a run's feasibility score its
\emph{verdict}. \emph{Sign accuracy} is the share of runs whose score has
the sign of the gold label, with 0 counted as a sign of its own. A run
\emph{abstains} when it scores 0, and run-to-run variance
is the standard deviation of a claim's five scores. Every interval is a
95\% percentile interval from a bootstrap over claims with 10,000
resamples, paired by claim, and where an interval includes zero we say
that the conditions did not differ measurably.\footnote{Appendix~\ref{app:stats}
also resamples groups of claims among the 24 that share a chemical
system.}
Intervals are given in the tables and in footnotes.

\section{Results}
\label{sec:results}

\subsection{Claims Leave Much of Their Interpretation Unstated}
\label{sec:underspec}

Using the provenance marks, we counted how much of each card the claim
itself supplies. In the 300 cards from card runs on the 60 claims (five runs
each), 42\% of field values were not marked stated in
materials-science cards and 44\% in biomedical cards
(Figure~\ref{fig:unstated} in Appendix~\ref{app:underspec-more}).\footnote{The marks are the parser's judgment, but the audit judge, which
saw each claim's first-run card together with the claim, also called 33\%
of materials-science fields and 46\% of biomedical fields reasonable
defaults, values that the claim does not give but that an expert would
accept.} Materials-science cards almost always marked
the system (98\%) and the observable (97\%) as stated. The gaps fell
instead in fields that can determine whether the claim is feasible, even
though experts wrote these claims. Units were not marked stated in 68\% of
materials-science cards and 99\% of biomedical cards, the comparator in
68\% and 61\%, and the threshold in 35\% and 45\%. Cards for the 24 claims
run with the computational tools left the same kinds of field open, and 36\% of their values were not
marked stated (Appendix~\ref{app:underspec-more}).

\subsection{Interpretations Vary Between Runs, Most in the Threshold, Comparator and Conditions}
\label{sec:consistency}

Does the agent fill these gaps the same way in every run, and does it
say which values it assumed? In no-card runs, nearly every first reasoning
step named the entity the claim is about and the evidence the agent would
seek, but in both conditions few first steps used the words ``default'' or
``assume'' (Appendix~\ref{app:consistency-more}). On the 50 of the 60 claims judged
in both conditions, consistency was 74\% for a claim's five explicit cards
and 70\% for the interpretations recovered from no-card runs, and the two
did not differ measurably.\footnote{The paired difference was $+0.04$
\ci{$-0.01$}{$+0.09$}. Appendix~\ref{app:consistency-more} also compares
interpretations recovered in the same way from both conditions.} The runs with a card
agreed least on the threshold (60\%), comparator (62\%) and conditions
(69\%), against 88\% for the system and 83\% for the quantifier
(Figure~\ref{fig:consistency} in Appendix~\ref{app:consistency-more}),
and these are fields that the parser may fill in when the first reasoning
step leaves them implicit (Section~\ref{sec:limitations}). The quantifier was unanimous across
the five cards for 34 of 51 claims (Fleiss' $\kappa$ 0.70), but all nine
fields were judged to express the same interpretation for only 2 of 51
claims, and for 1 of 22 with the computational tools. Over five runs, nearly every
claim thus received more than one interpretation in at least one field.
Without computational libraries, runs of a claim whose cards agreed on the threshold
had the same sign about as often as runs whose cards differed (31
claims),\footnote{Sign agreement was 0.03 \ci{$-0.03$}{$+0.09$} higher
between runs whose thresholds agreed.} and the same held for the quantifier
(Appendix~\ref{app:consistency-more}).

\subsection{Changing the Interpretation Can Change the Verdict}
\label{sec:control}

\begin{table}[t]
\centering
\footnotesize
\setlength{\tabcolsep}{3pt}
\begin{tabular}{@{}lrlrc@{}}
\toprule
 & & & Did not & Below\,/ \\
Edit & $n$ & Shift & increase & above \\
\midrule
\multicolumn{5}{@{}l}{\emph{Without computational libraries (60 claims)}} \\
Quantifier & 56 & $-0.62$ \ci{$-0.91$}{$-0.35$} & 47 & 16\,/\,0 \\
Threshold & 58 & $-0.54$ \ci{$-0.79$}{$-0.32$} & 50 & 14\,/\,1 \\
\midrule
\multicolumn{5}{@{}l}{\emph{With the computational tools (24 claims)}} \\
Quantifier & 20 & $-0.55$ \ci{$-1.04$}{$-0.10$} & 16 & 4\,/\,1 \\
Threshold & 24 & $-0.48$ \ci{$-0.80$}{$-0.20$} & 23 & 2\,/\,0 \\
\midrule
\multicolumn{5}{@{}l}{\emph{Search only (24 claims)}} \\
Quantifier & 20 & $-0.48$ \ci{$-0.80$}{$-0.20$} & 19 & 4\,/\,0 \\
Threshold & 24 & $-0.14$ \ci{$-0.43$}{$+0.13$} & 18 & 3\,/\,2 \\
\bottomrule
\end{tabular}
\caption{Score shifts after one card field is edited to make the claim harder to satisfy (Section~\ref{sec:conditions}), with bootstrap intervals. The last two columns count claims whose edited score did not exceed the unedited mean and claims whose edited score fell below every unedited run or rose above the unedited range. The two identical shifts on the 24 claims are separate estimates.}
\label{tab:edits-main}
\end{table}

To test whether a different interpretation can change the verdict, we ran
the agent once with each edited card, in which one field of the frozen
card was changed to make the claim harder to satisfy (Section~\ref{sec:conditions}). On
the 60 claims, both edits lowered the score on average, as a stricter
interpretation implies (Table~\ref{tab:edits-main}). The quantifier edit
changed the score by $-0.62$ points over 56 claims,
and the threshold edit by $-0.54$ over 58. The edited
score fell below every unedited run of its claim for 16 of 56 claims under
the quantifier edit (3.7 expected had the edit no effect) and 14 of 58
under the threshold edit (3.5 expected), and its sign differed from the
most common unedited sign for 19 claims under each edit (7.5 and 8.3
expected; Appendix~\ref{app:edits-more}).\footnote{Both edits lowered
the score in each domain, most on biomedical claims ($-0.80$ and $-0.66$),
whose runs seldom abstained. The shifts barely changed without the edited
runs that the search outage affected (Table~\ref{tab:edits} in
Appendix~\ref{app:edits-more}), and the threshold shift did not rest on
the edits that also demand a standard of evidence.} For the autophagy claim of
Table~\ref{tab:card-example}, all five unedited runs scored $+2$, the
correct verdict. With the quantifier set to universal, the score was $-1$,
and the agent's own answer said ``likely false as stated ($-1$),
because the universal quantifier is contradicted.''

With the computational tools, both edits also lowered the score
(Table~\ref{tab:edits-main}).\footnote{Without the claims on which an
edited run lost a CALPHAD result or misread a mole fraction as a
percentage, the shifts were $-0.47$ for the quantifier and $-0.29$ for the
threshold, and both intervals still exclude zero (Table~\ref{tab:edits}
in Appendix~\ref{app:edits-more}).} With search only, the quantifier edit
lowered the score, but the threshold edit's interval includes zero.

On materials-science claims, a stricter threshold often left the agent
without evidence that met it, and the edited run abstained
(Appendix~\ref{app:edit-abstention}).\footnote{Of the threshold-edited
materials-science runs without computational libraries, 19 of 29 scored 0,
against 49 of 145 unedited runs, and abstention also rose with search
only.}

The scorer receives the card, so a lower score could come from the scorer
alone. The edited value, however, also reached what the agent itself wrote. In
89 of the 114 edited runs on the 60 claims, the agent's own answer stated
a score, and, descriptively, shifts computed from those stated scores alone
were $-0.67$ and $-0.48$, with intervals that exclude zero
(Appendix~\ref{app:edits-more}). With the computational tools, the edited
threshold appeared in the questions the agent sent to the tools or the
search agent in 22 of 24 edited runs and in none of 120 unedited runs
(Appendix~\ref{app:propagation}).

The edits thus show that an interpretation given to the agent reaches its
verdict, through what the agent itself writes and not only through the
scorer. They do not show which interpretation of a claim is right, and because every edit made the
claim harder to satisfy, they cannot separate the edited value from a
general caution under added constraints (Section~\ref{sec:limitations}).

\subsection{How Well a Card Matches Its Claim, and What It Costs}
\label{sec:faithful}
\label{sec:overhead}

The audit judge flagged few fields as unsupported by or
contradicting the claim (3\% of materials-science fields and 4\% of
biomedical fields), and 25 of 30 cards in each domain had no flagged
field. The provenance mark agreed with the judge's label on 77\% of
materials-science fields and 59\% of biomedical fields.\footnote{A stated
mark agrees with a faithful label and any other mark with any other label.
Most biomedical disagreements were fields marked stated that the judge
called reasonable defaults (Appendix~\ref{app:faithful-more}), as for the
\field{Conditions} and \field{Quantifier} values of the card in
Table~\ref{tab:card-example}. We traced
two of the flagged errors to the parser, which does not see the claim
(Figure~\ref{fig:overview}a shows another disputed mark).} Recording the
interpretation added about a tenth to the tokens and time of a run without
computational libraries.\footnote{It added 13\% more GPT-5.4 input tokens
and 8\% more wall-clock time (Appendix~\ref{app:faithful-more}).}

\subsection{Observations on Accuracy and Stability}
\label{sec:observations}

A card might also be expected to make verdicts more accurate or more
stable. With GPT-5.4, runs with and without a card did not differ
measurably in sign accuracy or run-to-run variance in either set of claims
(Table~\ref{tab:outcomes}). On
the 60 claims, sign accuracy was 0.68 with a card and 0.71 without,
and on the 24, with the computational tools, 0.67 and 0.64. The interval of the difference on the 60 claims excludes a gain from the card larger than
0.01.\footnote{Card minus no card was $-0.03$ \ci{$-0.07$}{$+0.01$}.
Always predicting a positive score would give a sign accuracy of 0.58 over
the 60 claims and 0.50 on their materials-science claims, where card and
no-card runs (0.53 and 0.56) were about as accurate.} Descriptively, frozen runs on the 60 claims were not measurably more
stable than card runs.\footnote{Descriptively, frozen runs were also less
often right about the sign than no-card runs ($-0.09$), mostly because
they abstained more often ($+0.08$), and the extra abstentions fell on
materials-science claims (Appendix~\ref{app:abstention}).} Where the agent's own
answer stated a score, in 478 of the 900 repeated runs on the 60 claims, the
scorer repeated it in 472, so in those runs little of the variation arose
in the scorer. We think most of it arises in which evidence a run finds
and in how the agent weighs that evidence.
Most abstaining materials-science runs, for example, had at least one
search that came back empty, against about a third of those that decided
the claim (Appendix~\ref{app:abstention}). With the computational tools, even runs
that obtained the same calculation did not always reach the same verdict
(Appendix~\ref{app:samecalc}).

\subsection{Effect of Claim Cards on Tool Use}
\label{sec:tooluse}

With the computational tools, card runs on the 24 claims called CALPHAD or CHGNet in 64 of 120 runs and no-card runs in
72, and per claim the two conditions did not differ measurably
(Table~\ref{tab:tooluse}). A card in the agent's prompt, as in frozen and edited
runs, changed what it asked the tools instead. On each of the three claims
whose edited threshold named a composition range, the edited run
calculated over that range, and the frozen card's wording also reached the
calculation questions (Appendix~\ref{app:tooluse-more}).

\section{Related Work}
\label{sec:related}

\paragraph{Claim verification and feasibility:} Feasibility assessment is
related to evidence-based claim verification, as in FEVER
\citep{thorne-etal-2018-fever} and SciFact \citep{wadden-etal-2020-fact}.
Closest to our task is Matter-of-Fact \citep{jansen-etal-2025-matter}, which labels 8.4k
materials-science claims as feasible or infeasible. Its labels are binary,
so none can record that a claim is too underspecified to judge, as the
expert's 0 on our opening claim does. Its baselines also output a binary
label, and every comparison in this study is between conditions of one
agent. We therefore use SFBench, whose labels
come with an expert's rationale, and do not compare with those baselines.
Claim decomposition into literal and implied subquestions
\citep{chen-etal-2022-generating} and program-guided verification
\citep{pan-etal-2023-fact} also make a claim's parts explicit before
checking it, but neither records the value adopted for each open detail or
its source.

\paragraph{Ambiguity and underspecification:} Language models often
interpret underspecified or ambiguous sentences poorly
\citep{wildenburg-etal-2024-dust, liu-etal-2023-afraid}, and benchmarks
such as AmbigQA \citep{min-etal-2020-ambigqa} test whether models recover
interpretations that annotators listed in advance. No dataset we know of
lists the interpretations of a scientific claim. AmbiFC
\citep{glockner-etal-2024-ambifc}, the closest work on ambiguous claims,
traces annotators' disagreement about veracity partly to
underspecification and models it as a label distribution, with no
interpretation to compare with a card's. Dis2Dis
\citep{staliunaite-vlachos-2025-dis2dis} rewrites an ambiguous claim once
the evidence is in hand, whereas a card is extracted before any evidence
is gathered. For these reasons we compare with neither.

\paragraph{Agents and visible interpretations:} Agents given
underspecified tasks usually proceed on unstated assumptions instead of
asking \citep{vijayvargiya-etal-2026-ambigswe, wang-etal-2025-learning}.
Methods that decide when
a model should ask a clarifying question \citep{zhang-choi-2025-clarify} need
someone to answer during the run, which batch assessment of benchmark
claims does not have. Interfaces that show a model's interpretation of a request and let
users edit it have helped users steer systems in user studies
\citep{tian-etal-2023-interactive, kazemitabaar-etal-2024-improving}.
Our edits are an intervention test in the spirit of
\citet{lanham-etal-2023-measuring}. Appendix~\ref{app:related-more}
discusses further work.

\section{Conclusion}
\label{sec:conclusion}

We studied how an LLM agent resolves the ambiguity of underspecified
scientific claims and introduced Claim Cards to record the interpretation
it adopts at the start of each run. A feasibility verdict is a verdict on
one interpretation of the claim, and a system should therefore report
each verdict with the interpretation behind it. What an agent calculates
can also depend on the interpretation, and a card printed as the run
starts lets a user see the interpretation and stop or re-run the agent before the
calculations finish. The expert who labeled the Al--Zn claim that opens
this paper gave it 0, ``cannot determine,'' because its temperature range
is unstated. An agent that always adopts one interpretation will not
reach such a label for the expert's reason, even when its calculation
agrees with the expert's, so some disagreements between a system and a
gold label may be disagreements about what the claim means. A benchmark
that published the interpretation behind each label would let a system be
scored on the claim the expert assessed.

Without computational libraries, runs whose cards recorded different interpretations
of a claim did not measurably differ more in their verdicts, and runs given
the same frozen card varied about as much as card runs, although a
reduction of up to about half is not ruled out. With the computational
tools, runs whose card recorded a stricter quantifier scored lower on
average, while other comparisons in those runs pointed the other way. The
edits give controlled evidence that a different
interpretation can change the verdict. We think much of
the run-to-run variation arises later, in which evidence a run finds and
how it weighs it, so more stable verdicts will probably need changes
there. Some of what decided the verdicts is
not yet on the card, such as a run's own rule that one counterexample makes
the claim negative or the tolerances under which the coding agent judges a
calculation, and fields for both are worth testing
(Appendix~\ref{app:discussion-more}). Our first test, however, would be
whether domain experts find the fields marked as default useful and
whether their corrections improve verdicts, on more datasets and with
more than one LLM backbone.

\section{Limitations}
\label{sec:limitations}

\paragraph{Sample size:} Our two sets of claims are small, 60 claims and
24, six of them in both. Every claim has an expert label and was run five
times in each condition, so most comparisons set a claim against itself,
but each edited card was run once per setting. Our main result rests on the
edits on the 60 claims, whose score shifts and changes of sign both have
$p < 0.001$ (Appendix~\ref{app:edits-more}). Intervals for the other
contrasts are a few tenths of a point wide on the $-2$ to $+2$ scale for
the 60 claims and wider for the 24, wide enough to miss small effects
\citep{card-etal-2020-little}. We did not adjust them for multiple
comparisons and report those contrasts as exploratory. Several of the 24 claims concern the
same chemical system. With groups of such claims as the unit, nine
intervals included, reached or excluded zero just as they did over single
claims (Appendix~\ref{app:stats}). Because the two sets differ in both
their claims and their setting, we compare them only descriptively.

\paragraph{Backbone, guidelines and data:} We used one LLM backbone, GPT-5.4
at high reasoning effort. In a 20-claim pilot with two other backbones and
one run per claim, the backbones differed more from one another than card
runs did from no-card runs (Appendix~\ref{app:pilot}), so other agents may
resolve the same claims differently. How the agent resolves ambiguity
probably also depends on its guideline, which tells it to prefer the
weakest faithful interpretation. We refined the materials-science
guidelines on claims from earlier SFBench releases, which may include some
of the evaluated claims or claims like them, so the materials-science
results are in-domain in a way the biomedical results are not. SciFact's
claims and labels have been public since 2020 and may be in GPT-5.4's
training data. Its labels refer to the abstracts cited with each claim,
whereas the agent searched the open web. The claims themselves are
factual assertions, while \system{} and its guidelines were built for feasibility
assessment, and we score them by sign only.

\paragraph{Edits:} Every edit made the claim harder to satisfy. Edits that
loosen a field, such as universal to existential, or reword it without
changing its meaning would separate the edited value from a general caution
under added constraints, since changes to a prompt's format alone can
change a model's answers \citep{sclar-etal-2024-quantifying}. By keyword
rules that one coder applied, 30 of the 58 threshold edits on the 60 claims
and 16 of the 24 on the 24 claims also add a standard of evidence, such as
independent replication. On the 60 claims, the threshold shift did not rest
on those edits (Appendix~\ref{app:edits-more}). Setting an existential
quantifier to universal, as in 30 of the 56 quantifier edits, may impose an
interpretation that the claim's wording does not support. The agent was
told to use the frozen card exactly as given, and language models tend to
match their answers to views that a user states
\citep{sharma-etal-2024-towards}, so the edits measure the effect of an
interpretation given to the agent, which may differ from that of an
interpretation the agent adopts on its own.

\paragraph{Card construction:} The parser writes each card from the agent's
first reasoning step, without the claim, and fills fields that the step
leaves implicit. A card therefore mixes what the agent decided with what
the parser filled in, and a value can be marked stated because the agent's
step states it, whether or not the claim does. The audit judge, which saw
the claim and checked one card for each of the 60 claims, flagged 3\% of
materials-science fields and 4\% of biomedical fields as unsupported by or contradicting it
(Section~\ref{sec:faithful}), but the audit could not show which values
came from the agent. The fields that the first step can leave implicit,
such as the threshold, comparator and conditions, are also those on which
cards agreed least (Section~\ref{sec:consistency}), and part of their
run-to-run variation may come from the parser, since what it fills in
depends on the wording of each run's first step.

\paragraph{Judges and coding:} The field audit and the consistency groupings
come from judge models, and agreement between LLM judges and people varies
widely by task \citep{bavaresco-etal-2025-llms}. Beyond the cases traced in
Section~\ref{sec:faithful}, their labels were not checked systematically by
a person, and the audit judge saw each card's provenance marks, which may
have anchored its labels. The consistency analysis covers 51 of the 60
claims and 22 of the 24 (Appendix~\ref{app:consistency-more}). Other
analyses, among them the reasons for abstention and the comparison of the
agent's calculations with the experts', rest on codes that one coder
assigned.

\paragraph{Tool settings:} The libraries that the computational tools need
were not installed for the runs on the 60 claims, so materials-science runs
relied mainly on search, and their accuracy and abstention reflect that
setting (Section~\ref{sec:tools-setup}). Biomedical runs
could call a general coding agent for any check, which may account for
part of the gap in abstention between the domains
(Appendix~\ref{app:abstention}). The results with the tools mostly describe
CALPHAD use, which made up 56 of the tool calls in card runs on the 24
claims, against 10 for CHGNet.

\paragraph{The 24 claims and their gold labels:} Although these claims were
drawn at random from the 37 claims tagged for computational tools, their
labels lean positive. Of their 24 labels, 14 are positive, and 8 of the 12
tagged claims labeled $-2$ are not among them, an imbalance that a random
draw would rarely produce (Appendix~\ref{app:stats}). Accuracy on these
claims should be judged against the 0.58 that always predicting a positive
score would reach, whereas paired contrasts compare each claim with itself
and are affected only if conditions act differently on claims with
different labels. The CALPHAD claims were distributed with the COST~507
database file that the agent also used, so an agent's CALPHAD calculation
and the expert's probably share a database, and their agreement is not
independent evidence that either is right. The expert's code or explanation has documented problems on 8 of
the 24 claims, such as Mg and Zn fractions swapped relative to the claim,
and on two claims, one of them among the eight, the expert computed a proxy
for the claimed quantity (Appendix~\ref{app:discussion-more}). We
therefore treat accuracies on these claims as descriptive, and
Appendix~\ref{app:gold} repeats the accuracy contrasts without these nine
claims.

\paragraph{Run timing:} The frozen and edited-card runs on the 60 claims were
made together, hours after most card and no-card runs, and a search outage
made searches fail in 17 frozen and 10 edited-card runs. On the 24 claims,
the frozen cards came from a separate preliminary run, the frozen runs and
their quantifier edits were made a day before the other runs, and some card
and no-card runs that called a tool used a thermodynamic database other
than COST~507 (Appendix~\ref{app:tools}). Comparisons of frozen
runs with card or no-card runs span these differences, and we report them
descriptively. The
edit shifts on the 60 claims compare runs made together, and they barely
changed without the runs hit by the outage (Section~\ref{sec:control}). On
the 24 claims, the threshold-edit shifts compare runs made a day apart.

\paragraph{Controls and users:} In the no-card condition, the guideline also
loses its sections on verb mapping, defaults and comparators, and the
scorer and the report step receive no card, so the card and no-card
comparison tests the card condition as a whole rather than recording the
interpretation alone. Every edited card reached both the agent and the
scorer, and Section~\ref{sec:control} separates the two paths through the
scores that the agent itself stated. No domain expert edited a card or
re-ran the agent in this study. Whether seeing the interpretation at the
start of a run helps a user catch a wrong default, or leads users to trust
a verdict more than they should, is a question for a user study with an
interface for the card.

\section*{Ethics Statement}

Claim Cards are meant to make an automated verdict easier to inspect and
to contest. They do not make a verdict correct, and a card that reads well
can still rest on a wrong default. We intend the provenance marks to direct
a user's attention to the fields marked as default. A card
may also lead users to rely on a verdict more than they should, since
explanations can increase reliance on a system regardless of whether it is
right \citep{bansal-etal-2021-whole, si-etal-2024-large}, although they
can reduce over-reliance when they make the system's answer cheaper to
check \citep{vasconcelos-etal-2023-explanations}. The provenance marks are themselves a
second model's judgment, not a check of the claim.

Because a frozen card reaches both the agent and the scorer, a card is
also a lever on the verdict. Without computational libraries, an edit to one field lowered scores by 0.54 to
0.62 points on average (Section~\ref{sec:control}), so a verdict produced
under an edited card should say so, and a saved card should record who
changed which field. A score of 0 can reflect failed retrieval rather than
an underspecified claim, and most abstaining materials-science runs had at
least one empty search (Section~\ref{sec:observations}). A report that
lists failed searches and tool calls next to the score lets a reader tell
the two apart. The guideline's preference for the weakest faithful
interpretation leans toward existential interpretations, which a claim
satisfies more easily, and the card's quantifier shows when a positive
verdict rests on one. Verdicts on
biomedical claims are not assessments of medical evidence, and SciFact's
labels refer to the abstracts cited with each claim.

The SciFact claims, from its authors' release, are distributed under the
CC BY 4.0 license. We use the claims, labels, rationales and expert code
of SFBench as distributed, and the CALPHAD claims
came with the COST~507 database file. The released run records will
contain text that the search agent returned from public web pages, with
each page's URL. No human subjects were involved.

\ifreviewversion\else
\section*{Acknowledgments}

This research was developed with funding from the Defense Advanced Research
Projects Agency's (DARPA) SciFy program (Agreement No.\ HR00112520300). The
views expressed are those of the authors and do not reflect the official
policy or position of the Department of Defense or the U.S. Government.
\fi

\ifaclstyle\else\bibliographystyle{acl_natbib}\fi
\bibliography{references}

\appendix

\raggedbottom

\section{Claim Cards in Detail}
\label{app:card-more}

\subsection{Underspecification in a Single Claim}
\label{app:gan}

Even a claim that looks complete leaves details open, like this
materials-science claim:

\ClaimBox{SFBench, materials science}{GaN thin films that have been
homoepitaxially deposited on c-plane GaN substrates can be bonded to a
flexible polymer substrates [\emph{sic}] followed by subsequent removal of the
homoepitaxial substrate. Mechanical straining of the film/polymer substrate
resulted in strain-induced bandgap tuning over a 200 meV range.}

\noindent The claim names the material, the substrates, the process, and a number,
but does not say how thick the film is, how it is bonded to the polymer, whether
the strain is uniaxial or biaxial, tensile or compressive, whether the
film deforms only elastically, how the bandgap was measured, or whether
the 200\,meV range spans tension and compression or is measured from the
unstrained film. It leaves at least the following thirteen details open.

\begin{enumerate}[leftmargin=*,widest=13,itemsep=1pt,topsep=3pt]
\item The film thickness and lateral dimensions, which determine strain transfer efficiency, fracture limits, and achievable property modulation, are unspecified.
\item The bonding method and interfacial layer between the film and the flexible substrate are not described, leaving unclear how effectively strain is transferred and whether the interface alters properties.
\item The type and mechanical properties of the flexible substrate are unspecified, and they affect which strain states can be reached and how uniform and controllable the strain is.
\item The applied stress state is ambiguous (uniaxial vs.\ biaxial, tensile vs.\ compressive, magnitude, and crystallographic direction), which influences band structure changes.
\item It is unclear whether the deformation is purely elastic and reversible or involves plasticity or defect formation, which affects feasibility and repeatability.
\item The measurement modality and conditions for bandgap determination are unspecified (e.g., optical method, temperature control, mitigation of heating), which can confound attribution to strain.
\item The meaning of the reported tuning ``range'' is ambiguous, i.e., whether it spans both tensile and compressive extremes or is relative to an unstrained baseline under a single loading mode.
\item The spatial uniformity and area over which the strain is applied and measured are not stated, so an observed shift may be representative or only a local average.
\item The initial defect density and residual stress after transfer and substrate removal are unspecified, which influence mechanical limits and baseline bandgap.
\item Environmental conditions during straining and measurement (temperature, atmosphere, humidity) are not defined, yet can shift optical properties.
\item The strain magnitude required to achieve the stated bandgap modulation is not provided, nor is whether it is within the fracture limits of the material-substrate system.
\item Process yield, reproducibility, and durability under repeated strain cycles are not discussed.
\item Material properties such as doping level and surface passivation are unspecified, which can alter optical spectra and apparent bandgap.
\end{enumerate}

These gaps, like the open questions in the examples of
Section~\ref{sec:intro}, fall in the fields of Table~\ref{tab:card-example}.
For the autophagy claim, the \field{Quantifier} records how broadly the
claim applies across aged organisms, the \field{ObservableFoM} and
\field{Units} which assay measures autophagy and in what units, and the
\field{Threshold} the size of decline that counts as support. In the GaN
claim, the unstated stress state belongs to the \field{Manipulation}, and
whether the 200\,meV range is measured from the unstrained film belongs to
the \field{Threshold}.

\begin{figure*}[p]
\centering
\includegraphics[width=\textwidth]{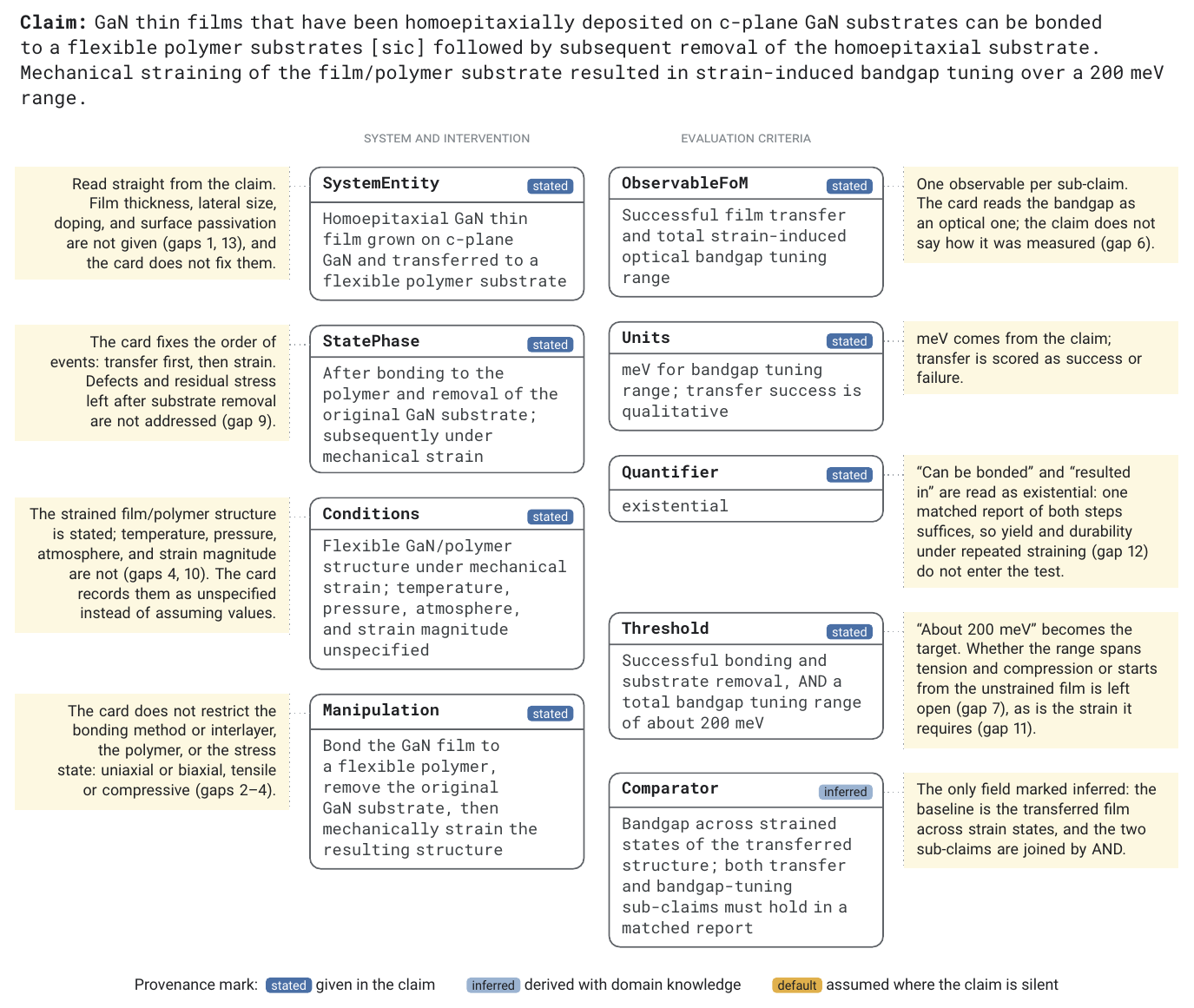}
\caption{The Claim Card that \system{} produced for the GaN claim, with
field values and provenance marks as output by the parser. The margin notes
are ours; gap numbers refer to the list in Appendix~\ref{app:gan}.}
\label{fig:card661}
\end{figure*}

Figure~\ref{fig:card661} shows the Claim Card that \system{} produced
for this claim, with margin notes on which of these gaps each field fills
in or leaves open. The card marks eight of its nine fields stated. A mark
applies to a whole field, so it cannot show that a stated value leaves
details of that field open.

\subsection{The Quantifier and the Provenance Marks}

\paragraph{Why the quantifier is a field:} The quantifier records whether the
claim is \emph{existential} (``can,'' ``there exists,'' meaning that at
least one instance satisfies it), \emph{typical} (``generally,'' ``tends to''), or
\emph{universal} (``always,'' ``for all''). The three values call for different
evidence and different stopping rules. Under the verification guideline, one
condition-matched instance that clears the threshold suffices for an
existential claim, and failing to find one is not negative evidence. For a
typical or universal claim, the guideline treats one condition-matched
counterexample as decisive negative evidence, and a few positive instances
do not establish the claim. A verifier that silently treats ``can'' as ``always'' may
reject feasible claims, and one that narrows a universal claim without
saying so may accept false ones. Recording the quantifier before evidence is
gathered lets a user see either error on the card.
Section~\ref{sec:control} shows that editing the quantifier can change the
feasibility score in the expected direction.

\paragraph{Provenance marks:} The mark on each content field tells a user
whether its value is given in the claim (\emph{stated}), follows from the
text with domain knowledge (\emph{inferred}), or was assumed where the text
is silent (\emph{default}). We use the share of fields that are not stated
to estimate what a claim leaves open, although the marks themselves are a
model's judgment. Every field is required by the schema
(Appendix~\ref{app:schema}), so a field that the claim leaves open is
filled with an explicit default or marked as unspecified. A qualitative
claim therefore also gets a \field{Units} value, such as ``Assay-dependent;
no units specified'' on the autophagy card of
Table~\ref{tab:card-example}.

\paragraph{Relation to PICO:} For biomedical claims, \field{SystemEntity} and
\field{StatePhase} play the role of PICO's population, \field{Manipulation}
its intervention, \field{ObservableFoM} its outcome, and \field{Comparator}
its comparator \citep{nye-etal-2018-corpus}. The card adds the quantifier,
threshold, units, conditions, and provenance.

\subsection{How a Card Is Produced and Used}
\label{app:card-production}

\system{} produces a card in a
step we call the \emph{preflight analysis}. This step turns the claim into
an explicit interpretation before evidence is gathered. Where the claim is silent, the
agent or the parser (Section~\ref{sec:generation}) supplies a value from
domain knowledge, and the value is marked inferred or default. A card
therefore holds both what the claim says and what \system{} decided it
means.

\paragraph{The two stages:} The preflight analysis has two stages. In the \emph{interpretation} stage,
our verification guideline instructs the agent, before it uses any tool,
to record its interpretation of the card fields
(Section~\ref{sec:fields}), the defaults it assumes, and the evidence that
would decisively change the score. Where the claim leaves a gap, the
guideline tells the agent to ``prefer the weakest faithful interpretation
of a claim.'' It also tells the agent not to upgrade quantifiers or add
hidden constraints, to choose the weakest defensible threshold when the
claim gives none, and not to change a default or narrow the scope once
evidence gathering begins. A table in the guideline maps verbs to testable
criteria. In this table, ``enables'' requires only that the downstream
capability be achieved in at least one instance, not that it improve in
magnitude.
In practice the agent writes its interpretation as free text in its first
reasoning step. This is a short paragraph (median about 90 words) that
names the system and the evidence the agent will seek, and it can leave
other fields to the next stage.

In the \emph{structuring} stage, a second model, the \emph{parser}
(GPT-6 Sol), converts the first reasoning step into JSON conforming to the
card schema (Appendix~\ref{app:schema}). We separate the stages so that
every decision the agent states is made by the model that has the
verification guideline and the full problem context, and so that the agent
does not have to write its interpretation as JSON. The field audit
(Section~\ref{sec:faithful} and Appendix~\ref{app:faithful-more}) checked the values and marks of one card for
each of the 60 claims with a judge that sees the claim.

\paragraph{What the parser decides:} The parser's instruction
(Appendix~\ref{app:schema}) asks it only to convert its input, but the
schema requires a value for every field and a provenance mark for each
value. The first reasoning step seldom says which values the agent
assumed (Appendix~\ref{app:consistency-more}), so where the step leaves a
field implicit, the parser supplies the value itself, and it assigns every
mark. A mark is the parser's judgment
of where a value came from, not a statement by the agent. The
parser sees neither the claim nor the verification guideline, so a value
that the agent used but did not restate in its first step can be missing
from the card, as in the two flagged fields that
Appendix~\ref{app:faithful-more} traces to the parser.

\paragraph{When the card appears:} In card mode, the agent is asked to state its own interpretation.
\system{} extracts the card from the agent's first reasoning step as soon
as the agent has written that step, before the tool call that the step
selects is executed. The card is printed at that point and saved
with the report when the run ends (Figure~\ref{fig:overview}b). It
therefore exists before any evidence is gathered, after the first of about
three agent steps per run (Appendix~\ref{app:faithful-more}). The run waits for
the extraction, so the time the extraction takes is part of the run's
wall-clock time. We did not measure how many seconds into a run the card
appears. If extraction fails, the run continues and
the first step's free text stands in for the card. The parser returned a
card in all 300 card runs without computational libraries and in 118 of
the 120 card runs with the computational tools.

\paragraph{What each step receives:} The agent gathers evidence under its own written interpretation, which
stays in its context, and in card mode it is not shown the parsed card.
The quantifier it wrote selects the guideline's stopping rule, and the
threshold and conditions define what counts as matched evidence. The
parsed card is passed only to the steps that end a run. The scorer, which
assigns the feasibility score, receives the claim, the parsed card, the
agent's reasoning, and its final answer, and scores the claim \emph{as
interpreted}. The step that writes the report is
instructed to list the card as an evidence item next to the verbatim claim.
Both the scorer and this step receive the card in card mode and in frozen
mode, and neither receives one in runs without a card.

A card can also be supplied from outside. In \emph{frozen} mode, \system{}
injects a stored card into the agent's prompt, with an instruction to use it
exactly as given and not to change the claim's interpretation, quantifier,
threshold, or defaults, and passes the same card to the scorer and the
report step. The agent is told not to write a card of its own. We used
frozen mode to give all five runs of a claim the same interpretation and
to run cards edited in one field (Sections~\ref{sec:conditions}
and~\ref{sec:control}).

\subsection{Preflight Schema}
\label{app:schema}

In the structuring stage (Appendix~\ref{app:card-production}), the parser
receives the instruction and JSON Schema below as its system prompt and the
agent's first reasoning step as the user message. The listing gives the
general-science version, used for the biomedical claims, with the
instruction and every field description verbatim. It shows the schema's
structure compactly rather than as JSON. Every field is required, every
field other than \field{Provenance} is of type \texttt{string}, and
\field{Quantifier} is limited to the three values shown.
\field{Provenance} holds one mark for each of the nine content fields,
limited to \emph{stated}, \emph{inferred} and \emph{default}. Besides the
card, the schema has five guideline-check fields (\field{VerbMapping} to
\field{MinimalDecisiveDesign}), which we do not analyze in this paper.

\begin{lstlisting}[literate={≥}{{$\geq$}}1 {→}{{$\rightarrow$}}1 {·}{{$\cdot$}}1]
Convert the unstructured input containing
preflight information into a JSON object with
the following schema. Do not add any other
text before and after the JSON.

PreFlight
  ClaimCard
    Structured claim definition including
    system, state, conditions, manipulation,
    observable, quantifier, threshold,
    comparator, and the provenance of each
    field.
    SystemEntity
      The physical, computational, or
      conceptual system under study.
    StatePhase
      The operational or temporal state of the
      system (e.g., 'steady-state',
      'transient', 'pre-treatment',
      'post-intervention').
    Conditions
      The environmental or boundary conditions
      under which the claim is made, including
      any controlled or measured variables.
    Manipulation
      The intervention, treatment, or change
      applied to the system.
    ObservableFoM
      The measurable quantity or Figure of
      Merit the claim refers to.
    Units
      The units associated with the observable
      (e.g., SI units or domain-appropriate
      units).
    Quantifier  enum [existential, typical,
                      universal]
      Scope of the claim: 'existential'
      asserts existence ('can', 'may', 'there
      exists'); 'typical' asserts the typical
      case ('generally', 'typically', 'tends
      to'); 'universal' asserts every case
      ('will', 'always', 'for all'). Use the
      weakest reading that is faithful to the
      text.
    Threshold
      The predefined quantitative or
      qualitative threshold for significance
      (e.g., '≥10% improvement', 'beyond
      method uncertainty').
    Comparator
      The baseline or control the claim is
      compared against, and the AND/OR logic
      across sub-claims (e.g., 'vs. undoped
      baseline from the same route; single
      claim', 'A AND B must both hold'). Write
      'none' if the claim makes no comparison.
    Provenance  for each field above, enum
                [stated, inferred, default]
      For each ClaimCard field, where its
      value comes from: 'stated' (given in the
      claim text), 'inferred' (follows from
      the claim text with domain knowledge),
      or 'default' (the claim text is silent
      and a default was assumed).
  VerbMapping
    Explanation of how the main verb in the
    claim maps to a measurable or modeled
    quantity change (e.g., 'improves' →
    increase in FoM).
  MetricIntegrity
    Description of how the observable
    maintains measurement integrity,
    calibration, and traceability.
  RepresentationGate
    Statement of how the claim's
    representation aligns with the measured or
    modeled data, ensuring consistency and
    validity.
  UnitsAndBaseline
    Defines measurement units and the baseline
    condition or control used for comparison.
  MinimalDecisiveDesign
    Describes the minimal experimental or
    computational design sufficient to
    decisively test the claim (replicates,
    controls, statistics).
\end{lstlisting}

\noindent The materials-science version, used for all materials-science
claims, replaces the descriptions of six content fields with the
following and is otherwise the same.

\begin{lstlisting}[literate={≥}{{$\geq$}}1 {→}{{$\rightarrow$}}1 {·}{{$\cdot$}}1]
SystemEntity
  The physical, computational, or conceptual
  system under study (e.g., 'Lithium-ion
  battery cell', 'Catalyst A').
StatePhase
  The operational or temporal state or phase
  of the system (e.g., 'charging phase',
  'steady-state', 'post-treatment').
Conditions
  The environmental or boundary conditions,
  including temperature (T), pressure (P), and
  atmosphere or medium.
Manipulation
  The intervention, dosage, or compositional
  change applied to the system (e.g.,
  'addition of 2% additive', 'voltage step to
  4.2 V').
ObservableFoM
  The measurable quantity or Figure of Merit
  (FoM) the claim refers to (e.g., 'Coulombic
  efficiency', 'reaction yield').
Units
  The units associated with the observable
  (e.g., '%', 'A·h/g', 'mV/dec').
\end{lstlisting}

\paragraph{Domain adaptation:} \system{} supports materials science, AI and
machine learning, and general science. It uses the same card schema in all
three domains and changes only the field descriptions shown to the parser
and the guideline texts shown to the agent and the scorer.
\field{StatePhase} is a phase or operating state for a material (e.g., bcc,
discharge), a lifecycle stage (training, inference) for a model, and the
state of an organism or sample in general science (aged versus young or
adult in Table~\ref{tab:card-example}). \field{Conditions} covers
temperature and atmosphere for a material, and the dataset and evaluation
protocol for a model.

\section{Experimental Setup in Detail}
\label{app:setup-more}

Table~\ref{tab:design} lists the runs of the study with their claims,
conditions and tool settings.

\begin{table*}[t]
\centering
\footnotesize
\setlength{\tabcolsep}{4pt}
\begin{tabularx}{\textwidth}{@{}>{\raggedright\arraybackslash}p{3.0cm} r >{\raggedright\arraybackslash}p{3.4cm} r >{\raggedright\arraybackslash}X@{}}
\toprule
Conditions & Claims & Setting & Runs & Frozen or edited card \\
\midrule
Card, no card, frozen & 60 & Without computational libraries$^{a}$ & 900 & Frozen: the card of each claim's first card run$^{\dagger}$ \\
\addlinespace
Edited cards & 56\,+\,58 & Without computational libraries$^{a}$ & 114 & The first card run's card with one field edited, the quantifier (56 claims) or the threshold (58 claims)$^{\dagger}$ \\
\addlinespace
Card, no card, frozen & 24 & With the computational tools & 360 & Frozen: a card from a separate preliminary card run, not one of the reported runs \\
\addlinespace
Frozen & 24 & Search only & 120 & The same frozen card \\
\addlinespace
Edited cards & 20\,+\,24 & With the computational tools and with search only & 88 & The frozen card with one field edited, the quantifier or the threshold, each run once in each setting \\
\addlinespace
Pilot (Appendix~\ref{app:pilot}) & 20 & Three LLM backbones, card and no card, one run each & 120 & -- \\
\bottomrule
\end{tabularx}
\caption{The runs in the study, by condition and setting. We ran 60 claims without computational libraries (30 materials science, 30 biomedical) and 24 materials-science claims with the computational tools and with search only, six of them among the 60. A frozen run injects a stored card into the agent's prompt. An edited card is a stored card with one field changed, run once in the same way. $^{a}$The CALPHAD and CHGNet tools could be called, but their libraries were not installed. Biomedical claims had a general coding agent instead (Appendix~\ref{app:tools}). $^{\dagger}$A search outage affected some of these runs (Section~\ref{sec:conditions}).}
\label{tab:design}
\end{table*}

\subsection{System and Tool-Use Policies}

\system{} iteratively decomposes a scientific claim and judges whether the
claim is feasible as stated. Before searching, it checks the quantities the
claim implies against hard physical bounds. The guidelines also have
rubrics for specific topics, such as phase stability and batteries for
materials-science claims, and controls and statistics for biomedical
claims.

The search agent returns cited source excerpts, and the coding agent writes
and runs code in its own workspace. A tool-use policy in the agent's
prompt, also chosen by domain, says how often each tool may be called. For
materials-science claims, the agent reaches the coding agent only through
the two computational tools of Section~\ref{sec:tools-setup}, and the
policy allows up to three searches and one call to each computational tool.
For biomedical claims, it allows up to five searches and one call to the
coding agent, which can be asked to write and run code for any check. That
call was made in 16 of the 450 biomedical runs (2 with a card, 4 without,
and 10 with a frozen card).

A separate step writes the report from the full trajectory, and a scorer
guided by the rubric assigns the verdict, a feasibility score from $-2$
(infeasible) to $+2$ (feasible), with $0$ for ``cannot determine.'' The
parser (Section~\ref{sec:generation}) runs on GPT-6 Sol with reasoning off
at temperature 0. Before the coding agent's output reaches the agent,
Gemini 3.8 Flash condenses it into a short summary of the method and the
result.

\paragraph{Model identifiers:} The agent, the scorer, the report step
and the figure captions used the vendor identifier \texttt{gpt-5.4} at
high reasoning effort, and the parser used \texttt{gpt-6-sol} with
reasoning effort \texttt{none} and temperature 0. The search agent ran on
\texttt{claude-sonnet-4-6} with web search, the coding agent on
\texttt{claude-sonnet-5-5} and the condenser on
\texttt{gemini-3.8-flash}. The consistency judge and the threshold edits
used \texttt{claude-opus-5-5}, with \texttt{gemini-3.1-pro-preview} as the fallback for the 60 claims, and the audit judge used \texttt{gemini-3.1-pro-preview}.
Appendix~\ref{app:prompts} gives the prompts.

\paragraph{Search restrictions and leakage:} The search tool was set to
block SciFact's code repository and raw files, its dataset pages, its file
store, its leaderboard and its demo. Pages hosting SFBench were not blocked, the block did not cover the coding agent,
which could look things up on the web itself, and the agent's prompt
contains the claim's problem id. We could not confirm that the runs
without computational libraries used the same block list, so for all runs we
rely on a scan of the stored records for pages hosting SciFact, SFBench or
SciFy and for problem ids. In those runs the scan found no such page
in 2{,}198 search requests, 4{,}831 search results, 129 coding-agent
questions, 129 coding-agent answers and 269 files in the coding agent's
working directories. A problem id appeared in 2 search requests of 1 run and
in 2 coding-agent questions. On the 24 claims it found no
such page in 915 search requests, 2{,}163 search results, 228
coding-agent questions, 228 coding-agent answers and 735 files, and no
problem id in a search request or coding-agent question. The stored
search results and coding-agent answers are condensed, so the scan cannot
rule out that a run saw such a page.

\subsection{Computational Tools and Run Settings}
\label{app:tools}

\paragraph{What the tools compute:} Besides phase equilibria, the CALPHAD
tool has the coding agent compute battery voltages, and the COST~507
database it uses covers light-metal alloys. Besides energies and forces,
the CHGNet tool has it compute stresses and magnetic moments.

\paragraph{Playbooks and their scores:} Each computational tool passes a
question from the agent to a coding agent together with a playbook, the
instruction file for one library. The
CALPHAD playbook restricts the coding agent to the thermodynamic databases
in a local directory and asks it to record which one it chose. It also asks
for a closing summary that includes ``a Likert label/score'' judged with
``default tolerances for T, composition, and phase fractions.'' The CHGNet
playbook gives a rubric from $+2$ (clear support) to $-2$ (clear
contradiction), tells the coding agent to ``define gray zones per
method,'' and has it take the starting structures for its simulations from
the Materials Project \citep{jain-etal-2013-materials}.
The coding agent's answer, with any score of its own, returns to the agent
as an observation. No card field sets the tolerances behind that score.
Appendix~\ref{app:tooluse-more} reports how often the tool's score agreed
with the run's final score.

\paragraph{Tool-call policy and the search-only setting:} The
materials-science tool-call policy limits how often the agent may call each
tool. It tells the agent to stop once a tool gives
conclusive evidence and notes that calling both computational tools is
usually unnecessary. Biomedical claims run under the general-science
policy, which offers a general coding agent in place of the two
computational tools. For the search-only runs, the CALPHAD and CHGNet tools
were removed from the agent's tool list, so the agent could not reach the
coding agent. The search-only runs let us compare the two tool settings
under the same frozen and edited cards (Section~\ref{sec:control} and
Appendix~\ref{app:tools-search}). They kept the same guidelines and
tool-call policy as runs with the computational tools, including the
policy's description of tools they could not call. Apart from the tool
setting, runs on the 60 claims and on the 24 used the same guidelines, tool-call policy
and card handling, and the same models for the agent, the scorer, the
parser, the search agent and the coding agent (Section~\ref{sec:setup}).

\paragraph{Calls without computational libraries:} The materials-science runs
on the 60 claims could call the CALPHAD and CHGNet tools, but the libraries those tools need were
not installed. The agent called one in 86 of the 450 materials-science runs (29
card, 31 no-card and 26 frozen runs) and made 90 calls, 51 to CALPHAD and
39 to CHGNet. Only one call returned a calculation, a Gibbs-energy hull
that the coding agent computed from the COST~507 database file without
pycalphad, in a no-card run on the Al--Zn single-phase claim. Of the 89
calls that returned no calculation, 81 reported that a required library was
missing, and in 43 the coding agent gave literature, recalled or
hand-estimated values instead. The general coding agent offered to biomedical claims was not
affected.

\paragraph{Days and computers for the 24 claims:} The frozen runs on these claims (with the tools and with search only) and
their quantifier-edited runs ran on one day, and their card, no-card and
threshold-edited runs on the next. Most card and no-card runs that called a computational tool ran
on a different computer from the frozen and edited runs (55 card and 47
no-card runs, compared with 7 card and 17 no-card runs on the computer of
the frozen runs).

On the computer used by most tool-calling card and no-card runs, the coding
agent sometimes loaded a thermodynamic database other than COST~507. Every
CALPHAD answer in the frozen and edited runs that named a database gave
COST~507, whereas 21 CALPHAD calls in the card and no-card runs (12 card and
9 no-card) used another Al--Fe or Al--Zn database, 20 of them on that
computer. On the three Fe--Al claims, 12 of 15 card-run CALPHAD calls and 7
of 14 no-card calls used such a database, against none of 15 frozen calls.
Many card and no-card calculations on those claims therefore rest on a
different thermodynamic database from the frozen ones.

\paragraph{Library versions:} The coding agent used pycalphad 0.11.2
(0.11.0 in one run) and CHGNet 0.3.8 or 0.4.2, the older CHGNet only on
the earlier day, with the v0.3.0 CHGNet weights in every run that records
them.

\subsection{Claims}

Three of the materials-science claims among the 60 include a figure, which
GPT-5.4 captioned, and the captions are stored in the run records of all
51 runs on these claims. The SciFact claims were run under the general-science
domain, in which the agent finds its own evidence without the SciFact
corpus. The six claims in both sets include the
opening claim. The gold labels of the 24 claims are more often positive than
those of the 13 tagged claims left out of them
(Section~\ref{sec:limitations}). The claims tagged for pycalphad were
distributed with the same COST~507 database file that the CALPHAD tool
uses. Appendix~\ref{app:ids} lists the claims with their labels and library
tags, and Appendix~\ref{app:gold} compares the experts' calculations,
recorded as code in the gold entries, with the agent's.

\subsection{Conditions and Edits}

Comparing runs with and without a card shows what asking for a card
changes. In no-card runs, we removed the guideline's subsections on the
card, verb mapping, defaults and envelopes, and comparators, and kept its
physical-plausibility checks (derived quantities, hard bounds, hazard
priors) and quantifier-based stopping rules. Neither the scorer nor the
report step sees a card. Frozen runs received their stored card in frozen
mode (Section~\ref{sec:visibility}). For the 24 claims, that card comes
from a preliminary card run made with the same settings as the five
reported card runs. That run is not one of the five, and its card is
identical to none of the 118 cards they produced. The six shared claims therefore have a different frozen card in each set. Each of these 24 frozen cards was run five times with the computational
tools and five times with search only. Every frozen run, in any setting,
carried its stored card unchanged.

The threshold edits were written by Claude Opus 5.5 with the same prompt
for all claims, with Gemini 3.1 Pro as a fallback for the 60 claims run
without computational libraries. The
prompt (Appendix~\ref{app:prompts}) asked for a threshold that is strictly
more demanding than the original and that a domain expert could plausibly
ask for, with a concrete quantitative threshold where the original was
qualitative and a larger required effect where it was quantitative. We did
not check the edits against that request.
Some edits also made a demand on the evidence itself, such as replication,
a statistical test, or a particular measurement method, and the keyword
rules that counted such demands differ between the 60 claims and the 24
(Appendix~\ref{app:propagation}). Two of the 60 claims have no threshold
edit, because their first-run cards were not yet available when the edits
were written. Each edited card of the 24 claims differs from its frozen
card only in the edited field, and each of their edited runs carried its edited card. The 20-claim pilot ran with and without a card on GPT-5.4 and two
other LLM backbones (Appendices~\ref{app:outcomes-more}
and~\ref{app:pilot}). Appendix~\ref{app:abstention} describes the search
outage, and Appendix~\ref{app:tools} the days and computers on which the
24 claims were run.

\subsection{Measures}

The field audit checks whether a card's values and their provenance marks
are faithful to the claim, and it covers the first-run card of each of
the 60 claims. For consistency, the second judge grouped the five runs'
values of each field into sets that express the same interpretation, and
consistency is the share of run pairs whose values fall in the same set,
averaged over the nine fields. For the quantifier we also report Fleiss'
$\kappa$ \citep{fleiss-1971-measuring}. An explicit card comes from the
first step alone, whereas the parser recovers a no-card run's
interpretation from its first step and final report, so it also recovered
interpretations from the card runs in the same way for a like-for-like
comparison. Gemini 3.1 Pro was available as a fallback for the consistency
grouping of the runs without computational libraries, and the records do
not show which of the two models grouped a given claim.

Run-to-run variance is also measured by sign disagreement, the share of
claims whose runs do not all have the same sign. Both sign measures treat
$0$ as a sign of its own. An abstention therefore counts as an error in
sign accuracy on any claim with a nonzero label, and as a different sign in
sign disagreement. Overhead compares card and no-card runs in GPT-5.4
tokens, wall-clock time, agent steps, and searches per run, and
Appendix~\ref{app:faithful-more} says how its cost was computed. Searches
per run are reported by domain, because the two tool-use policies allow
different numbers of searches.

For each edit we report the mean shift with its interval, the share of
claims whose edited score did not increase, the claims whose edited score
fell below every unedited run or rose above the unedited range, and the
edited runs that abstained. The change in absolute score shows whether an
edit made the verdict less decisive rather than more negative. It is the
edited run's absolute score minus the mean absolute score of the unedited
runs. Appendix~\ref{app:propagation} defines how we counted runs whose text
carries an edited value.

With the computational tools, we counted the runs that called a
computational tool, the calls that returned a calculation to the agent, and
the runs that received at least one calculation. A call returned a
calculation if the coding agent ran a calculation and the agent received
its result, in full or in part. In the runs without computational
libraries, a call counted as
returning a calculation if the coding agent wrote a result file other than
its script.

\subsection{Statistics}
\label{app:stats}

\paragraph{Intervals:} Every interval in the paper was computed by the same
script, which we will release with our code. It reports a 95\% percentile
interval from a bootstrap over claims with 10{,}000 resamples. For a paired
comparison it resamples the per-claim differences, so each claim
contributes its own contrast, and a ratio such as a relative overhead is
recomputed on each resample of claims. Because each comparison starts a
fresh random-number generator with the same seed, a value does not depend
on which other values were computed before it. We recomputed 13
intervals on the 60 claims and 9 on the 24, among them the edit shifts and the
card contrasts, with 20 other seeds. Their bounds moved by at most 0.01 for
the 60 claims and 0.02 for the 24, and none changed whether it
includes, reaches or excludes zero. Values and bounds are rounded half away
from zero to two decimals, and a bound that rounds to zero prints as 0.00.
When a bound prints as 0.00, we say the interval reaches zero if the
unrounded bound is zero and includes zero if it lies beyond zero, as for the
biomedical consistency difference in Appendix~\ref{app:consistency-more}.
Each consistency grouping was judged once, so the consistency intervals do
not include variation between judge calls. Tables give intervals in
brackets. The running text gives an interval only where a statement
depends on it, and then in a footnote.

\paragraph{Tests:} The sign test over claims drops ties and uses the
binomial distribution, and the sign-flip test of a mean per-claim
difference is exact, enumerating every sign pattern of the nonzero
differences. The edit nulls of Appendix~\ref{app:edits-more} treat a
claim's edited run and its five unedited runs as exchangeable and give
exact one-sided $p$-values, with counts over claims following a
Poisson-binomial distribution, and the one-in-six bound there uses a
binomial tail. The tests of which claims got a tool call permute condition labels within each
claim, either among all three conditions with the tools or among the two
compared, with a million permutations and the same seed. The group tests
below flip the signs of per-group mean differences. A random draw of 24 of
the 37 claims tagged for computational tools includes four or fewer of the
12 tagged claims labeled $-2$ with a hypergeometric probability of 0.008,
whereas our 24 claims include four (Section~\ref{sec:limitations}).

\paragraph{Claims that share a chemical system:} Several of the 24 claims
concern the same chemical system, such as the two Al--Zn claims or the
three Fe--Al claims, so their outcomes may not be independent. The script
also provides a cluster bootstrap that resamples groups of such claims
instead of single claims. We grouped the 24 claims by the chemical system
or host compound they concern, which gives 12 groups. Five claims, the
Al--Mg anode potential, silver bulk-modulus, VO$_2$ distortion,
La$_2$CuO$_4$ $c$-axis and P-in-Si interstitial claims, each form a group
of their own. The other seven groups are Al--Zn, Li--Al--Cu, doped
Fe$_2$O$_3$, Fe--Al, Al--Mg--Zn, doped Co$_3$O$_4$ and solutes in fcc Al.
Table~\ref{tab:cluster} repeats nine contrasts under this
cluster bootstrap. Each of its intervals includes, reaches or excludes
zero just as its counterpart from the bootstrap over claims does, and some
are narrower than their counterparts. With only 12 groups, five of them single claims, a percentile
cluster bootstrap can understate uncertainty
\citep{cameron-etal-2008-bootstrap}, so we also ran an exact sign-flip
test on the per-group mean differences, over every sign pattern of the
groups (11 groups for the quantifier edits, which cover 20 claims, and 12
otherwise). Its $p$-values are in the same table, and each test agrees
with the interval over claims on whether the contrast differs from zero.

\begin{table}[t]
\centering
\footnotesize
\setlength{\tabcolsep}{3pt}
\begin{tabular}{@{}lll@{}}
\toprule
Contrast & Bootstrap over groups & Sign-flip $p$ \\
\midrule
\multicolumn{3}{@{}l}{\emph{Quantifier edit, shift}} \\
\quad with the tools & $-0.55$ \ci{$-0.93$}{$-0.20$} & 0.039 \\
\quad search only & $-0.48$ \ci{$-0.68$}{$-0.28$} & 0.004 \\
\multicolumn{3}{@{}l}{\emph{Threshold edit, shift}} \\
\quad with the tools & $-0.48$ \ci{$-0.75$}{$-0.23$} & 0.016 \\
\quad search only & $-0.14$ \ci{$-0.42$}{$+0.15$} & 0.26 \\
\multicolumn{3}{@{}l}{\emph{Card $-$ no card}} \\
\quad sign accuracy & $+0.03$ \ci{$-0.04$}{$+0.09$} & 0.50 \\
\quad abstention & $-0.03$ \ci{$-0.08$}{$+0.04$} & 0.63 \\
\quad standard deviation & $+0.09$ \ci{$-0.03$}{$+0.19$} & 0.41 \\
\quad runs calling a tool & $-0.07$ \ci{$-0.16$}{$+0.03$} & 0.41 \\
\multicolumn{3}{@{}l}{\emph{Frozen, tools $-$ search only}} \\
\quad sign accuracy & $+0.12$ \ci{$+0.04$}{$+0.18$} & 0.016 \\
\bottomrule
\end{tabular}
\caption{Contrasts on the 24 materials-science claims, with the claims
grouped by chemical system.
The interval is a percentile interval from a bootstrap that resamples
groups of claims, and $p$ is from an exact sign-flip test over the
per-group mean differences. The edit shifts are defined as in
Table~\ref{tab:edits}, and the other rows are paired differences as in
Tables~\ref{tab:outcomes}, \ref{tab:tooluse} and~\ref{tab:tools-search-diff}.}
\label{tab:cluster}
\end{table}

\section{Selected Claims}
\label{app:ids}

So that the claims run without computational libraries can be
reconstructed from the public releases, Tables~\ref{tab:ids-sfbench}
and~\ref{tab:ids-scifact} list all 60, on which the repeated runs and the
card edits in that setting were made (Section~\ref{sec:data}). We identify the materials-science claims by their SFBench
problem id, prefixes included, and SciFact claims by their claim id in the development set. Each half was drawn at random
within strata of the gold label, and each label received a share of the 30
claims in proportion to its share of the claims it was drawn from. Six of
the materials-science claims were also run with the computational tools,
and Table~\ref{tab:ids-tools} marks them.

\begin{table}[H]
\centering
\footnotesize
\begin{tabular}{l r}
\toprule
Problem id & Label \\
\midrule
\texttt{matsci\_db\_002} & $+2$ \\
\texttt{matsci\_db\_003} & $+2$ \\
\texttt{matsci\_db\_006} & $-2$ \\
\texttt{mof\_alloys\_0007} & $+2$ \\
\texttt{mof\_batteries\_0010} & $-1$ \\
\texttt{vis\_reas\_alloys\_0002} & $+2$ \\
\texttt{vis\_reas\_batteries\_0002} & $-2$ \\
\texttt{vis\_reas\_superconductors\_0002} & $+2$ \\
\texttt{umbc\_alloys\_0003} & $+2$ \\
\texttt{umbc\_semiconductors\_0005} & $-2$ \\
\texttt{umbc\_alloys\_0005} & $+1$ \\
\texttt{jhu-alloys-0010} & $-2$ \\
\texttt{jhu-superconductors-0001} & $+1$ \\
\texttt{gtri\_0007} & $-2$ \\
\texttt{gtri\_0013} & $-2$ \\
\texttt{str\_dft\_06} & $-1$ \\
\texttt{str\_dft\_08} & $-2$ \\
\texttt{str\_spectulative\_01} & $-1$ \\
\texttt{str\_speculative\_04} & $+1$ \\
\texttt{str\_speculative\_11} & $+2$ \\
\texttt{str\_speculative\_12} & $+1$ \\
\texttt{str\_speculative\_16} & $-2$ \\
\texttt{computational\_tools\_0003} & $0$ \\
\texttt{computational\_tools\_0016} & $-1$ \\
\texttt{computational\_tools\_0017} & $+2$ \\
\texttt{computational\_tools\_0025} & $-2$ \\
\texttt{computational\_tools\_0028} & $+2$ \\
\texttt{computational\_tools\_0036} & $+2$ \\
\texttt{modalities\_0003} & $+2$ \\
\texttt{modalities\_0005} & $-2$ \\
\bottomrule
\end{tabular}
\caption{The 30 materials-science claims run without computational libraries, from SFBench \citep{costello-etal-2026-sfbench}. Each is listed by its SFBench problem id, with its gold feasibility label ($-2$ to $+2$). Table~\ref{tab:ids-tools} names the six \texttt{computational\_tools} claims. Of the others, the text refers to \texttt{matsci\_db\_003} as the MgO surface-energy claim, \texttt{str\_dft\_08} as the rock-salt cathode capacity claim and \texttt{str\_speculative\_11} as the toluene polysulfide claim.}
\label{tab:ids-sfbench}
\end{table}

\begin{table}[H]
\centering
\footnotesize\setlength{\tabcolsep}{3.5pt}
\begin{tabular}{l r l r l r}
\toprule
Claim id & Label & Claim id & Label & Claim id & Label \\
\midrule
56 & $+2$ & 532 & $-2$ & 1021 & $-2$ \\
118 & $+2$ & 539 & $+2$ & 1062 & $+2$ \\
148 & $+2$ & 540 & $+2$ & 1140 & $-2$ \\
239 & $+2$ & 589 & $+2$ & 1144 & $-2$ \\
248 & $+2$ & 636 & $+2$ & 1221 & $-2$ \\
249 & $-2$ & 770 & $+2$ & 1241 & $+2$ \\
275 & $+2$ & 781 & $-2$ & 1259 & $+2$ \\
327 & $+2$ & 805 & $+2$ & 1320 & $-2$ \\
399 & $+2$ & 847 & $-2$ & 1337 & $+2$ \\
443 & $+2$ & 1019 & $+2$ & 1370 & $-2$ \\
\bottomrule
\end{tabular}
\caption{The 30 SciFact claims \citep{wadden-etal-2020-fact}, by claim id in the SciFact development set, with the gold label mapped to our scale (\textsc{support} $=+2$, \textsc{contradict} $=-2$). Claim 148 is the autophagy claim of Section~\ref{sec:intro}.}
\label{tab:ids-scifact}
\end{table}

SFBench tags 37 claims for computational tools (19 for
pycalphad and 18 for CHGNet), and 24 of them were chosen at random, before any run, for the runs with the computational tools (Section~\ref{sec:data}).
Table~\ref{tab:ids-tools} lists them with their gold labels and the
library each is tagged for, eleven for pycalphad and thirteen for CHGNet.
It also gives the short name by which the text refers to each claim, so
that a name can be matched to its id, and the captions of
Tables~\ref{tab:ids-sfbench} and~\ref{tab:ids-scifact} do the same for
the claims among the 60 that the text names. All 24 claims keep the collection's
text and label. The other 13 tagged claims, which we did not use, have
labels that lean negative (one $+2$, two $+1$, one $0$, one $-1$ and eight
$-2$). The 24 we used have eleven $+2$, three $+1$, one $0$, five
$-1$ and four $-2$, so 14 of our 24 labels are positive.
Section~\ref{sec:limitations} discusses this skew.

\begin{table}[H]
\centering
\footnotesize
\setlength{\tabcolsep}{3pt}
\begin{tabular}{@{}l>{\raggedright\arraybackslash}p{2.85cm}rlc@{}}
\toprule
Id & Short name & Label & Library & \begin{tabular}[b]{@{}c@{}}Tool-\\calling\end{tabular} \\
\midrule
\texttt{0002} & Al--Zn eutectic & $+2$ & pycalphad & $\bullet$ \\
\texttt{0003}$^{\dagger}$ & Al--Zn single-phase (the opening claim) & $0$ & pycalphad & $\bullet$ \\
\texttt{0006} & Al--Mg anode potential & $-2$ & pycalphad &  \\
\texttt{0009} & Li--CuAl capacity & $-1$ & pycalphad & $\bullet$ \\
\texttt{0010} & Li--AlCu reaction & $+2$ & pycalphad & $\bullet$ \\
\texttt{0011} & silver bulk-modulus & $+2$ & CHGNet &  \\
\texttt{0013} & VO$_2$ distortion & $-2$ & CHGNet &  \\
\texttt{0016}$^{\dagger}$ & La$_2$CuO$_4$ $c$-axis & $-1$ & CHGNet &  \\
\texttt{0017}$^{\dagger}$ & Al-doped Fe$_2$O$_3$ magnet & $+2$ & CHGNet &  \\
\texttt{0018} & P-in-Si interstitial & $+2$ & CHGNet &  \\
\texttt{0023} & Fe--Al Al$_2$Fe-onset & $+2$ & pycalphad & $\bullet$ \\
\texttt{0024} & Fe--Al equiatomic strengthening & $+1$ & pycalphad & $\bullet$ \\
\texttt{0025}$^{\dagger}$ & Fe--Al Al-rich strengthening & $-2$ & pycalphad & $\bullet$ \\
\texttt{0026} & Al--Mg--Zn two-phase solidification & $+2$ & pycalphad & $\bullet$ \\
\texttt{0027} & Al--Mg--Zn three-phase solidification & $-1$ & pycalphad & $\bullet$ \\
\texttt{0028}$^{\dagger}$ & Al--Mg--Zn $\tau$-phase & $+2$ & pycalphad & $\bullet$ \\
\texttt{0030} & Fe$_2$O$_3$ dopant-ranking & $+1$ & CHGNet &  \\
\texttt{0031} & Fe-doped Co$_3$O$_4$ & $+2$ & CHGNet & $\bullet$ \\
\texttt{0032} & Nd-doped Co$_3$O$_4$ & $-1$ & CHGNet &  \\
\texttt{0033} & Cu-in-Al stiffening & $+2$ & CHGNet &  \\
\texttt{0034} & Cu-in-Al softening & $-2$ & CHGNet &  \\
\texttt{0035} & Mg-in-Al stiffening & $+1$ & CHGNet &  \\
\texttt{0036}$^{\dagger}$ & lattice-parameter ranking & $+2$ & CHGNet &  \\
\texttt{0037} & Cu-in-Al lattice-parameter & $-1$ & CHGNet & $\bullet$ \\
\bottomrule
\end{tabular}
\caption{The 24 materials-science claims run with the computational tools, by problem id (\texttt{computational\_tools\_}\emph{id}), with the short name by which the text refers to the claim, the expert label and the library the claim is tagged for. A bullet marks the 12 tool-calling claims, the post hoc group of claims on which at least three of the five frozen runs with the computational tools called a tool (Appendix~\ref{app:tools-search}). $^{\dagger}$Also among the 60 claims run without computational libraries (Table~\ref{tab:ids-sfbench}).}
\label{tab:ids-tools}
\end{table}

\clearpage

\section{Unstated Fields and Run-to-Run Consistency}
\label{app:underspec-more}

\begin{figure}[t]
\centering
\includegraphics[width=\columnwidth]{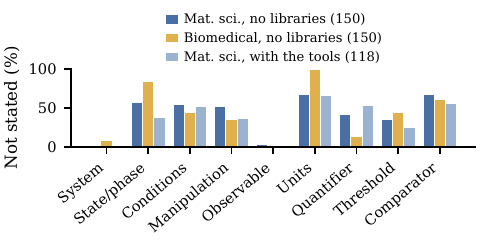}
\caption{Share of cards that do not mark a field stated, that is, mark it
inferred or default, in the 300 cards of the card condition without
computational libraries (150 per domain) and the 118 cards parsed in the
card runs with the computational tools.
Figure~\ref{fig:provenance} splits each bar into inferred and default.
Fields are labeled with short forms of the names in
Table~\ref{tab:card-example}.}
\label{fig:unstated}
\end{figure}

\begin{figure*}[t]
\centering
\includegraphics[width=0.9\textwidth]{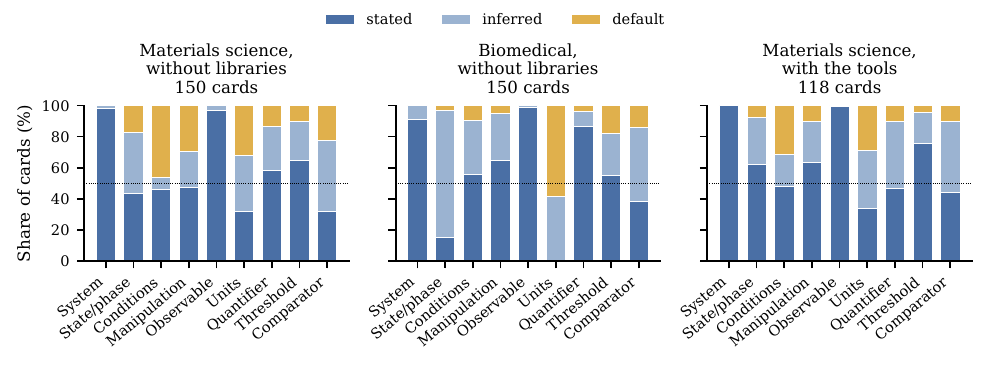}
\caption{Where each card field's value comes from. The left and middle
panels show the 300 cards from the card runs on the 60 claims (150
materials-science, 150 biomedical), and the right panel the 118 cards
parsed in the card runs on the 24 claims. The dark segment of each bar is the
share of cards that mark the field stated, and the segments above it the
shares that mark it inferred or default. The parser assigned the marks
(Section~\ref{sec:generation}). The dotted line is a 50\% reference.
Fields are labeled as in Figure~\ref{fig:unstated}.}
\label{fig:provenance}
\end{figure*}

Figure~\ref{fig:provenance} splits each bar of Figure~\ref{fig:unstated}
into the fields marked inferred and those marked default. In the card runs
on the 60 claims, the 42\% of materials-science field values that
were not stated divide into 23\% inferred and 19\% default, and the 44\% of
biomedical field values into 31\% inferred and 13\% default. Besides the
units, comparator and threshold of Section~\ref{sec:underspec}, the state
was not stated in 57\% of materials-science cards and 85\% of biomedical
cards, and the quantifier in 42\% and 13\%. Most biomedical cards recorded
the claim as typical (101 of 150), and most materials-science cards as
existential (110 of 150).

The 118 cards parsed in the card runs with the computational tools hold 1{,}062 field
values, of which 25\% were marked inferred and 11\% default. The units were
unstated in 66\% of these cards, the comparator in 56\%, the quantifier in
53\% and the conditions in 52\%, while the system was unstated in none and
the observable in 1\%. Of the 118 cards, 62 recorded the claim as
existential, 42 as typical and 14 as universal. Descriptively, on the six claims that appear in both sets, 62.5\% of field
values were marked stated in 29 cards from runs with the computational
tools and 65.2\% in 30 from runs without computational libraries.

\subsection{Consistency}
\label{app:consistency-more}

\begin{figure}[t]
\centering
\includegraphics[width=\columnwidth]{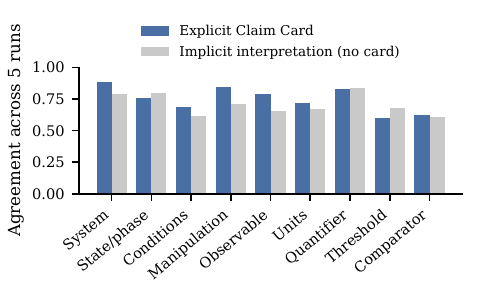}
\caption{Agreement across five runs, by field, for explicit Claim Cards and
for implicit interpretations recovered from runs without a card, in the
setting without computational libraries. A judge model grouped the explicit interpretations of 51 of
the 60 claims and the recovered ones of 56, and the paired comparison uses
the 50 claims in both. Fields are labeled with short forms of the names
in Table~\ref{tab:card-example}.}
\label{fig:consistency}
\end{figure}

\paragraph{What the first reasoning step says:} What Section~\ref{sec:consistency}
says about the first reasoning steps rests on keyword counts. Of the 300
first steps without a card, 295 contain a word of the system field of the
claim's frozen card, and all 300 name a source of evidence such as a
search, a study or a calculation. A looser rule, which matches words such
as ``decisive,'' ``enough'' or ``sufficient'' and conditionals such as
``would refute'' or ``if the,'' matches 274 of them, so it does not show
how many steps said what would decide the claim. The word ``default''
appears in none of the 300 and in 6 of the 300 first steps with a card, and
the stem ``assum'' in 1 first step without a card and 3 with one. These
counts are a lower bound, since a step can state an assumed value without
either word. In card mode the parser supplies the provenance marks, and
without a card nothing records them.

\paragraph{Coverage:} The judge grouped the explicit cards of 51 claims,
those whose card runs had finished when it ran, and the recovered no-card
interpretations of 56. The paired comparison of Section~\ref{sec:consistency}
uses the 50 claims in both, and the like-for-like comparison the 49 claims
whose recovered card and no-card interpretations were both grouped. The
nine claims without grouped explicit cards, four materials-science and
five biomedical, are claims on which the agent did better than on the 51.
In card runs they had an abstention rate of 0.02 against 0.16 and a sign
accuracy of 0.96 against 0.64, and in no-card runs 0.00 against 0.12 and
0.91 against 0.68. The standard deviation of their card-run scores was
0.39, against 0.31.

\paragraph{Like for like, by domain and by field:} Recovered from the
first reasoning step and final report in both conditions
(Appendix~\ref{app:setup-more}), interpretations agreed on 74\% of run
pairs with a card and 71\% without, on the 49 claims grouped both ways, a
difference of $+0.03$. Split by domain, the paired difference in
consistency between explicit cards and the interpretations recovered from
no-card runs was $+0.08$ on biomedical claims and $0.00$ on
materials-science claims. All three intervals include
zero.\footnote{The like-for-like difference
was $+0.03$ \ci{$-0.02$}{$+0.08$}, the biomedical difference $+0.08$
\ci{0.00}{$+0.17$} and the materials-science difference $0.00$
\ci{$-0.05$}{$+0.05$}.} On the claims whose card runs were grouped both
ways, explicit cards and the interpretations recovered from the same runs
(Appendix~\ref{app:setup-more}) agreed equally often (75\% each). Without
a card, the recovered interpretations agreed least on the comparator
(60\%) and conditions (61\%; Figure~\ref{fig:consistency}). For the
quantifier of the explicit cards, Fleiss' $\kappa$ was 0.70.

\paragraph{With the computational tools:} Extraction failed in two of the
120 card runs with the computational tools, and the text of the first step
stood in for the card. On the 22 claims with five parsed cards, explicit cards agreed on 76\% of run pairs and recovered no-card
interpretations on 73\%, a difference of $+0.03$.
With interpretations recovered in the same way from both conditions, the
difference was $+0.02$. As in the runs without computational libraries, explicit cards and
recovered interpretations did not differ measurably in
consistency.\footnote{The two differences were $+0.03$
\ci{$-0.04$}{$+0.09$} and $+0.02$ \ci{$-0.03$}{$+0.08$}.} The quantifier was unanimous
for 10 of 22 claims ($\kappa$ of 0.52).

\begin{table*}[t]
\centering
\footnotesize
\setlength{\tabcolsep}{4pt}
\begin{tabular}{@{}lrrrlrrl@{}}
\toprule
 & & \multicolumn{3}{c}{Same sign} & \multicolumn{3}{c}{Absolute score difference} \\
\cmidrule(lr){3-5}\cmidrule(lr){6-8}
Field compared & Claims & Alike & Differ & Alike $-$ differ & Alike & Differ & Differ $-$ alike \\
\midrule
\multicolumn{8}{@{}l}{\emph{Without computational libraries (60 claims)}} \\
Threshold & 31 & 85\% & 83\% & $+0.03$ \ci{$-0.03$}{$+0.09$} & 0.40 & 0.43 & $+0.02$ \ci{$-0.09$}{$+0.17$} \\
Quantifier's value & 20 & 83\% & 84\% & $-0.01$ \ci{$-0.08$}{$+0.07$} & 0.48 & 0.52 & $+0.03$ \ci{$-0.11$}{$+0.19$} \\
Judge's quantifier groups & 17 & 82\% & 87\% & $-0.04$ \ci{$-0.11$}{$+0.01$} & 0.42 & 0.50 & $+0.08$ \ci{$-0.06$}{$+0.25$} \\
All nine fields & 30 & 88\% & 82\% & $+0.06$ \ci{$-0.03$}{$+0.16$} & 0.27 & 0.34 & $+0.07$ \ci{$-0.07$}{$+0.21$} \\
\multicolumn{8}{@{}l}{\emph{Without computational libraries, card runs whose answer stated a score}} \\
Threshold & 14 & 82\% & 77\% & $+0.05$ \ci{$-0.06$}{$+0.21$} & \multicolumn{3}{c}{--} \\
Quantifier's value & 8 & 88\% & 79\% & $+0.08$ \ci{$-0.13$}{$+0.38$} & \multicolumn{3}{c}{--} \\
All nine fields & 13 & 82\% & 75\% & $+0.07$ \ci{$-0.07$}{$+0.23$} & \multicolumn{3}{c}{--} \\
\multicolumn{8}{@{}l}{\emph{With the computational tools (24 claims)}} \\
Threshold & 15 & 79\% & 86\% & $-0.07$ \ci{$-0.18$}{$+0.02$} & 0.49 & 0.54 & $+0.05$ \ci{$-0.18$}{$+0.31$} \\
Quantifier's value & 13 & 78\% & 85\% & $-0.08$ \ci{$-0.13$}{$-0.03$} & 0.60 & 0.42 & $-0.19$ \ci{$-0.33$}{$-0.06$} \\
Judge's quantifier groups & 12 & 76\% & 84\% & $-0.08$ \ci{$-0.14$}{$-0.03$} & 0.65 & 0.45 & $-0.20$ \ci{$-0.35$}{$-0.07$} \\
All nine fields & 15 & 86\% & 83\% & $+0.02$ \ci{$-0.09$}{$+0.14$} & 0.40 & 0.49 & $+0.09$ \ci{$-0.26$}{$+0.44$} \\
\bottomrule
\end{tabular}
\caption{Whether a claim's card runs that recorded the same
interpretation also reached the same verdict. For each field, pairs of
card runs whose values fell in the same consistency group (\emph{alike})
are compared with pairs whose values fell in different groups
(\emph{differ}), on the claims that have both kinds of pair. \emph{Same
sign} is the share of pairs whose two scores have the same sign, and
\emph{absolute score difference} is the mean absolute difference between
the two scores. Shares and means are averaged within each claim and then
over claims, and both differences are paired by claim and oriented so that
a positive value is what tracking would predict. \emph{Quantifier's value}
compares the quantifier written on the card, and \emph{judge's quantifier
groups} the consistency judge's grouping of it. Under \emph{all nine
fields}, alike pairs agree in every field and the other pairs differ in at
least one. The rows for runs whose answer stated a score use that score,
which leaves the scorer out, and give sign agreement only. Brackets give
bootstrap intervals over claims.}
\label{tab:interp-verdict}
\end{table*}

\paragraph{Interpretation and verdict within a claim:} To see whether the
variation in interpretation went together with variation in verdicts, we
compared, within each claim, pairs of card runs whose values of a field
fell in the same consistency group with pairs whose values fell in
different groups (Table~\ref{tab:interp-verdict}). All of these comparisons are exploratory, and we did
not adjust them for multiple comparisons. They are also observational,
and the scorer receives each run's card, so an association could also run
through the scorer. Without computational libraries, pairs whose threshold agreed had
the same sign in 85\% of cases and pairs whose threshold differed in 83\%,
over the 31 claims with both
kinds of pair. For the quantifier's value the shares were 83\% and 84\%
(20 claims), for the judge's quantifier
groups 82\% and 87\% (17 claims), and for
pairs alike in all nine fields against pairs that differed in any field
88\% and 82\% (30 claims). Measured by the
absolute difference in score, pairs that differed in the threshold were
$+0.02$ points further apart than pairs that agreed,
pairs that differed in the quantifier's value $+0.03$
and pairs that differed in any field $+0.07$.
Restricted to the card runs
whose own answer stated a score, which leaves the scorer out, the threshold
difference in sign agreement was $+0.05$ over 14
claims, the quantifier difference $+0.08$ over 8 and
the all-field difference $+0.07$ over 13. On the 20 claims whose five cards
did not share one quantifier, the score of runs with the stricter
quantifier minus that of runs with the weaker one was $-0.16$
on average, negative on 8 claims, zero on 9 and
positive on 3. Across claims, consistency and the standard
deviation of a claim's card-run scores had a Spearman correlation of
$-0.12$ (51 claims). On the 60 claims, none of these comparisons has an
interval that excludes zero.\footnote{The stricter-minus-weaker difference was
$-0.16$ \ci{$-0.43$}{$+0.13$} and the Spearman correlation $-0.12$
\ci{$-0.39$}{$+0.17$}. Table~\ref{tab:interp-verdict} gives the intervals
of the pair comparisons.}

With the computational tools, the threshold difference in sign agreement was $-0.07$
(15 claims), and pairs whose quantifier's value
agreed had the same sign less often than pairs whose value differed (78\%
against 85\%, 13 claims) and lay further apart in score. The stricter
minus weaker difference in score was $-0.25$ there
(negative on 4, zero on 8 and positive on 1 of 13 claims), and the
Spearman correlation was $-0.30$ (22 claims). The intervals of the
quantifier comparisons and of the stricter-minus-weaker difference exclude
zero, and those of the threshold difference and the correlation do
not.\footnote{With the tools, the stricter-minus-weaker difference was
$-0.25$ \ci{$-0.53$}{$-0.02$} and the Spearman correlation $-0.30$
\ci{$-0.70$}{$+0.16$}.}
Within a claim, the variation in interpretation that we recorded did not
consistently track the variation in verdicts. In the runs with the tools,
a stricter quantifier went with a lower score, but pairs whose quantifier
agreed also had the same sign less often, the opposite of what tracking
would predict.

\section{Card Edits in Detail}
\label{app:edits-more}

\begin{figure*}[t]
\centering
\includegraphics[width=0.9\textwidth]{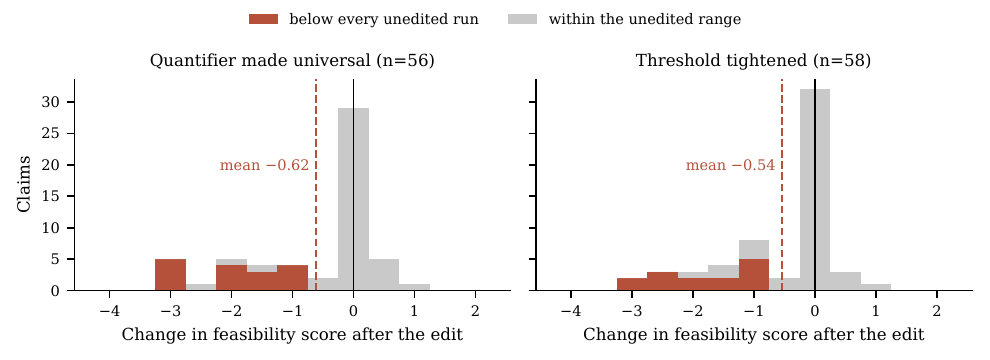}
\caption{Change in the feasibility score after editing the card, relative to
the mean of five runs with the unedited card, on the 60 claims. Dark bars
are claims whose edited score is below all five unedited runs; light bars
are all other claims, including the one claim whose threshold-edited run
scored above the unedited range.}
\label{fig:edits}
\end{figure*}

Table~\ref{tab:edits} gives every edit shift with its abstention counts and
robustness rows, and Figure~\ref{fig:edits} shows, claim by claim, the
shifts without computational libraries.

\begin{table*}[t]
\centering
\footnotesize
\setlength{\tabcolsep}{2.15pt}
\begin{tabular}{@{}lrlrcrrll@{}}
\toprule
 & & & Did not & Below\,/ & \multicolumn{3}{c}{Runs scoring 0} & $|$Score$|$ \\
\cmidrule(lr){6-8}
Edit and claims & $n$ & Shift & increase & above & Ed. & Un. & Diff. & change \\
\midrule
\multicolumn{9}{@{}l}{\emph{Without computational libraries (60 claims)}} \\
Quantifier, all claims & 56 & $-0.62$ \ci{$-0.91$}{$-0.35$} & 84\% & 16\,/\,0 & 11 & 54 & 0.00 \ci{$-0.09$}{$+0.10$} & $-0.18$ \ci{$-0.35$}{$-0.03$} \\
\quad materials science & 28 & $-0.44$ \ci{$-0.77$}{$-0.13$} & 82\% & 8\,/\,0 & 9 & 54 & $-0.06$ \ci{$-0.22$}{$+0.11$} & $-0.01$ \ci{$-0.21$}{$+0.19$} \\
\quad biomedical & 28 & $-0.80$ \ci{$-1.26$}{$-0.37$} & 86\% & 8\,/\,0 & 2 & 0 & $+0.07$ \ci{0.00}{$+0.18$} & $-0.36$ \ci{$-0.60$}{$-0.14$} \\
\quad no outage-hit runs & 54 & $-0.61$ \ci{$-0.91$}{$-0.34$} & 83\% & 15\,/\,0 & -- & -- & \multicolumn{1}{c}{--} & \multicolumn{1}{c}{--} \\
\addlinespace
Threshold, all claims & 58 & $-0.54$ \ci{$-0.79$}{$-0.32$} & 86\% & 14\,/\,1 & 20 & 49 & $+0.18$ \ci{$+0.09$}{$+0.28$} & $-0.39$ \ci{$-0.55$}{$-0.25$} \\
\quad materials science & 29 & $-0.43$ \ci{$-0.68$}{$-0.18$} & 90\% & 5\,/\,1 & 19 & 49 & $+0.32$ \ci{$+0.15$}{$+0.48$} & $-0.47$ \ci{$-0.71$}{$-0.24$} \\
\quad biomedical & 29 & $-0.66$ \ci{$-1.08$}{$-0.28$} & 83\% & 9\,/\,0 & 1 & 0 & $+0.03$ \ci{0.00}{$+0.10$} & $-0.32$ \ci{$-0.51$}{$-0.14$} \\
\quad no outage-hit runs & 50 & $-0.57$ \ci{$-0.82$}{$-0.33$} & 90\% & 13\,/\,0 & -- & -- & \multicolumn{1}{c}{--} & \multicolumn{1}{c}{--} \\
\midrule
\multicolumn{9}{@{}l}{\emph{With the computational tools (24 claims)}} \\
Quantifier & 20 & $-0.55$ \ci{$-1.04$}{$-0.10$} & 16 & 4\,/\,1 & 3 & 14 & $+0.01$ \ci{$-0.14$}{$+0.15$} & $-0.23$ \ci{$-0.51$}{$+0.02$} \\
\quad without one claim & 19 & $-0.47$ \ci{$-0.97$}{$-0.03$} & -- & 3\,/\,-- & -- & -- & \multicolumn{1}{c}{--} & \multicolumn{1}{c}{--} \\
Threshold & 24 & $-0.48$ \ci{$-0.80$}{$-0.20$} & 23 & 2\,/\,0 & 5 & 14 & $+0.09$ \ci{$-0.01$}{$+0.22$} & $-0.29$ \ci{$-0.52$}{$-0.11$} \\
\quad without two claims & 22 & $-0.29$ \ci{$-0.49$}{$-0.12$} & -- & -- & -- & -- & \multicolumn{1}{c}{--} & \multicolumn{1}{c}{--} \\
\midrule
\multicolumn{9}{@{}l}{\emph{Search only (24 claims)}} \\
Quantifier & 20 & $-0.48$ \ci{$-0.80$}{$-0.20$} & 19 & 4\,/\,0 & 5 & 27 & $-0.02$ \ci{$-0.18$}{$+0.15$} & $-0.04$ \ci{$-0.20$}{$+0.11$} \\
Threshold & 24 & $-0.14$ \ci{$-0.43$}{$+0.13$} & 18 & 3\,/\,2 & 12 & 28 & $+0.27$ \ci{$+0.09$}{$+0.45$} & $-0.38$ \ci{$-0.62$}{$-0.16$} \\
\quad without one claim & 23 & $-0.19$ \ci{$-0.47$}{$+0.07$} & -- & -- & -- & -- & $+0.23$ \ci{$+0.06$}{$+0.42$} & \multicolumn{1}{c}{--} \\
\bottomrule
\end{tabular}
\caption{Score shifts when one card field is edited to make the claim harder to satisfy. The quantifier edit sets the quantifier to universal, and the threshold edit makes the threshold stricter. Each edited card was run once with the card frozen, and its score is compared with the five unedited frozen runs of the same claim in the same setting. \emph{Shift} is the edited score minus the mean unedited score, averaged over the $n$ claims. \emph{Did not increase} counts claims whose edited score was at or below that mean (as a percentage of claims in the rows without computational libraries). \emph{Below} and \emph{above} count claims whose edited score fell below every unedited run or rose above the unedited range. \emph{Runs scoring 0} counts the edited runs that abstained (Ed.), out of $n$, and the unedited runs that abstained (Un.), out of five per claim, and \emph{Diff.}\ is the paired difference in the share abstaining. \emph{$|$Score$|$ change} is the change in the absolute score, so a negative value means that scores moved toward 0. Rows marked \emph{no outage-hit runs} leave out the edited runs that lost searches to the search outage (Section~\ref{sec:conditions}). Rows marked \emph{without} leave out claims whose edited run in that setting was irregular. With the computational tools, both such rows leave out the Li--AlCu reaction claim, whose quantifier-edited run lost its CALPHAD result and whose threshold-edited run replaced an attempt that had lost one, and the threshold row also leaves out the Al--Mg--Zn $\tau$-phase claim, whose threshold-edited run read a mole fraction as a percentage. With search only, the row leaves out the Fe--Al equiatomic strengthening claim, whose threshold-edited run found only empty searches. Brackets give bootstrap intervals over claims, and dashes mark values not computed.}
\label{tab:edits}
\end{table*}

\paragraph{Without computational libraries:} Under the quantifier edit, the score did not
increase for 84\% of claims and fell below every unedited run for 29\%.
Under the threshold edit, it did not increase for 86\% and fell below every
unedited run for 24\%. The edited score rose above the unedited range for
no claim under the quantifier edit and for 2\% of claims under the
threshold edit.

\paragraph{A null reference:} If an edit had no effect, the edited run
would be like a sixth unedited run of its claim. Given a claim's six
scores, each run would then be equally likely to be the edited one, so the
edited run would fall below every unedited run with probability one in six
when the lowest score is unique and never otherwise. Summed over claims,
this permutation null expects 3.7 claims below every unedited run under
the quantifier edit and 3.5 under the threshold edit, against the 16 and
14 observed ($p < 0.001$ for each), and a score that did not increase on
63\% and 65\% of claims, against the 84\% and 86\% observed ($p <
0.001$). It expects 1.0 and 1.2 claims above the unedited range, against
the 0 and 1 observed, so that count carries no information. The
permutation $p$-values of both mean shifts are below 0.001. Ignoring ties,
the chance of falling below every other run is one in six, which 16 of 56
and 14 of 58 exceed with binomial tail probabilities of 0.018 and 0.092,
but this bound is conservative because ties make the true rate much
lower. A like-for-like comparison, in which each unedited run in turn played the
edited run against the other four, gave a rate of 4\% below all four,
against 29\% and 26\% for the edited runs compared with four unedited runs
at a time, and both paired differences have intervals that exclude
zero.\footnote{The differences were $+0.26$ \ci{$+0.14$}{$+0.38$} for the
quantifier edit and $+0.22$ \ci{$+0.11$}{$+0.33$} for the threshold
edit.}

\paragraph{Sign changes:} The edited run's sign differed from the most
common sign of its claim's five unedited runs, with every tied sign
counted as most common, for 19 of 56 claims under the quantifier edit and
19 of 58 under the threshold edit. The permutation null of the previous
paragraph expects 7.5 and 8.3 such claims ($p < 0.001$ for each). All of
the quantifier changes (19 of 19) and 17 of 19 threshold changes moved to
a lower sign. Under the quantifier edit, 11 of 19 changes left a gold sign
that most unedited runs had taken, and 5 of 19 reached a gold sign that
most had missed. Under the threshold edit, these counts were 13 of 19 and
2 of 19. The gold label refers to the claim as written, not to the edited
interpretation. The sign changed on 10 of 28 materials-science and 9 of
28 biomedical claims under the quantifier edit, and on 12 of 29 and 7 of
29 under the threshold edit. On the 24 claims, it changed on 6 of 20
under the quantifier edit and 4 of 24 under the threshold edit with
the computational tools (2.2 and 2.0 expected, $p = 0.006$ and 0.11), and
on 6 of 20 and 7 of 24 with search only (3.5 and 3.3 expected, $p =
0.055$ and 0.022).

\begin{table*}[t]
\centering
\footnotesize
\setlength{\tabcolsep}{4pt}
\begin{tabular}{@{}lrlrl@{}}
\toprule
Claims & $n$ & Quantifier edit & $n$ & Threshold edit \\
\midrule
\multicolumn{5}{@{}l}{\emph{Without computational libraries (60 claims)}} \\
All claims & 56 & $-0.62$ \ci{$-0.91$}{$-0.35$} & 58 & $-0.54$ \ci{$-0.79$}{$-0.32$} \\
Without the claim whose unedited runs all scored $-2$ & 55 & $-0.63$ \ci{$-0.92$}{$-0.35$} & 57 & $-0.55$ \ci{$-0.81$}{$-0.33$} \\
Unedited mean above 0 & 33 & $-1.01$ \ci{$-1.43$}{$-0.60$} & 34 & $-0.99$ \ci{$-1.32$}{$-0.66$} \\
Unedited mean 0 or below & 23 & $-0.05$ \ci{$-0.21$}{$+0.09$} & 24 & $+0.08$ \ci{$-0.01$}{$+0.20$} \\
Card existential before the edit & 30 & $-0.65$ \ci{$-1.01$}{$-0.31$} & & \multicolumn{1}{c}{--} \\
Card typical before the edit & 26 & $-0.58$ \ci{$-1.05$}{$-0.17$} & & \multicolumn{1}{c}{--} \\
Edit adds no evidence standard$^{a}$ & & \multicolumn{1}{c}{--} & 28 & $-0.69$ \ci{$-1.07$}{$-0.33$} \\
Edit adds an evidence standard$^{a}$ & & \multicolumn{1}{c}{--} & 30 & $-0.41$ \ci{$-0.71$}{$-0.14$} \\
Universal marker in the first step$^{b}$ & 34 & $-0.99$ \ci{$-1.39$}{$-0.59$} & & \multicolumn{1}{c}{--} \\
No universal marker in the first step$^{b}$ & 22 & $-0.04$ \ci{$-0.25$}{$+0.14$} & & \multicolumn{1}{c}{--} \\
Edited run did not abstain$^{c}$ & 45 & $-0.63$ \ci{$-0.97$}{$-0.32$} & 38 & $-0.57$ \ci{$-0.90$}{$-0.27$} \\
Stated scores only$^{d}$ & 41 & $-0.67$ \ci{$-1.00$}{$-0.35$} & 45 & $-0.48$ \ci{$-0.78$}{$-0.22$} \\
\midrule
\multicolumn{5}{@{}l}{\emph{With the computational tools (24 claims)}} \\
All claims & 20 & $-0.55$ \ci{$-1.04$}{$-0.10$} & 24 & $-0.48$ \ci{$-0.80$}{$-0.20$} \\
Unedited mean above 0 & 11 & $-1.07$ \ci{$-1.73$}{$-0.42$} & 11 & $-0.89$ \ci{$-1.46$}{$-0.40$} \\
Unedited mean 0 or below & 9 & $+0.09$ \ci{$-0.24$}{$+0.42$} & 13 & $-0.12$ \ci{$-0.31$}{$+0.02$} \\
Edit adds no evidence standard$^{a}$ & & \multicolumn{1}{c}{--} & 8 & $-0.93$ \ci{$-1.70$}{$-0.28$} \\
Edit adds an evidence standard$^{a}$ & & \multicolumn{1}{c}{--} & 16 & $-0.25$ \ci{$-0.45$}{$-0.08$} \\
Edited run did not abstain$^{c}$ & 17 & $-0.51$ \ci{$-1.05$}{$-0.01$} & 19 & $-0.42$ \ci{$-0.80$}{$-0.13$} \\
Stated scores only$^{d}$ & 18 & $-0.48$ \ci{$-1.01$}{$-0.01$} & 21 & $-0.44$ \ci{$-0.79$}{$-0.16$} \\
\midrule
\multicolumn{5}{@{}l}{\emph{Search only (24 claims)}} \\
All claims & 20 & $-0.48$ \ci{$-0.80$}{$-0.20$} & 24 & $-0.14$ \ci{$-0.43$}{$+0.13$} \\
Unedited mean above 0 & 11 & $-0.76$ \ci{$-1.24$}{$-0.33$} & 11 & $-0.49$ \ci{$-0.93$}{$-0.07$} \\
Unedited mean 0 or below & 9 & $-0.13$ \ci{$-0.36$}{0.00} & 13 & $+0.15$ \ci{$-0.12$}{$+0.43$} \\
Edited run did not abstain$^{c}$ & 15 & $-0.55$ \ci{$-0.95$}{$-0.19$} & 12 & $-0.02$ \ci{$-0.25$}{$+0.20$} \\
Stated scores only$^{d}$ & 19 & $-0.48$ \ci{$-0.83$}{$-0.14$} & 23 & $-0.16$ \ci{$-0.46$}{$+0.12$} \\
\bottomrule
\end{tabular}
\caption{Edit shifts in subsets of claims. As in Table~\ref{tab:edits},
a shift is the edited score minus the mean of the claim's five unedited
frozen runs, averaged over the $n$ claims, and the rows for all claims
repeat that table. The unedited mean is the mean score of those five runs.
$^{a}$By the keyword rules of Appendix~\ref{app:propagation}, applied by
one coder, which differ between the 60 claims and the 24. $^{b}$Whether the
quantifier-edited run's first reasoning step contains the universal marker
(Appendix~\ref{app:propagation}). $^{c}$These rows select on what happened
after the edit and are descriptive. $^{d}$Computed from the agent's own
stated scores, on claims where the edited run and at least three unedited
runs stated one, a selection that the edit itself may influence. Brackets
give bootstrap intervals over claims, and dashes mark splits that concern
only the other edit.}
\label{tab:edit-splits}
\end{table*}

\paragraph{Where the shifts came from:} All five unedited runs scored $-2$
on one claim in each edit and $+2$ on 14. Without the claim at the floor,
the shifts were $-0.63$ for the quantifier edit and
$-0.55$ for the threshold edit (Table~\ref{tab:edit-splits}). The shifts came from
claims whose unedited runs scored positive on average, $-1.01$
over 33 claims and $-0.99$
over 34, whereas on claims with an unedited mean of 0 or below they were
$-0.05$ and $+0.08$, neither measurably different from zero. A
stricter interpretation thus lowered verdicts that were positive and left
the others about where they were. On the 24 claims, in both settings, the shifts on claims with a positive
unedited mean ranged
from $-1.07$ to $-0.49$ across the two edits, and those on the other
claims from $-0.13$ to $+0.15$. Without computational libraries, the
threshold shift was if anything larger on the edits
that add no evidence standard ($-0.69$, 28 claims) than on those that add
one ($-0.41$, 30 claims), and with the computational
tools likewise ($-0.93$ over 8 claims against
$-0.25$ over 16), so the threshold effect does not
rest on the added standards. Setting the quantifier to universal lowered
the score about as much on cards that had been existential, 30 of the 56,
as on cards that had been typical, 26 of the 56 ($-0.65$
and $-0.58$).

Two further splits condition on what happened after the edit and are
descriptive. On claims whose edited run did not abstain, the shifts were
$-0.63$ (45 claims) and $-0.57$
(38 claims). With search only,
the threshold shift on such claims was $-0.02$ (12
claims), so in that setting what lowering there was came from runs that
abstained. When the
quantifier-edited run's first reasoning step carried the universal marker
(Appendix~\ref{app:propagation}), the shift was $-0.99$
(34 claims), and otherwise $-0.04$
(22 claims).

\paragraph{The agent's own scores:} In 89 of the 114 edited runs without
computational libraries, the agent's own answer stated a score, and the scorer repeated it in
86 of them. Computed from the stated scores alone, on claims where the
edited run and at least three unedited runs stated one, a selection that
the edit itself may influence, the quantifier shift was $-0.67$
over 41 claims and the threshold shift $-0.48$
over 45, and both intervals exclude zero. On the 24 claims, 83 of the
88 edited runs stated a score and the scorer repeated it in 81, and the
shifts computed from stated scores excluded zero in every setting except
the threshold edit with search only ($-0.16$; Table~\ref{tab:edit-splits}).

\paragraph{A threshold edit:} The autophagy claim of
Section~\ref{sec:control} shows the effect of a quantifier edit, and the
MgO surface-energy claim, whose card the audit flagged
(Appendix~\ref{app:faithful-more}), shows that of a threshold edit. Four of
its five unedited runs scored $+2$ and the fifth $+1$. A threshold of
$\pm 3$\,meV/\AA$^2$ around the stated value gave a score of $0$, below
every unedited run. The agent found the literature values too scattered to
confirm a window that narrow. In both examples \system{} assessed the claim
as the edited card states it. A threshold edit like this one may test
whether the claim can be verified at all as much as how strict the
criterion is.

\paragraph{On the 24 claims:} With the computational tools, the
quantifier edit changed the score by $-0.55$ over 20
claims, and the threshold edit by $-0.48$ over 24 (Table~\ref{tab:edits}).
The score did not increase for 16 of 20 claims under the quantifier edit
and for 23 of 24 under the threshold edit. One quantifier-edited run scored
above the unedited range, on the Fe$_2$O$_3$ dopant-ranking claim, whose
five unedited runs all abstained. The irregular edited runs on two claims
are described with the edit propagation (Appendix~\ref{app:propagation}).
With search only, the quantifier edit changed the score by $-0.48$,
and the score did not increase for 19 of 20 claims.
The threshold edit's shift was $-0.14$, and its interval
includes zero. Under the permutation null, the quantifier edit put
more claims below every unedited run than expected, 4 against 0.8 with the
tools ($p = 0.003$) and 4 against 1.0 with search only ($p = 0.009$). The
threshold edit's 2 such claims with the tools and 3 with search only, against
0.8 and 1.0 expected, were not clearly more ($p = 0.20$ and 0.062), so for
that edit the shift intervals are the evidence. The permutation $p$-values
of the shifts were 0.002 and below 0.001 for the quantifier edit and below
0.001 and 0.17 for the threshold edit, whose edited runs were made a day
after the unedited ones. Two threshold-edited runs scored above the unedited range,
on the Li--CuAl capacity and Fe--Al equiatomic strengthening claims. On
both, all five unedited runs scored $-1$ and the edited run abstained.

\subsection{A Stricter Threshold That Search Cannot Meet Can Lead to Abstention}
\label{app:edit-abstention}

Section~\ref{sec:control} reports that a stricter threshold often left the
agent without evidence that met it on materials-science claims. Without the runs affected by the search outage, 15 of 24 threshold-edited
materials-science runs on the 60 claims and 45 of 120
unedited runs scored 0, a difference in the share abstaining of $+0.25$.
The difference was $+0.27$ with search only, or $+0.23$ without the
Fe--Al equiatomic strengthening claim, whose edited run found only empty
searches. With the computational tools, 5 of 24 edited runs abstained
against 14 of 120 unedited runs, a difference of $+0.09$
(Table~\ref{tab:edits}). Under
the stricter threshold, the search-only runs searched more, by $+0.28$
searches per run, and more of their searches came back
empty (22 of 60 against 71 of 267).\footnote{Excluding the outage-hit
runs, the difference in abstention without computational libraries was $+0.25$
\ci{$+0.08$}{$+0.43$}. The difference in searches per run with search only
was $+0.28$ \ci{$+0.02$}{$+0.55$}.} We think the agent abstained because it
could not find a value that met the stricter threshold. The quantifier edit did not raise
abstention (9 of 28 edited materials-science runs against 54 of 140
unedited), and only 1 of 29 threshold-edited biomedical runs abstained. The
threshold edit still made scores less decisive in every setting, lowering
the absolute score by between 0.29 and 0.47 points on average
(Table~\ref{tab:edits}). Abstention thus rose with search only, though
not measurably with the computational tools, and because abstention moves
negative scores up toward 0, it may explain why the interval of the
search-only threshold shift includes zero. Without the Fe--Al equiatomic
strengthening claim, the search-only threshold shift was $-0.19$,
and its interval still includes zero.

\subsection{Edit Propagation}
\label{app:propagation}

\begin{table*}[t]
\centering
\footnotesize
\setlength{\tabcolsep}{3pt}
\begin{tabular}{@{}>{\raggedright\arraybackslash}p{4.4cm}rrrrrrrrrrrr@{}}
\toprule
 & \multicolumn{4}{c}{\begin{tabular}{@{}c@{}}Without computational\\ libraries (60 claims)\end{tabular}} & \multicolumn{4}{c}{\begin{tabular}{@{}c@{}}With the computational\\ tools (24 claims)\end{tabular}} & \multicolumn{4}{c}{\begin{tabular}{@{}c@{}}Search only\\ (24 claims)\end{tabular}} \\
\cmidrule(lr){2-5}\cmidrule(lr){6-9}\cmidrule(lr){10-13}
 & \multicolumn{2}{c}{Quantifier} & \multicolumn{2}{c}{Threshold} & \multicolumn{2}{c}{Quantifier} & \multicolumn{2}{c}{Threshold} & \multicolumn{2}{c}{Quantifier} & \multicolumn{2}{c}{Threshold} \\
\cmidrule(lr){2-3}\cmidrule(lr){4-5}\cmidrule(lr){6-7}\cmidrule(lr){8-9}\cmidrule(lr){10-11}\cmidrule(lr){12-13}
Text that mentions the value & Ed. & Un. & Ed. & Un. & Ed. & Un. & Ed. & Un. & Ed. & Un. & Ed. & Un. \\
\midrule
Runs & 56 & 280 & 56 & 280 & 20 & 100 & 24 & 120 & 20 & 100 & 24 & 120 \\
\midrule
First reasoning step & 34 & 1 & 23 & 0 & 15 & 0 & 18 & 1 & 15 & 1 & 18 & 1 \\
Any reasoning step & 43 & 6 & 34 & 5 & 16 & 4 & 22 & 3 & 16 & 2 & 21 & 2 \\
Tool inputs$^{a}$ & 22 & 1 & 20 & 2 & 9 & 0 & 22 & 0 & 8 & 0 & 16 & 1 \\
\quad search queries alone & \multicolumn{2}{c}{--} & \multicolumn{2}{c}{--} & \multicolumn{2}{c}{--} & \multicolumn{2}{c}{--} & 0 & 0 & 1 & 0 \\
\quad calculation questions$^{b}$ & \multicolumn{2}{c}{--} & \multicolumn{2}{c}{--} & 4/11 & 0/53 & 14/15 & 0/64 & \multicolumn{2}{c}{--} & \multicolumn{2}{c}{--} \\
Final reasoning & 44 & 47 & 39 & 14 & \multicolumn{2}{c}{--} & \multicolumn{2}{c}{--} & \multicolumn{2}{c}{--} & \multicolumn{2}{c}{--} \\
Agent's answer$^{c}$ & 24 & 20 & 24 & 10 & 12 & 12 & 24 & 2 & 9 & 3 & 23 & 4 \\
Tool inputs asking for a counterexample & 18 & 10 & \multicolumn{2}{c}{--} & 8 & 3 & \multicolumn{2}{c}{--} & 8 & 1 & \multicolumn{2}{c}{--} \\
\midrule
Scorer rationale$^{d}$ & 32 & 26 & 19 & 3 & 12 & 7 & 23 & 5 & 14 & 3 & 21 & 1 \\
\bottomrule
\end{tabular}
\caption{Where the edited value appears, in the edited-card runs (Ed.) and in the five unedited frozen runs of the same claims (Un.). Each cell counts runs whose text mentions the value, and the first row gives the number of runs. The agent's own outputs come first and the scorer's rationale last. A quantifier edit is matched on \emph{universal} outside its CHGNet sense (``universal potential''). On the 60 claims, a threshold edit is matched on word trigrams of the edited threshold that carry a number absent from the claim and the unedited card, so those threshold counts cover only the edits with such a trigram. On the 24 claims, a threshold edit is matched on a pattern for the values it introduced, written per claim, including phrases that also occur without the edit (Appendix~\ref{app:propagation}). $^{a}$Questions passed to the computational tools, and search queries together with the agent-written context passed to the search agent; in runs with search only or without computational libraries, only the search queries with that context. $^{b}$Among runs that asked a calculation question. $^{c}$The agent's own final answer, not the composed assessment. $^{d}$The scorer is given the card, so its mentions show only that the edit reached the score, not that the agent used it. Dashes mark values not computed.}
\label{tab:propagation}
\end{table*}

Table~\ref{tab:propagation} counts, by setting, the edited and
unedited runs whose text carries the edited value, from the agent's first
reasoning step to its final answer and the scorer's rationale. These counts support the
argument of Section~\ref{sec:control} that an edit reached what the agent
itself wrote and not only the scorer.

\paragraph{The edits:} Of the frozen cards of the 24 claims, the four that were
already universal, and so received no quantifier edit, are those of the
Al--Zn single-phase, Li--CuAl capacity, VO$_2$ distortion and Cu-in-Al
softening claims.

Many threshold edits are compound. By keyword rules applied by one coder,
30 of the 58 edits on the 60 claims (19 materials-science and 11 biomedical) and 16 of
the 24 edits on the 24 claims add an evidence standard absent from the original
threshold, such as replication, a statistical criterion or a named
measurement method. The rule for the 24 claims also counts demands for matched
conditions and for confirmation or detection, so the two counts are not
strictly comparable. Some edits also constrain what another field
describes. On the 24 claims, the edit on the Al--Mg--Zn
$\tau$-phase claim caps the Zn
addition, which the \field{Manipulation} field specifies, and the one on
the Al--Zn single-phase claim requires the single phase to persist over a
composition window around the nominal composition.

\paragraph{Where the edited value appears:} For each edited run and the
five unedited frozen runs of the same claim, we searched the agent's first
reasoning step, all its reasoning steps, its tool inputs, its own final
answer and the scorer's rationale for the edited value. The final answer
is the agent's answer, not the assessment composed afterwards, which also
receives the card. Tool inputs are the questions sent to CALPHAD or CHGNet
and, for a search, the search request, meaning the query together with the
context the agent writes for the search agent. The scorer receives the
card, so its mentions show only that the edit reached the score, not that
the agent used it. With the tools, its rationale mentioned the edited
threshold in 23 of 24 edited runs and 5 of 120 unedited runs.

A marker is a word or text pattern that signals an edited value, and we
counted the value as present in any text that contains its marker. For
the universal quantifier, the marker is the word ``universal'' or ``universally,'' except in the sense of
a universal interatomic potential. For a threshold edit on the 24 claims,
it is a hand-written pattern, one per claim, that matches the values the
edit introduced. The patterns include phrases of the edited value that also
occur in some unedited runs, so the patterns alone do not make the unedited
counts low. Leaving those phrases out would change the edited counts by at most one
run. For the 60 claims, the threshold
marker is any word trigram of the edited threshold that contains a number
found in neither the claim nor the unedited card, and 56 of the 58 edits
have one.

In every setting, the edited value appeared far more often in
edited runs than in unedited ones, from the first reasoning step on
(Table~\ref{tab:propagation}). On the 56 threshold edits to the 60 claims
that introduced a number found in neither the claim nor the unedited card,
the edited threshold appeared in the agent's reasoning in 34 edited runs,
against 5 of 280 unedited runs. With the tools, 14 of the 15 edited runs
that asked a calculation question carried the edited threshold in it,
against none of 64 unedited runs (10 of 15 under a stricter rule that
requires three shared trigrams), and the threshold appeared in the agent's
reasoning in 22 of 24 edited runs, against 3 of 120 unedited runs. With
search only, it reached the search
requests of 16 of 24 edited runs, against 1 of 120 unedited runs, but the
query alone in only 1. The universal quantifier reached the tool inputs of 9 of
20 edited runs with the tools and 8 with search only, against none of 100
unedited runs, and 8 edited runs in each setting asked the tools or the
search agent for a counterexample, against 3 unedited runs with the tools
and 1 with search only, out of 100 each.
Without computational libraries, where runs could search but not calculate, the edited
threshold reached the search requests of 20 of 56 edited runs against 2 of
280 unedited ones, and the universal quantifier reached those of 22 edited
runs against 1 unedited run. After the quantifier edit on these claims,
the search requests asked for counterexamples in 18 of 56 edited runs against
10 of 280.
Unedited runs often use ``universal'' in contrast, as in ``typical rather
than universal,'' so the quantifier marker is less specific in the agent's
later text. On the 60 claims it appeared in the reasoning that
accompanies the final answer in 44 of 56 edited runs and 47 of 280 unedited
runs, and in the final answer itself in 24 edited and 20 unedited runs.

\paragraph{How a threshold edit changed the calculation:} With the tools,
a threshold edit changed what the agent asked the tools more than whether
it called them. Edited runs called a tool in 15 of 24 runs and unedited
runs in 64 of 120, a per-claim difference of $+0.09$ whose interval
includes zero (Table~\ref{tab:tooluse}). The quantifier edit did not change the call rate
measurably either (11 of 20 edited against 53 of 100 unedited runs, $+0.02$).
Threshold-edited runs with the tools searched
more, by $+0.43$ searches per run, as did those with
search only.\footnote{With the tools, the difference was $+0.43$
\ci{$+0.15$}{$+0.78$} searches per run.} On the
three claims whose edited threshold named a composition range, the edited
run calculated over that range, whereas the unedited runs had calculated
at the nominal composition or over a different range. The edited run on
the Al--Zn single-phase claim computed $x(\mathrm{Zn})$ of 0.45, 0.50 and
0.55, against 0.50 in all five unedited runs. On the Fe--Al
Al$_2$Fe-onset claim, the edited run scanned 45 to 60\,at.\% Al, a narrower window than
the unedited scans, which ran from 40 or 45 up to 75--85\,at.\%. For the
Cu-in-Al lattice-parameter claim, the edited run put 1 to 5 Cu atoms into
cells of 108 and 256 atoms (0.39 to 1.95\,at.\% Cu). Each of the three
unedited runs that called CHGNet put a single Cu atom into each cell (0.20
to 3.13\,at.\%). The edit on the Li--AlCu reaction claim instead changed
the criterion itself, from AlCu coexisting with a lithium-containing phase
to exactly two phases. The edited run, one of the irregular runs described
below, obtained the same three-phase field as all five unedited runs,
judged it against the new criterion and scored 0 where each of them had
scored $+2$.

\paragraph{Irregular runs:} Three edited runs with the tools were
irregular. On the Li--AlCu reaction claim, a CALPHAD result was lost under
both edits. The quantifier-edited run lost it and scored 0, against $+2$ in
all five unedited runs, and the threshold-edited run, which also scored 0,
replaced an attempt that had lost it (Appendix~\ref{app:tools-search}). On
the Al--Mg--Zn $\tau$-phase claim, the threshold-edited run took the
tool's $\tau$-phase mole fraction for a mole percentage, an error rather
than a different interpretation, and scored $-1$, against $+2$ in all five
unedited runs. Table~\ref{tab:edits} gives the
shifts without these claims.

\section{Same Calculation, Different Verdicts}
\label{app:samecalc}

With the computational tools, among runs of a claim with the same calculation
result, the share of pairs with the same score, averaged over claims, was
0.77 under the frozen card (12 claims) and 0.72 in card runs (9 claims).
Table~\ref{tab:samecalc} lists five claims on which the same calculation
did not settle the verdict. On four of them the
runs reached opposite verdicts, and on the opening claim they agreed with
one another but not with the expert. The cases illustrate the pattern
rather than estimate how often it occurs.

\begin{table*}[t]
\centering
\footnotesize
\setlength{\tabcolsep}{4pt}
\begin{tabularx}{\textwidth}{@{}>{\raggedright\arraybackslash}p{2.0cm}>{\raggedright\arraybackslash}X>{\raggedright\arraybackslash}X>{\raggedright\arraybackslash}p{2.9cm}c@{}}
\toprule
Contrast & Same calculation & What differed & Scores & Gold \\
\midrule
\multicolumn{5}{@{}l}{\emph{Al--Zn single-phase (the opening claim)}} \\
Agent vs expert, every condition with the tools & A single solid phase only above about 600\,K in 15 of 15 runs, as in the expert's calculation, and the full phase sequence in 11 of 15 & Every card-run card and the frozen card were universal, requiring one solid phase at every temperature in the claimed range, while the expert labeled the claim 0 because the temperature range is not specified & $-1$ in 15 of 15 runs & $0$ \\
\addlinespace
\multicolumn{5}{@{}l}{\emph{Li--AlCu reaction}} \\
Between conditions & The same three-phase field, $\eta$-AlCu + $\zeta$ + T1, at the onset of lithiation, in six card and no-card runs (all scoring $-1$) and in 5 of 5 frozen runs & Whether a two-phase reaction means two coexisting phases (card and no-card runs) or coexistence with a Li-containing phase rather than a single-phase solid solution (frozen card) & \emph{Card}\newline $+2$, $-1$, $+1$, $-1$, $-1$\newline \emph{No card}\newline $-1$, $-1$, $-1$, $+1$, $-1$\newline \emph{Frozen}\newline $+2$, $+2$, $+2$, $+2$, $+2$ & $+2$ \\
\addlinespace
\multicolumn{5}{@{}l}{\emph{Al--Mg--Zn $\tau$-phase}} \\
Within the card condition & The same $\tau$-phase fractions in 5 of 5 card runs, 0.12 at 2\,at.\% Zn and 0.22 at 3\,at.\% Zn with $x(\mathrm{Mg}) = 0.345$, at 298--300\,K & Not established for the runs that scored $+1$ and $+2$. All five cards were existential with the same threshold, and the first run's report interpreted the claim as typical although its card was existential & \emph{Card}\newline $-1$, $+1$, $+1$, $+2$, $+2$\newline \emph{Frozen}\newline $+2$, $+2$, $+2$, $+2$, $+2$ & $+2$ \\
\addlinespace
\multicolumn{5}{@{}l}{\emph{Fe--Al Al$_2$Fe-onset}} \\
Within one frozen card & The same B2 phase limit in 5 of 5 frozen runs & A decision rule set in the first reasoning step, that a single counterexample makes the claim negative (2 of 5 runs) & \emph{Frozen}\newline $+2$, $-1$, $+1$, $-1$, $-1$ & $+2$ \\
\addlinespace
\multicolumn{5}{@{}l}{\emph{Li--CuAl capacity}} \\
Within one frozen card & The same lithiation limit, $x = 0.38$ and $0.385$ in two frozen runs & How the tool's answer framed the same result, as a lithiation ceiling close to the claimed value or as no boundary at the claimed value & Frozen $+2$ and $-1$ & $-1$ \\
\bottomrule
\end{tabularx}
\caption{Runs with the computational tools that obtained the same calculation but reached different scores, on five of the 24 claims. The cases illustrate how verdicts can diverge and do not estimate how often this happens. \emph{Contrast} says whether the runs compared share a condition or one frozen card, come from different conditions, or are compared with the expert. The playbooks ask the coding agent for a score of its own, and some tool answers returned to the agent carry one (Appendix~\ref{app:tools}, Table~\ref{tab:tooluse}). Scores are listed in run order, and gold is the expert label. The cases were coded by a single coder and checked by a second reader.}
\label{tab:samecalc}
\end{table*}

Where the runs' interpretations differed, the verdicts
followed how a run interpreted the claim's temperature range or its phase
criterion. Under one frozen card, the verdicts depended instead on how the
agent applied the card or how the tool's answer framed the result. The Al--Mg--Zn
$\tau$-phase claim requires a $\tau$-phase fraction over 20\% after a few
atomic percent of Zn is added. All five card runs computed the same $\tau$
fractions near room temperature, 0.12 at 2\,at.\% Zn and 0.22 at
3\,at.\% Zn, on either side of the 20\% threshold, and all five cards
were existential with that threshold, yet the runs scored $-1$, $+1$,
$+1$, $+2$ and $+2$. The run that scored $-1$ preferred room temperature,
and its report interpreted the claim as typical although its card was
existential. What separated the runs that scored $+1$ from those that
scored $+2$ remains open. The Li--AlCu reaction claim, ``The initial reaction of Li with
AlCu is a two phase reaction,'' has a gold label of $+2$. The six card and
no-card runs whose calculation gave $\eta$, $\zeta$ and T1 at the onset of
lithiation all scored $-1$. The frozen card counted the reaction as
two-phase if AlCu coexisted with a Li-containing phase rather than forming a
single-phase solid solution, and all five frozen runs, which obtained the
same phase field as those runs, scored $+2$. Because this contrast is between conditions,
it may also reflect when and on which computer the runs were made and where
the frozen card came from (Section~\ref{sec:limitations} and
Appendix~\ref{app:tools}). The claim does not say which sense of
``two-phase reaction'' it intends.

Under one frozen card, the five runs on the Fe--Al Al$_2$Fe-onset claim
computed the same B2 limit and scored $+2$, $-1$, $+1$, $-1$ and $-1$, and
two of them had set a rule in their first step that a single counterexample
would rate the claim negative. On the Li--CuAl capacity claim, which puts
the maximum practical capacity at $x = 0.4$, one frozen run computed a
lithiation limit of $x = 0.38$ and scored $+2$, and another computed
$x = 0.385$ and scored $-1$. The tool answers framed these nearly equal
limits in opposite terms, one as a lithiation ceiling close to the claimed
value and the other as no boundary at the claimed value. The opposite framing
was already in the coding agent's own output, so the condensing step did
not create it, although the condensed answer of the run that scored $-1$
left out two of the coding agent's qualifications, that 0.385 is close to
0.4 and that the value is only approximately right. On the
opening claim, where the comparison is with the expert, all 15 runs with
the computational tools found a single solid phase only above about
600\,K, as the expert's diagram does, 11 of them with the full phase
sequence. All 15 scored $-1$ where the expert gave 0
(Figure~\ref{fig:overview} and Section~\ref{sec:conclusion}). Every card
required a single solid phase at every temperature below 700\,K, which
the guideline's preference for the weakest faithful interpretation would
not select if the existential interpretation is also faithful. We have not
established why the cards took this interpretation, or why the runs scored
$-1$ rather than $-2$.

\subsection{Agent and Expert Calculations}
\label{app:gold}

The gold entries of 35 of the 37 tagged claims record the expert's
calculation as code, including 22 of the 24 that we ran, so the agent's calculations can be compared with the
experts'. The other two of the 24 are the lattice-parameter claims, which have no code and
give only an explanation. We read the code as text and did not
run it. One coder compared each claim's agent calculations, over the card,
no-card and frozen runs with the tools, with the expert's. A claim was coded as a
match when the agent computed the same quantity with a compatible set-up
and obtained a value within the expert's stated precision. Otherwise it
was coded as the same quantity with a different value, a different set-up
(phase set, composition, cell or figure of merit), or no agent calculation.

The agent matched the expert's calculation on 11 of the 24 claims, computed
the same quantity with a different value on 1 (the Fe-doped Co$_3$O$_4$
claim), used a different set-up on 9, and calculated nothing on 3 (the
silver bulk-modulus, VO$_2$ distortion and Fe$_2$O$_3$ dopant-ranking
claims). Among the 12 tool-calling claims, 7 matched, 1 differed in value
and 4 differed in set-up. On the Al--Mg--Zn $\tau$-phase claim, the
$\tau$-phase fraction of 0.261 found by a frozen run
(Appendix~\ref{app:tools-search}) agrees with the expert's explanation,
which gives about 26\%. Runs on the Fe--Al Al-rich strengthening claim
that used COST~507 put the mole fraction of Al$_2$Fe at 65\,at.\% Al at
0.89--0.90, and the expert put it at 90\%. A frozen run on the P-in-Si
interstitial claim found a tetrahedral interstitial P atom in Si 2.06\,eV
higher in energy than a substitutional one, where the expert's calculation
gives 2.20\,eV. On the Al--Zn single-phase claim, 11 of 15 runs found fcc and
hcp phases below about 550\,K, two fcc phases from 550 to 597\,K and a
single fcc phase from about 597\,K. The expert's explanation gives the
boundaries as round values, 500 and 600\,K, and the agent's lower boundary
of about 550\,K agrees with the monotectoid temperature in the assessment
of \citet{murray-1983-al}. Yet we found no evidence that runs whose calculation matched
the expert's agreed more often with the sign of the gold label. The CALPHAD
claims were distributed with the same COST~507 file the
agent used (Appendix~\ref{app:setup-more}), so a CALPHAD match most likely means
that the agent set up the expert's calculation with the same database, not
that two independent thermodynamic descriptions agree.

Several differences trace to the expert's code rather than to the agent.
For the two Al--Mg--Zn solidification claims, the code swaps Mg and Zn
relative to the claim. The code for the Cu-in-Al stiffening, Cu-in-Al
softening and Mg-in-Al stiffening claims uses an 8-atom cell, which holds
12.5\,at.\% solute, and labels it 4\,at.\%. For the two Li--Al--Cu
claims, the phase list omits bcc and liquid, so lithium metal cannot form
in the expert's calculation. In the explanation for the Al--Zn eutectic
claim, the components of the eutectic composition sum to only 90\,at.\%.
We have not established whether any of these problems changes a gold
label. Without the nine claims whose expert code or explanation has such a
problem or computes a proxy for the claimed quantity
(Section~\ref{sec:limitations}), card minus no card in sign accuracy was
$-0.01$, and the gain of frozen runs with the
computational tools over search only was $+0.08$. Neither interval
excludes zero.\footnote{Card minus no card was $-0.01$
\ci{$-0.05$}{$+0.03$}, and the gain was $+0.08$ \ci{$-0.01$}{$+0.19$}.}
Always predicting a positive score would give 0.53 on the remaining 15
claims. Several CHGNet claims concern quantities that a general-purpose
machine-learned potential trained on density functional theory (DFT) data
may not settle, such as the saturation magnetization of a doped cobalt
oxide, the stiffening of Al by dissolved Cu or Mg or the distortion of
VO$_2$, and such potentials tend to underestimate the curvature of the
energy surface \citep{deng-etal-2025-systematic}. Where a gold entry
records a CHGNet calculation, its label may reflect what that calculation
found rather than what the material does.

A calculation did not by itself determine the verdict. On the opening
claim the agent's calculation matched the expert's, and the verdicts still
differed from the expert's label. On the Li--AlCu reaction claim, the
agent's calculations found the $\eta$\,+\,$\zeta$\,+\,T1 field at the
onset of lithiation, and those that traced the lithium chemical potential
found it flat across that field. In a ternary
system at constant temperature and pressure, the chemical potentials stay
constant as the composition changes only where three phases coexist,
because only there does the phase rule leave no degree of freedom. Card
and no-card runs that found this field scored the claim $-1$, and all five
frozen runs found the same field and scored $+2$ (Table~\ref{tab:samecalc}).

\section{Card Audit and Overhead}
\label{app:faithful-more}

The field audit labeled every field of the first-run cards of the 60
claims against its claim (Table~\ref{tab:audit}). A field judged faithful
here is faithful to the claim. Whether a card records the interpretation
that its own run went on to apply is a separate question, which we did not
measure systematically, and the paragraph on the opening claim below
describes one case in which the two differed.

\begin{table}[t]
\centering
\small
\setlength{\tabcolsep}{4pt}
\begin{tabularx}{\columnwidth}{@{}>{\raggedright\arraybackslash}Xrr@{}}
\toprule
& Materials & Biomed. \\
\midrule
Fields judged faithful & 64\% & 49\% \\
Fields judged reasonable default & 33\% & 46\% \\
Fields judged unsupported & 1\% & 4\% \\
Fields judged contradicting & 2\% & 0\% \\
Cards with no flagged field & 25 / 30 & 25 / 30 \\
Mark agrees with judge & 77\% & 59\% \\
Compatible with rationale$^\dagger$ & 11 / 11 & -- \\
\bottomrule
\end{tabularx}
\caption{Judge audit of the first-run cards of the 60 claims (30 per
domain). $^\dagger$With the expert rationale, on cards where the judge
returned a ruling; it returned none for the other 19 materials-science cards.}
\label{tab:audit}
\end{table}

The judge saw each card with its provenance marks and labeled every field
as \emph{faithful} (the value restates the claim explicitly or by direct
implication), a \emph{reasonable default} (the claim gives no value and an
expert would accept the one chosen), \emph{unsupported} (adds a constraint
the claim does not support), or \emph{contradicting}. A provenance mark
agreed with the judge's label when a field marked \emph{stated} was judged
\emph{faithful}, or a field marked \emph{inferred} or \emph{default}
received any other label. On materials-science claims, most disagreements were fields
the card marked inferred and the judge called faithful. On biomedical
claims, most were fields the card marked stated and the judge called a
reasonable default, so there the marks, if anything, understate how much
\system{} supplies. For the materials-science claims, we also
asked the judge whether the expert's rationale for the gold label relies on
an interpretation of the claim that the card excludes. It returned a ruling
for 11 of the 30 cards, and in all 11 the card was compatible with the
rationale.

Among the flagged fields was the threshold of the card for ``The surface
energy of a (100) MgO surface is 61.8\,meV/\AA$^2$.'' The card recorded it
as ``exact value not supplied,'' and the judge objected that the claim
supplies one. The judge also flagged a card that marked the conditions
unspecified for a claim that lists annealing temperatures. These are the
two errors that Section~\ref{sec:faithful} traces to the parser, which sees
only the agent's first reasoning step and not the claim. That step did not
restate the values, although the agent's own searches used them.

On the opening claim, all five cards from runs without computational
libraries marked as stated the threshold that requires a single solid phase at every
temperature below 700\,K (Figure~\ref{fig:overview}), and four of the five
from runs with the computational tools did the same, although the expert who labeled the claim judged the
temperature range unspecified. A card extracted from the first step can
also differ from the interpretation a run finally applies, as it did in
one card run with the tools on the Al--Mg--Zn $\tau$-phase claim, the run
that preferred room temperature (Appendix~\ref{app:samecalc}).
The cards from the runs with the computational tools were not audited.

\begin{table*}[t]
\centering
\footnotesize
\setlength{\tabcolsep}{5pt}
\begin{tabular}{@{}lrrlr@{}}
\toprule
Per run & Card & No card & Card $-$ no card & Rel. \\
\midrule
\multicolumn{5}{@{}l}{\emph{Without computational libraries (60 claims)}} \\
Input tokens & 40.4K & 35.6K & $+4.8$K \ci{$+3.9$K}{$+5.6$K} & $+13$\% \\
Output tokens & 11.1K & 10.3K & $+0.7$K \ci{$+0.5$K}{$+1.0$K} & $+7$\% \\
Cost (US\$) & \$0.27 & \$0.24 & $+0.023$ \ci{$+0.018$}{$+0.028$} & $+9$\% \\
Prompt tokens served from the cache & 65\% & 65\% & -- & -- \\
Wall-clock time (min) & 3.8 & 3.5 & $+16$\,s \ci{$+9$\,s}{$+24$\,s} & $+8$\% \\
Agent steps & 3.2 & 3.1 & $+0.07$ \ci{$-0.03$}{$+0.17$} & $+2$\% \\
Searches, materials science & 2.3 & 2.2 & $+0.06$ \ci{$-0.07$}{$+0.19$} & $+3$\% \\
Searches, biomedical & 1.9 & 1.8 & $+0.12$ \ci{$-0.03$}{$+0.27$} & $+7$\% \\
\midrule
\multicolumn{5}{@{}l}{\emph{With the computational tools (24 claims)}} \\
Input tokens & 39.3K & 33.7K & $+5.6$K \ci{$+3.5$K}{$+7.8$K} & $+16$\% \\
Output tokens & 10.5K & 9.9K & $+0.64$K \ci{$-0.07$K}{$+1.32$K} & $+6$\% \\
Cost (US\$) & \$0.256 & \$0.232 & $+0.023$ \ci{$+0.008$}{$+0.038$} & $+10$\% \\
Prompt tokens served from the cache & 67\% & 66\% & -- & -- \\
Wall-clock time (min) & 6.1 & 5.2 & $+55$\,s \ci{$-31$\,s}{$+138$\,s} & $+18$\% \\
\quad outside tool calls & -- & -- & $+26$\,s \ci{$+4$\,s}{$+48$\,s} & -- \\
Agent steps & 2.97 & 2.84 & -- & -- \\
Searches, materials science & 1.42 & 1.24 & $+0.18$ \ci{$+0.02$}{$+0.36$} & -- \\
\bottomrule
\end{tabular}
\caption{What a card adds to a run. Rows give per-run means for card and no-card runs. Tokens and cost count only GPT-5.4, which runs the agent, the report step and the scorer. Cost is at list price and does not apply the discount for cached prompt tokens, which were most of the prompt tokens. The difference is paired over claims, with bootstrap intervals in brackets, and \emph{Rel.}\ is the relative difference in means. Searches are reported by domain because the tool-use policies allow different numbers of searches in each domain (Appendix~\ref{app:setup-more}). The 24 claims are all in materials science. Dashes mark values not computed.}
\label{tab:overhead}
\end{table*}

\paragraph{Overhead:} Table~\ref{tab:overhead} compares card and no-card
runs. We counted the tokens of GPT-5.4, which runs the agent,
the report step and the scorer, and priced them at list price, without the
discount for cached prompt tokens. Cached tokens were most of the prompt
tokens (65\% in the runs without computational libraries and 67\% in those
with the tools). On the 60 claims, runs with a card used 40.4K input tokens against 35.6K without,
7\% more output tokens, and 3.8 against 3.5 minutes of wall-clock time
($+16$\,s per run paired by claim). Agent steps (3.2
per run with a card, 3.1 without) and searches (2.1 and 2.0) were about the
same, and a card raised the cost per run by 9\% (\$0.27 against \$0.24,
rounded to the cent). With the
computational tools, card runs used 39.3K input tokens against 33.7K, or
16\% more. Output tokens rose
by 6\% and wall-clock time from 5.2 to
6.1 minutes ($+18\%$, a paired difference of $+55$\,s), and the
intervals of both paired differences include zero. The paired
difference in time spent outside the calls to the computational tools was
$+26$\,s. Runs took 2.97 steps with a card and 2.84 without,
and cost \$0.256 against \$0.232 ($+10\%$). Card runs with the tools
also searched more often (Appendix~\ref{app:tooluse-more}). The parser's
tokens were not counted, but it reads only the first reasoning step, a median of about 90 words
(Section~\ref{sec:generation}), and writes one card. The tokens of the
search agent, the coding agent and the condenser, which other models run,
were not counted either, so these figures are not the full cost of a run,
and the relative overhead could differ if those components were
counted.

\section{Accuracy, Abstention and Variance}
\label{app:outcomes-more}

Table~\ref{tab:outcomes} gives sign accuracy, abstention and variance by
setting, condition and domain.

\begin{table*}[t]
\centering
\footnotesize
\setlength{\tabcolsep}{2.5pt}
\begin{tabular}{@{}lllllc@{}}
\toprule
 & Sign accuracy & Abstention & SD & Sign disagreement & \begin{tabular}[b]{@{}c@{}}Runs calling\\a tool\end{tabular} \\
\midrule
\multicolumn{6}{@{}l}{\emph{Without computational libraries (60 claims)}} \\
Card & 0.68 & 0.14 & 0.32 & 16/60 & -- \\
No card & 0.71 & 0.10 & 0.29 & 15/60 & -- \\
Frozen & 0.62 & 0.18 & 0.28 & 15/60 & -- \\
Card $-$ no card & $-0.03$ \ci{$-0.07$}{$+0.01$} & $+0.04$ \ci{0.00}{$+0.08$} & $+0.03$ \ci{$-0.05$}{$+0.11$} & $+0.02$ \ci{$-0.10$}{$+0.13$} & \\
Frozen $-$ card & $-0.06$ \ci{$-0.13$}{0.00} & $+0.04$ \ci{$-0.01$}{$+0.10$} & $-0.04$ \ci{$-0.16$}{$+0.07$} & $-0.02$ \ci{$-0.15$}{$+0.12$} & \\
Frozen $-$ no card & $-0.09$ \ci{$-0.17$}{$-0.02$} & $+0.08$ \ci{$+0.02$}{$+0.15$} & $-0.01$ \ci{$-0.13$}{$+0.10$} & 0.00 \ci{$-0.13$}{$+0.13$} & \\
\midrule
\multicolumn{6}{@{}l}{\emph{Without computational libraries, materials science (30 claims)}} \\
Card & 0.53 & 0.26 & 0.34 & 11/30 & 29/150 \\
No card & 0.56 & 0.19 & 0.36 & 13/30 & 31/150 \\
Frozen & 0.43 & 0.36 & 0.36 & 13/30 & 26/150 \\
Card $-$ no card & $-0.03$ \ci{$-0.11$}{$+0.03$} & $+0.07$ \ci{$-0.01$}{$+0.15$} & $-0.02$ \ci{$-0.12$}{$+0.07$} & $-0.07$ \ci{$-0.23$}{$+0.10$} & \\
Frozen $-$ card & $-0.10$ \ci{$-0.17$}{$-0.03$} & $+0.10$ \ci{0.00}{$+0.21$} & $+0.02$ \ci{$-0.17$}{$+0.20$} & $+0.07$ \ci{$-0.13$}{$+0.27$} & \\
Frozen $-$ no card & $-0.13$ \ci{$-0.23$}{$-0.05$} & $+0.17$ \ci{$+0.05$}{$+0.29$} & 0.00 \ci{$-0.20$}{$+0.19$} & 0.00 \ci{$-0.23$}{$+0.23$} & \\
\midrule
\multicolumn{6}{@{}l}{\emph{With the computational tools (24 claims)}} \\
Card & 0.67 & 0.11 & 0.38 & 7/24 & 64/120 \\
No card & 0.64 & 0.13 & 0.29 & 8/24 & 72/120 \\
Frozen & 0.68 & 0.12 & 0.31 & 6/24 & 64/120 \\
Card $-$ no card & $+0.03$ \ci{$-0.03$}{$+0.08$} & $-0.03$ \ci{$-0.08$}{$+0.02$} & $+0.09$ \ci{$-0.03$}{$+0.23$} & $-0.04$ \ci{$-0.17$}{$+0.08$} & \\
Frozen $-$ card & $+0.02$ \ci{$-0.05$}{$+0.09$} & $+0.01$ \ci{$-0.02$}{$+0.04$} & $-0.07$ \ci{$-0.25$}{$+0.09$} & $-0.04$ \ci{$-0.17$}{$+0.08$} & \\
Frozen $-$ no card & $+0.04$ \ci{$-0.06$}{$+0.16$} & $-0.02$ \ci{$-0.09$}{$+0.05$} & $+0.02$ \ci{$-0.13$}{$+0.18$} & $-0.08$ \ci{$-0.25$}{$+0.08$} & \\
\bottomrule
\end{tabular}
\caption{Outcomes by condition, with paired differences. \emph{Sign accuracy} is the share of runs whose score has the sign of the gold label, with 0 counted as its own sign, and \emph{abstention} is the share of runs scoring 0. Always predicting a positive score would give a sign accuracy of 0.58 over all 60 claims, 0.50 on their materials-science claims and 0.67 on their biomedical claims, and 0.58 on the 24 claims. \emph{SD} is the standard deviation of a claim's five scores, averaged over claims, and \emph{sign disagreement} counts claims whose five runs do not all share a sign. \emph{Runs calling a tool} counts runs with at least one call to the CALPHAD or CHGNet tool, and Table~\ref{tab:tooluse} gives the paired difference for the runs with the computational tools. Only materials-science claims could call these tools, and in the runs on the 60 claims their libraries were not installed, so 89 of the 90 calls returned no calculation. For the 60 claims, the frozen card is the card of the first card run, and for the 24 it is a card from a separate preliminary run. Difference rows are paired over claims, with bootstrap intervals in brackets; for sign disagreement they give the difference in the share of claims. Tool calls are counted for the materials-science claims only, so the rows for all 60 claims show dashes.}
\label{tab:outcomes}
\end{table*}

\paragraph{Accuracy:} On the 60 claims, sign accuracy over five runs was
0.62 with a frozen card. Over runs with a nonzero score, the sign accuracy
of frozen runs was 0.76 against 0.79 without a card. Frozen and card runs
did not differ measurably in sign accuracy ($-0.06$).
With the computational tools, sign accuracy was 0.68 with a frozen card, a difference
of $+0.02$ from card runs. In a 20-claim pilot on
GPT-5.4 and two other LLM backbones, we again saw no consistent difference
between card and no-card runs (Table~\ref{tab:backbones} in
Appendix~\ref{app:pilot}), although one run per claim on 20 claims can
show only large effects. Scores were identical with and without a card
on 47 of 60 pairs of LLM backbone and claim, and the two conditions never
gave opposite signs on a claim both decided. The pilot's LLM backbones
differed more from one another than card runs did from no-card runs,
particularly in abstention. Under the same guidelines, GPT-6 Sol had an
abstention rate of 0.70 on materials-science claims with and without a
card, and Gemini 3.1 Pro a rate of 0.10--0.20.

\subsection{Abstention and Variance Sources}
\label{app:abstention}

\paragraph{Abstention without computational libraries:} On the 60 claims,
abstention was 0.14 with a card, 0.10 without and 0.18 with a frozen card. Nearly all
abstentions were on materials-science claims. Of the 150 materials-science
runs in each condition, 39 card, 29 no-card and 54 frozen runs scored 0,
against 3, 1 and 0 of the 150 biomedical runs per condition. Contrasts
pooled over both domains are therefore smaller than the materials-science
contrasts. Card minus no card in abstention was $+0.04$
over all 60 claims, $+0.07$ on materials-science claims and $+0.01$
on biomedical claims. Frozen minus no card was
$+0.08$ over all claims, $+0.17$
on materials-science claims and $-0.01$
on biomedical claims. Frozen minus card was $+0.04$,
$+0.10$ and $-0.02$. Of these contrasts, only frozen minus no card, over
all claims and on materials-science claims, has an interval that excludes
zero.\footnote{Table~\ref{tab:outcomes} gives the intervals over all
claims and on materials-science claims. On biomedical claims, card minus
no card was $+0.01$ \ci{$-0.01$}{$+0.05$}, frozen minus no card $-0.01$
\ci{$-0.02$}{$0.00$} and frozen minus card $-0.02$
\ci{$-0.05$}{$0.00$}.}

In the rest of this appendix, a \emph{zero} is a run that scored 0. One
coder gave each of the 122 materials-science zeros a primary reason, taken
from the scorer's rationale.
\begin{itemize}[leftmargin=1.6em,itemsep=0pt,topsep=2pt]
\item[E] The named material, system or route was not found, only analogs or
a mechanism.
\item[P] A partial match. The system is right, but a conjunct, a numeric
pair, linearity or the sign of a causal effect is not established.
\item[L] The evidence leans negative but is not decisive for an existential
or non-exhaustive claim.
\item[C] The decisive calculation or the searches failed, and the rationale
says so.
\item[U] The claim was judged underspecified.
\end{itemize}
Missing or partial evidence (E or P) was the primary reason for 32 of the
39 card zeros (82\%), 22 of 29 no-card zeros (76\%) and 41 of 54 frozen
zeros (76\%). No rationale gave conflicting evidence as a reason, and we
take the zeros to mark gaps in the evidence rather than conflicts in it.
Few rationales named the claim's underspecification as a primary or
secondary reason (4 card, 3 no-card and 2 frozen runs). The 15 extra zeros of frozen runs over card runs came mostly
from partial matches (16 against 7) and failed searches or calculations (7
against 2), while missing evidence was as common in both (25 runs each).
Twelve of the 15 fall on three claims (the toluene polysulfide,
Al--Mg--Zn $\tau$-phase and rock-salt cathode capacity claims). Without
them, frozen minus card was $+0.02$, and without the
Al--Mg--Zn $\tau$-phase claim alone it was $+0.08$.
Frozen runs had more zeros than card runs on 10 claims (21 runs) and fewer
on 4 (6 runs). The frozen card is the card of each claim's first card run,
and on materials-science claims the first card runs abstained less often (0.20)
than the other four card runs (0.275). We have not established why frozen
runs abstained more often on those three claims.

\paragraph{Runs whose tool call failed:} Without computational libraries, 85 of the 450
materials-science runs called a computational tool that returned no
calculation. Compared with the other runs of the same condition, they
abstained in 8 of 29 card runs against 31 of 121, in 10 of 30 no-card runs
against 19 of 120, and in 10 of 26 frozen runs against 44 of 124. Their
sign accuracy was 0.48 against 0.54 in card runs, 0.37 against 0.61 in
no-card runs and 0.38 against 0.44 in frozen runs. Without all 86 runs
that called a tool, including the one whose call returned a calculation, frozen
minus card abstention on materials-science claims was $+0.10$
(26 claims) and frozen minus no card $+0.18$
(25 claims), both with intervals that exclude zero, so failed tool calls
do not explain the extra abstention of frozen runs.\footnote{The
differences were $+0.10$ \ci{$+0.01$}{$+0.20$} and $+0.18$
\ci{$+0.08$}{$+0.30$}.} Both comparisons select on whether a run
called a tool, which the condition may influence.

Most abstaining runs had received an empty search result. At least one
search came
back empty in 32 of 39 card zeros, 26 of 29 no-card zeros and 50 of 54
frozen zeros, against 41 of 111, 38 of 121 and 35 of 96 materials-science runs that
decided the claim. Runs without an empty search abstained in 7 of 77, 3 of
86 and 4 of 65 cases. In the 17 materials-science claims with both zero and
nonzero runs, the zero runs had the larger share of empty searches in 16,
by $+0.28$ on average ($+0.26$
without the runs hit by the search outage). The scorer's rationale
mentioned the existential quantifier in 24 of 39 card zeros, 14 of 29
no-card zeros and 40 of 54 frozen zeros, against 27 of 111, 20 of 121 and
33 of 96 runs that decided the claim. Under the guideline, failing to find
an instance of an existential claim is not negative evidence
(Appendix~\ref{app:card-more}), which may leave 0 as the only score open to
a run that finds nothing.

\paragraph{The search outage:} The outage hit frozen and edited-card runs
on the 60 claims (Section~\ref{sec:conditions}). Each failed search reached the agent
as an empty result, like an ordinary search that finds nothing. Search
output that cannot be parsed goes through a repair step, which also returns
an empty result when it fails. Not every run's failed searches could be
detected, so the counts below are lower bounds. No
failure was found in card or no-card runs. In the frozen runs, at least 45 searches
failed in 17 runs, 18 searches in 11 materials-science runs (6 of which scored 0)
and 27 in 6 biomedical runs (none of which scored 0). In the edited runs,
at least 34 searches failed in 10 runs, 6 in two quantifier-edited runs and
28 in eight threshold-edited runs.

Every search failed in three frozen runs on the Al--Mg--Zn $\tau$-phase
claim, and all three scored 0. Every search also failed in five biomedical
frozen runs, yet all five decided the claim and four had the gold sign.
Four of the five called the general coding agent. Materials-science runs had no
such fallback, because they reached a coding agent only through the CALPHAD
and CHGNet tools, whose libraries were not installed for these runs.

We kept the affected runs, and dropping the 17 affected frozen runs
changes little.
Frozen abstention was 0.17 without them (48 of 283 runs) and 0.18 with them
(54 of 300). Frozen minus no card in sign accuracy was $-0.085$,
against $-0.090$ with all
runs, and its interval still excludes zero.\footnote{Without the affected
runs the difference was $-0.085$ \ci{$-0.160$}{$-0.011$}, and with all runs
$-0.090$ \ci{$-0.167$}{$-0.017$}.} On materials-science claims, frozen minus card abstention stayed $+0.10$
and frozen minus no card $+0.17$, with the same intervals as with all runs
(Table~\ref{tab:outcomes}).
Table~\ref{tab:edits} gives the edit shifts without the affected runs,
and Appendix~\ref{app:edit-abstention} the abstention of threshold-edited
runs.

With the failed searches left out, between 27\% and 36\% of searches came
back empty in every condition except the threshold-edited runs on the 24
claims. For the 60 claims, the share was 34\% in card runs (214
of 635) and 32\% in no-card runs (196 of 608). It was 36\% in frozen runs
without the failed searches (40\% with them, 269 of 671), 36\% in
threshold-edited runs (47\% with them, 76 of 161) and 32\% in quantifier-edited runs
(36\%, 44 of 123). On the 24 claims it was 59 of 170 in card runs, 52 of
149 in no-card runs, 50 of 153 in frozen runs with the tools and 71 of 267
with search only. In their quantifier-edited runs it was 8 of 27 with the
tools and 13 of 48 with search only. Their threshold-edited runs had a larger
share, both with the tools (18 of 41 searches) and with search only
(Appendix~\ref{app:edits-more}), so a stricter threshold can raise the
share without an outage.

\paragraph{Abstention on the 24 claims:} On these claims, 13 card runs,
16 no-card runs, 14 frozen runs with the tools and 28 search-only runs
scored 0, each out of 120. Abstention did not differ measurably between
conditions, with differences of $-0.03$ between card
and no-card runs and $+0.01$ between frozen and card
runs (Table~\ref{tab:outcomes}). Nearly every run of every condition abstained on
two claims, the Fe$_2$O$_3$ dopant-ranking and Nd-doped Co$_3$O$_4$ claims,
which account for 10 of the 13 card zeros, 9 of the 16 no-card zeros, 10 of
the 14 frozen zeros and 10 of the 28 search-only zeros. Besides E, L and U,
the coder used two further codes for these runs. Here E covers no matched
evidence, only analogs or unmatched studies. Code X marks a calculation
that ran but that the scorer rejected as a proxy for the claimed quantity,
and in search-only runs, code Q marks a missing quantity that a calculation
could have supplied. E was the primary reason for 11 of 13 card zeros, 7 of
16 no-card zeros, 11 of 14 frozen zeros and 16 of 28 search-only zeros, and
X for 2 card, 8 no-card, 2 frozen and 0 search-only zeros. Q accounts for
10 of the 28 search-only zeros, four on the Li--AlCu reaction claim, three
on the Cu-in-Al stiffening claim, two on the Al--Mg--Zn $\tau$-phase claim
and one on the Cu-in-Al softening claim. Every search came back empty in 11
card runs (5 of which scored 0), 7 no-card runs (4), 4 frozen runs with the
tools (3) and 8 search-only runs (7).

\paragraph{Variance:} On the 60 claims, the standard deviation of a
claim's five scores was 0.32 with a card, 0.29 without and 0.28 with a
frozen card, and sign disagreement was 0.27, 0.25 and 0.25. Every paired
difference had an interval that includes zero (Table~\ref{tab:outcomes}),
for example $+0.03$ for the standard deviation with a card minus without
and $-0.04$ for frozen minus card. With the tools,
the standard deviations were 0.38, 0.29 and 0.31, with a card minus no-card
difference of $+0.09$, and 7, 8 and 6 of 24 claims had
runs of different signs.

\paragraph{Where score variation arises:} Among the frozen runs on the 24
claims, with the tools and with search only, one coder classed each run
whose score differed from its claim's median by the source of the
difference. Class (i) is a different decision
about calling a tool, (ii) a different calculation result, (iii) different
decisive search evidence, including empty or unverifiable retrieval, and
(iv) the same tool decision, calculation numbers and decisive sources with
a different verdict. A fifth class covers the scorer changing the agent's
stated score. Scores differed within 10 of the 24 claims with the tools and
13 of 24 with search only. With each run weighted by its squared deviation
from the median, class (iv) carried 26 of 44 units of squared deviation
(59\%) with the tools and 21 of 34 (62\%) with search only, and class (iii) 12 of 44 (27\%) and 12 of 34
(35\%). Class (i) carried 6 of 44 units (14\%) with the tools, class (ii)
none, and the scorer 1 of 34 (3\%) with search only. Four of the 16
deviating runs with the tools and 2 of the 19 with search only were
borderline between classes.

Most of class (iv) came from two claims in each setting. With the tools,
the Fe--Al Al$_2$Fe-onset and Li--CuAl capacity claims supplied 22 of its
26 units, and without them class (iv) carried 4 of 22 (18\%). With search
only, the Fe--Al Al$_2$Fe-onset and Al-doped Fe$_2$O$_3$ magnet claims
supplied 12 of 21, and without them it carried 9 of 18 (50\%). On the
Fe--Al Al$_2$Fe-onset claim, two of the five frozen runs with the tools
adopted in their first reasoning step the rule that one matched
counterexample would make the score negative, although the frozen card's
quantifier is typical. On the Li--CuAl capacity claim, two frozen runs
computed the same lithiation limit ($x$ about 0.38 and 0.385). The tool's
answer called it a maximum in one run and an arbitrary composition in the
other, and the runs scored $+2$ and $-1$. The scorer changed the agent's
stated score in 6 of the 478 runs on the 60 claims whose answer states one, and
its sign in 1, a no-card run on the Al--Mg--Zn $\tau$-phase claim that it
moved from 0 to $-1$. On the 24 claims it repeated the score in 391 of
the 395 runs whose answer stated one, and none of its 4 changes altered
the sign, so the scorer adds little of the variation.

\section{Pilot on Three LLM Backbones}
\label{app:pilot}

Table~\ref{tab:backbones} gives the results of the 20-claim pilot
mentioned in Section~\ref{sec:conditions} and discussed in
Appendix~\ref{app:outcomes-more}, in which GPT-5.4 and two other LLM
backbones were each run with and without a card.

\begin{table}[H]
\centering
\small
\setlength{\tabcolsep}{4pt}
\begin{tabular}{@{}lcrrrr@{}}
\toprule
& & \multicolumn{2}{c}{Materials} & \multicolumn{2}{c}{Biomed.} \\
\cmidrule(lr){3-4} \cmidrule(lr){5-6}
LLM backbone & Card & Sign & Abst. & Sign & Abst. \\
\midrule
GPT-6 Sol & yes & 0.20 & 0.70 & 0.70 & 0.30 \\
& no & 0.20 & 0.70 & 0.80 & 0.20 \\
GPT-5.4 & yes & 0.40 & 0.20 & 1.00 & 0.00 \\
& no & 0.30 & 0.40 & 1.00 & 0.00 \\
Gemini 3.1 Pro & yes & 0.60 & 0.20 & 0.90 & 0.00 \\
& no & 0.70 & 0.10 & 0.90 & 0.00 \\
\bottomrule
\end{tabular}
\caption{Sign accuracy and abstention rate on a 20-claim pilot (10 per
dataset, one run each) for three LLM backbones, with and without a card.}
\label{tab:backbones}
\end{table}

\section{Tool Use in Detail}
\label{app:tooluse-more}

For the runs with the computational tools, Table~\ref{tab:tooluse} counts
the tool calls, the calls that returned a calculation and the runs that
received one, under the definitions of Appendix~\ref{app:setup-more}. Several of the analyses in
this appendix use codes that one coder assigned, and a second reader
checked the counts taken from those codes.

\begin{table*}[t]
\centering
\footnotesize
\setlength{\tabcolsep}{4pt}
\begin{tabular}{@{}llllr@{}}
\toprule
 & Card & No card & Frozen & Card $-$ no card \\
\midrule
Runs calling a computational tool & 64/120 & 72/120 & 64/120 & $-0.07$ \ci{$-0.17$}{$+0.02$} \\
Calls, CALPHAD\,+\,CHGNet$^{a}$ & 66 (56\,+\,10) & 72 (54\,+\,18) & 64 (49\,+\,15) & $-0.05$ \ci{$-0.16$}{$+0.05$} \\
Calls returning a calculation$^{b}$ & 64/66 & 66/72 & 63/64 & -- \\
Runs receiving a calculation & 62/120 & 66/120 & 63/120 & $-0.03$ \ci{$-0.14$}{$+0.07$} \\
First reasoning step names CALPHAD & 97/120 & 97/120 & 105/120 & -- \\
First reasoning step names CHGNet & 41/120 & 42/120 & 52/120 & -- \\
\addlinespace
Searches per run & 1.42 & 1.24 & 1.28 & $+0.18$ \ci{$+0.02$}{$+0.36$} \\
Empty searches & 59/170 & 52/149 & 50/153 & -- \\
Runs with only empty searches (scoring 0) & 11 (5) & 7 (4) & 4 (3) & -- \\
\addlinespace
Tool answers carrying their own score$^{c}$ & 10/66 & 17/72 & 10/64 & -- \\
\quad its sign equals the run's final sign & \multicolumn{3}{c}{28/37, pooled} & -- \\
\bottomrule
\end{tabular}

\vspace{8pt}
\begin{tabular}{@{}>{\raggedright\arraybackslash}p{3.6cm}llll@{}}
\toprule
 & \multicolumn{2}{c}{Quantifier edit} & \multicolumn{2}{c}{Threshold edit} \\
\cmidrule(lr){2-3}\cmidrule(lr){4-5}
Edited cards, frozen, with the tools & Edited vs unedited & Difference & Edited vs unedited & Difference \\
\midrule
Runs calling a computational tool & 11/20 vs 53/100 & $+0.02$ \ci{$-0.13$}{$+0.19$} & 15/24 vs 64/120 & $+0.09$ \ci{$-0.05$}{$+0.25$} \\
Calls returning a calculation$^{b}$ & 10/11 & -- & 15/15 & -- \\
Edited value in calculation questions$^{d}$ & 4/11 vs 0/53 & -- & 14/15 vs 0/64 & -- \\
\bottomrule
\end{tabular}
\caption{How the agent used the computational tools on the 24 claims. The upper part compares the card, no-card and frozen-card conditions, and its last column gives card minus no card, paired over claims (per-claim shares or per-run means), with bootstrap intervals in brackets. An empty search is a search call that returned no results. The lower part compares each edited-card run with the five unedited frozen runs of the same claim, and its difference is in the share of runs calling a tool. $^{a}$The difference is in calls per run. $^{b}$A call returns a calculation if the agent receives the result, in full or in part, of a calculation that the coding agent ran (Appendix~\ref{app:setup-more}). $^{c}$A tool answer carries its own score when it states one on the agent's scale. The playbooks ask the coding agent for one (Appendix~\ref{app:tools}). $^{d}$Among runs that asked a calculation question (Table~\ref{tab:propagation}). Dashes mark values not computed.}
\label{tab:tooluse}
\end{table*}

\paragraph{Calls and calculations:} Frozen-card runs called a computational
tool in 64 of 120 runs, as many as card runs. Nearly every call returned a
calculation to the agent (64 of 66 calls with a card, 66 of 72 without, 63
of 64 with the frozen card), so the runs that received a calculation
followed the same pattern (62 card, 66 no-card and 63 frozen-card runs of
120). Card runs made 66 computational
calls (56 to CALPHAD and 10 to CHGNet; two runs called both), no-card runs
made 72 (54 and 18) and frozen runs with the tools made 64 (49 and 15).
Per claim, card minus no card was $-0.05$ in calls
per run and $-0.03$ in the share of runs receiving a
calculation.

\paragraph{What the first reasoning step and the card name:} The first
reasoning step named CALPHAD in 97 of 120 card runs, 97 of 120
no-card runs and 105 frozen runs. It named CHGNet in 41 card, 42 no-card
and 52 frozen runs. No card named a computational tool in any field,
whether it was written in a run (0 of the 118 parsed cards) or frozen (0 of
the 24 frozen cards).

\paragraph{Fewer calling runs with a card:} The number of runs calling a
tool was lower with a card, although the difference includes zero
(Section~\ref{sec:tooluse}). Each
call runs a coding agent that writes and executes code in its own
workspace, and a call that
returned a calculation took a median of 2.6 minutes (2.3 for CALPHAD and
6.0 for CHGNet), so fewer calls would be a saving if the difference held
up. Sign accuracy was no lower with a card ($+0.03$
against no card). On six claims, card runs called a tool in fewer runs than
no-card runs (6 against 18 calling runs). They are the Li--CuAl capacity,
P-in-Si interstitial, Cu-in-Al softening and Mg-in-Al stiffening claims and
the two Co$_3$O$_4$ doping claims, and on them sign accuracy was 0.57 in
both conditions. We
could not show the saving, however. Card runs took longer than no-card
runs, by an amount whose interval includes zero
(Appendix~\ref{app:faithful-more}), and they also searched more often,
1.42 times per run against 1.24 for no-card runs ($+0.18$; Table~\ref{tab:tooluse}).\footnote{Card minus no
card in searches per run was $+0.10$ \ci{0.00}{$+0.22$} on the
tool-calling claims of Appendix~\ref{app:tools-search}, an interval that
reaches zero, and $+0.25$ \ci{$-0.05$}{$+0.58$} on the other claims.}

\paragraph{Which claims got a tool call:} Frozen runs called a tool on
different claims from card and no-card runs. Against card runs the
difference lay entirely in which claims the calls fell on, since both
conditions called a tool in 64 of 120 runs. On the 12 tool-calling claims,
the agent called a tool in 0.83 of card
runs, 0.85 of no-card runs and 0.93 of frozen runs, and card minus no card
was $-0.02$. On the other 12 claims the shares were
0.23, 0.35 and 0.13. There, card minus no card was $-0.12$
and frozen minus no card $-0.22$, and only the frozen difference has an
interval that excludes zero.\footnote{The two differences were $-0.12$
\ci{$-0.30$}{$+0.05$} and $-0.22$ \ci{$-0.38$}{$-0.07$}.} To test whether the condition changed which claims
got a tool call, we permuted condition labels among each claim's 15 runs,
a million times per comparison, and also among only the 10 runs of the two
compared conditions (Appendix~\ref{app:stats}). Frozen against card runs gave $p = 0.03$ (0.014 when
only the two conditions' runs were permuted), frozen against no-card runs
$p = 0.007$ (0.012) and card against no-card runs $p = 0.25$ (0.28). On the
Cu-in-Al lattice-parameter claim, three of the five frozen runs called the
CHGNet tool, and no card or no-card run did. Its frozen card described the
manipulation in computational terms, as substituting Cu for Al ``in an fcc
Al supercell.'' These contrasts with frozen runs also span differences in
the day, the computer and the source of the frozen card
(Section~\ref{sec:conditions} and Appendix~\ref{app:tools}), which may
account for part of them, so we treat them as descriptive. Without the
three Fe--Al claims, on which many card and no-card calculations used
another database, permuting among each claim's 15 runs gave $p = 0.018$ for frozen
against card runs, 0.008 for frozen against no-card runs and 0.29 for card
against no-card runs, and card minus no card in the share of runs calling
a tool was $-0.09$.

\paragraph{Frozen-card wording in calculation questions:} The agent writes
each calculation question itself. To see whether a
frozen card's wording reached the tools, we took each calculation question,
kept its word trigrams that do not occur in the claim, and measured the
share that occur in the claim's frozen card, averaged per claim. Because
each card run produced its own card and none equals the frozen card, which
came from a separate preliminary card run, the share in card runs gives
the overlap expected from describing the same claim. On the 15 claims on which frozen and card runs both asked a
calculation question, the share was 0.15 in frozen runs and 0.06 in card
runs, a paired difference of $+0.09$. Against no-card
runs, also on 15 claims, the share was 0.14 in frozen runs and 0.06 in
no-card runs, a difference of $+0.08$. Restricted to
tool-calling claims with at least two calculation questions in each
condition,
the shares were 0.19 and 0.08 against card runs (10 claims) and 0.17 and
0.07 against no-card runs (11 claims), and leaving out the three Fe--Al
claims gave differences of $+0.10$ and $+0.09$. Each of these differences
has an interval that excludes zero.\footnote{Against card and no-card
runs, the differences were $+0.09$ \ci{$+0.06$}{$+0.14$} and $+0.08$
\ci{$+0.04$}{$+0.13$} on all 15 claims, $+0.10$ \ci{$+0.06$}{$+0.17$} and
$+0.11$ \ci{$+0.06$}{$+0.16$} on the tool-calling claims with at least two
questions, and $+0.10$ \ci{$+0.05$}{$+0.15$} and $+0.09$
\ci{$+0.04$}{$+0.14$} without the Fe--Al claims.} The frozen card sits in the agent's prompt, so some
reuse of its wording is expected, and we read this result as a check that
the card reached the tools rather than as a finding about tool use.

\paragraph{Edited cards:} Appendix~\ref{app:propagation} gives how often
the edited runs called a tool and the composition ranges they calculated
over. A card given to the agent, frozen or edited, changed the
questions sent to the tools, through its wording or its edited value, but
not measurably how often the tools were called.

\paragraph{The tools' own scores:} Following their playbooks
(Appendix~\ref{app:tools}), some coding-agent answers carried a score of
their own, which we detected by pattern (a Likert score, ``Verdict'' with a
number, or a signed score in parentheses). In the card, no-card and frozen
runs with the tools, 37 of the 202 answers did, 10 in card runs, 17 in no-card runs
and 10 in frozen runs. Of these 37, 5 came from CALPHAD and 32 from CHGNet.
The sign of the tool's score equaled the sign of the run's final score in 28
of the 37, and the two scores were equal in 22. In the edited runs with the
tools, 6 of 26 answers carried a score, all from CHGNet, and the score's
sign equaled the run's final sign in 4. We did not test whether these scores
influenced the agent's verdict. Part of the judgment of whether a result
meets the claim may thus be made by the coding agent, under tolerances
that no card field sets.

\paragraph{The composition basis on the Li--CuAl capacity claim:} One card
run and one no-card run took the claim's $x$ as a lithium count per
formula unit, whereas the expert's code and every other run used the
lithium atom fraction. The frozen card leaves the basis open as well. Its
\field{Units} value, ``Dimensionless composition fraction x,'' does not say
whether $x$ counts lithium atoms or formula units. A card field that
defined $x$, in \field{Units} or \field{SystemEntity}, would have recorded
this choice.

\paragraph{Same calculation, same score:} We also compared how often two
runs of a claim that obtained the same calculation result gave the same
score, under the frozen card and under the cards of the card runs. On the
tool-calling claims, the share of such pairs with the same score, averaged
over claims, was 0.77 under the frozen card (12 claims) and 0.72 in the
card runs (9 claims). On the 9 claims with such pairs in both conditions,
the paired difference of $+0.16$ was not measurably different from
zero,\footnote{The difference was $+0.16$ \ci{$-0.06$}{$+0.38$}.} and
most of it comes from the Al--Mg--Zn $\tau$-phase claim ($+0.80$) and the
Al--Mg--Zn two-phase solidification claim ($+0.50$).

\subsection{Frozen Cards with the Computational Tools and with Search Only}
\label{app:tools-search}

Only the frozen cards were run both with
the computational tools and with search only, so this comparison varies
whether the agent could calculate and keeps the card, the claims and the
day of the runs the same.

\begin{table}[H]
\centering
\footnotesize
\setlength{\tabcolsep}{2.5pt}
\begin{tabular}{@{}>{\raggedright\arraybackslash}p{1.85cm}rrrrrr@{}}
\toprule
 & \multicolumn{3}{c}{With the tools} & \multicolumn{3}{c}{Search only} \\
\cmidrule(lr){2-4}\cmidrule(lr){5-7}
Claims & Gold & Abst. & Wrong & Gold & Abst. & Wrong \\
\midrule
All claims & 82 & 14 & 24 & 68 & 28 & 24 \\
Tool-calling & 43 & 3 & 14 & 37 & 10 & 13 \\
Other claims & 39 & 11 & 10 & 31 & 18 & 11 \\
\midrule
SD & \multicolumn{3}{c}{0.31} & \multicolumn{3}{c}{0.31} \\
Sign disagr. & \multicolumn{3}{c}{6/24} & \multicolumn{3}{c}{8/24} \\
\bottomrule
\end{tabular}
\caption{Outcomes of frozen-card runs with the computational tools and with search only, on all 24 claims (120 runs per setting) and on two post hoc groups of 12 claims. The tool-calling claims are those on which at least three of the five frozen runs with the tools called a computational tool, and the other claims are the rest. \emph{Gold} counts runs whose score has the sign of the expert label, \emph{Abst.}\ runs scoring 0 on claims whose label is not 0, and \emph{Wrong} the remaining runs. SD and sign disagreement (\emph{Sign disagr.}) are over all claims and defined as in Table~\ref{tab:outcomes}.}
\label{tab:tools-search}
\end{table}

\begin{figure}[t]
\centering
\includegraphics[width=\columnwidth]{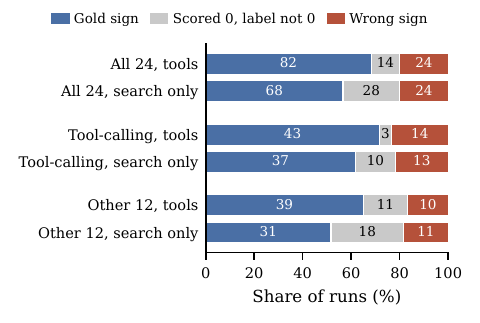}
\caption{Outcomes of the frozen-card runs with the computational tools and
with search only, on all 24 claims, on the 12
tool-calling claims and on the other 12. Each bar splits the runs into
those with the gold sign, those scoring 0 on a claim whose gold label is
not 0, and those with the wrong sign, which include nonzero scores on the
Al--Zn single-phase claim, the only claim labeled 0. The numbers in the bars are
run counts, out of 120 per bar
for all claims and 60 for each group of 12. A claim is tool-calling if at
least three of its five frozen runs with the tools called a computational
tool.}
\label{fig:tools-search}
\end{figure}

\begin{table}[t]
\centering
\footnotesize
\setlength{\tabcolsep}{3pt}
\begin{tabular}{@{}lrl@{}}
\toprule
Tools $-$ search only & Claims & Difference \\
\midrule
\multicolumn{3}{@{}l}{\emph{Sign accuracy}} \\
All claims & 24 & $+0.12$ \ci{$+0.03$}{$+0.23$} \\
\quad without the rerun claims$^{a}$ & 22 & $+0.06$ \ci{0.00}{$+0.15$} \\
\quad without nine claims$^{b}$ & 15 & $+0.08$ \ci{$-0.01$}{$+0.19$} \\
\quad without two claims$^{c}$ & 22 & $+0.06$ \ci{0.00}{$+0.14$} \\
Tool-calling claims & 12 & $+0.10$ \ci{$-0.05$}{$+0.28$} \\
Other claims & 12 & $+0.13$ \ci{$+0.03$}{$+0.25$} \\
Runs that made no call$^{d}$ & 15 & $+0.08$ \ci{0.00}{$+0.18$} \\
\multicolumn{3}{@{}l}{\emph{Abstention}} \\
All claims & 24 & $-0.12$ \ci{$-0.21$}{$-0.04$} \\
Tool-calling claims & 12 & $-0.12$ \ci{$-0.27$}{0.00} \\
Other claims & 12 & $-0.12$ \ci{$-0.23$}{$-0.02$} \\
\multicolumn{3}{@{}l}{\emph{All claims}} \\
Wrong sign & 24 & 0.00 \ci{$-0.03$}{$+0.03$} \\
Standard deviation & 24 & 0.00 \ci{$-0.16$}{$+0.17$} \\
Searches per run$^{e}$ & 24 & $+0.95$ \ci{$+0.55$}{$+1.38$} \\
\multicolumn{3}{@{}l}{\emph{Edit shift}} \\
Quantifier edit & 20 & $-0.07$ \ci{$-0.50$}{$+0.31$} \\
\quad without one claim & 19 & $+0.02$ \ci{$-0.41$}{$+0.39$} \\
Threshold edit & 24 & $-0.33$ \ci{$-0.70$}{$-0.01$} \\
\quad without one claim & 23 & $-0.30$ \ci{$-0.69$}{$+0.03$} \\
\bottomrule
\end{tabular}
\caption{Paired differences between the frozen-card runs with the
computational tools and those with search only, with
bootstrap intervals over claims. Sign accuracy, abstention and wrong signs
are the shares of runs counted in Table~\ref{tab:tools-search}, and the edit shifts
are those of Table~\ref{tab:edits}. $^{a}$Without the claims on which a
run with the tools was rerun. $^{b}$Without the nine claims whose gold
entries have documented problems (Section~\ref{sec:limitations}).
$^{c}$Without the Li--AlCu reaction and Al--Mg--Zn $\tau$-phase claims.
$^{d}$On the 15 claims with at least one frozen run with the tools that
made no computational call, those runs against the search-only runs.
$^{e}$Search only minus tools. The rows marked \emph{without one claim}
leave out the Li--AlCu reaction claim for the quantifier edit and the
Fe--Al equiatomic strengthening claim for the threshold edit.}
\label{tab:tools-search-diff}
\end{table}

With the computational tools, frozen-card runs had sign accuracy 0.68,
against 0.57 with search only, a paired difference of $+0.12$
(Figure~\ref{fig:tools-search} and
Tables~\ref{tab:tools-search} and~\ref{tab:tools-search-diff}). The whole difference came from abstention.
Runs with the tools abstained in 14 of 120 runs and search-only runs in 28,
a difference of $-0.12$. Wrong signs were as common
in both settings, 24 runs (0.20) in each. Runs with
the tools were more accurate on 8 claims, less accurate on 2 and tied on 14.
A sign test over the ten untied claims gave $p = 0.11$. An exact sign-flip
test of the mean difference, over every sign pattern of those ten
differences, gave $p = 0.045$. We report both tests and treat this
comparison as exploratory. The scores varied about as much in both
settings (SD 0.31 in each).
A few runs with the tools whose tool call failed were rerun, which can
favor the setting with the tools. Without
the claims on which a run with the tools was rerun, the difference was
$+0.06$ (sign-flip $p =
0.18$), and without the nine claims whose gold entries have documented
problems (Section~\ref{sec:limitations}) it was $+0.08$. The first of
these intervals reaches zero and the second includes it.

Two CALPHAD claims supplied half of the summed per-claim gain. Per claim,
the gain was $+0.8$ on the Li--AlCu reaction claim, $+0.6$ on the
Al--Mg--Zn $\tau$-phase and Cu-in-Al stiffening claims, and $+0.4$ or
$+0.2$ on the five other claims that gained. The two losses, on the
Li--CuAl capacity and Fe--Al Al$_2$Fe-onset claims, were $-0.2$ each. On the
Li--AlCu reaction claim (gold $+2$), all five runs with the tools obtained
the same $\eta$\,+\,$\zeta$\,+\,T1 phase field at the onset of lithiation
and scored $+2$. Four search-only runs abstained, and one scored $+1$. On
the Al--Mg--Zn $\tau$-phase claim (gold $+2$), all five runs with the tools
scored $+2$, and the search-only runs scored $0$, $0$, $+1$, $+1$ and $-1$.
One frozen run, for example, computed a $\tau$-phase fraction of 0.261 at
$x(\mathrm{Mg}) = 0.345$, $x(\mathrm{Zn}) = 0.04$ and 400\,K, above the
20\% the card's threshold requires. The fraction depends on the temperature
and the Zn content, and the card runs, which calculated at about 300\,K,
found 0.121 at 2\,at.\% Zn and 0.219 at 3\,at.\% Zn
(Table~\ref{tab:samecalc}). In every search-only run that abstained on
these two claims, the scorer's rationale gave as its reason that the
quantity on which the claim depends, a phase field or a phase fraction, was
missing from the evidence (our coding). Without these two claims, the
difference in sign accuracy was $+0.06$, and its
interval reaches zero.

The gain was not confined to claims on which the agent calculated. We call
a claim \emph{tool-calling} if at least three of its five frozen runs with
the tools called a computational tool. Twelve claims qualify, and they
carry a bullet in Table~\ref{tab:ids-tools}. The set was defined after the
runs with the tools, so it is post hoc, and contrasts within it are
descriptive. On the tool-calling claims, the difference in sign
accuracy was $+0.10$. On the other twelve, where only
8 of 60 frozen runs with the tools called a tool, it was $+0.13$, and
only this second interval excludes zero. Abstention fell by about as much in both groups,
$-0.12$ on the tool-calling claims (its interval
reaches zero) and $-0.12$ on the others. On the 15
claims with at least one frozen run with the tools that made no
computational call, those runs and the search-only runs did not differ
measurably in sign accuracy ($+0.08$). Runs
with the tools also searched less, 1.28 searches per run against 2.23, and 50 of their 153
searches came back empty, against 71 of 267. We have not established why runs
with the tools abstained less on claims where they seldom calculated, and we
do not attribute the gain to the calculations themselves.

Each edited card was run once with the tools and once with search only, so
the edit shifts can also be paired by claim. For the quantifier edit, the shift
with the tools minus the shift with search only was $-0.07$,
or $+0.02$ without the
Li--AlCu reaction claim. For the threshold edit it was $-0.33$,
and $-0.30$ without the
Fe--Al equiatomic strengthening claim, whose edited search-only run
received only empty searches (Table~\ref{tab:tools-search-diff}). Each of these uses one edited run per claim
and setting. The threshold difference also includes the irregular runs on
the Li--AlCu reaction and Al--Mg--Zn $\tau$-phase claims
(Appendix~\ref{app:propagation}), and its interval nearly reaches zero, so
we do not count it as a finding. With search only, half of the
threshold-edited runs abstained (Section~\ref{sec:control} and
Appendix~\ref{app:edits-more}).

\section{Extended Related Work}
\label{app:related-more}

\paragraph{Scientific claim verification and feasibility:} FEVER
\citep{thorne-etal-2018-fever} and SciFact \citep{wadden-etal-2020-fact}
cast verification as retrieving evidence and predicting a label.
SciFact-Open \citep{wadden-etal-2022-scifact-open} extends SciFact to
open-domain retrieval over 500K abstracts. Its analysis of the
retrieved evidence found abstracts that support only a special case of a
claim, so whether such a claim counts as supported depends on how broadly
it is interpreted. Because \system{} searches the web with pages that host
SciFact excluded, we use the SciFact development claims and labels without
the SciFact corpus. For the same reason, the verifiers that SciFact-Open
evaluates, which assess a claim against its corpus, are not comparable
with \system{}. Our task comes from SFBench
\citep{costello-etal-2026-sfbench}, which asks how likely an expert-written
claim is to be feasible, that is, achievable by a skilled team with
current knowledge and technology.

Matter-of-Fact \citep{jansen-etal-2025-matter}, the closest work to our
task, evaluates separate baselines for literature retrieval and for code
generation. Unlike its binary labels, a five-point scale like SFBench's
has a middle label, 0, with which an expert can record that a claim is too
underspecified to judge, as the expert who labeled our opening claim
did. In our experience,
Matter-of-Fact's labels were also too noisy for comparing an agent's
verdicts with gold labels claim by claim. We therefore took our
materials-science claims from SFBench, whose labels use a five-point scale
and come with an expert's written rationale, and the rationales also let a judge model
compare a card with the expert's interpretation
(Appendix~\ref{app:faithful-more}).

\citet{mohammadi-etal-2026-experiments} probe the feasibility judgments of
LLMs by varying how much experiment and outcome context
accompanies a claim, and observe that justifications given for the claim
alone often rest on generic plausibility rather than on explicit
experimental constraints. Their models are given the context. Because our
agent gathers its own evidence and we vary the interpretation, their probes
do not carry over to our setting.
\citet{liu-etal-2026-verification} build claims in which a less prominent
constraint is contradicted and find that verifiers often accept them.
Those claims contradict a stated constraint rather than omit one, so they
test whether a verifier checks every constraint a claim states, and we do
not use them. Listing each constraint as a card field might show which
ones a verifier tested. QuanTemp \citep{venktesh-etal-2024-quantemp}
collects real-world numerical claims, many of which lack precise
information. Drawn from news and politics and labeled for veracity rather
than feasibility, its claims do not fit our task.

\paragraph{Ambiguity and underspecification:} We follow
\citet{pezzelle-2023-dealing} in treating underspecification as a normal
property of language, which readers resolve from context and shared
knowledge. In machine learning, \citet{damour-etal-2022-underspecification}
call a training pipeline underspecified when it can return many models
with equally good held-out performance but different behavior in
deployment, and we use the term only for a claim that omits a detail
needed to assess it. \citet{vijayvargiya-etal-2026-ambigswe} separate ambiguity from
underspecificity, and \citet{kennedy-2011-ambiguity} separates ambiguity
from vagueness. By Kennedy's distinction, an unstated threshold for
``significant'' makes a claim vague rather than ambiguous. We use
ambiguity for the open choice among interpretations that
underspecification leaves, whatever its source. On DUST \citep{wildenburg-etal-2024-dust}, newer language models
identified underspecified sentences only moderately well when asked, and
interpreted them poorly and with little uncertainty, and crowdworkers judged GPT-4's
disambiguations of AmbiEnt sentences correct far less often than the
dataset's own \citep{liu-etal-2023-afraid}. Models also fill in
requirements that a prompt leaves unstated, but underspecified prompts are
more likely to regress when the model or the prompt changes
\citep{yang-etal-2026-prompts}, and adding explicit context to
underspecified queries can change evaluation conclusions
\citep{malaviya-etal-2025-contextualized}. AmbigQA
\citep{min-etal-2020-ambigqa} tests whether models recover the
interpretations of ambiguous questions, CLAMBER
\citep{zhang-etal-2024-clamber} whether they identify ambiguous queries
and ask clarifying questions, and AmbiBench \citep{tamkin-etal-2023-task}
whether they infer the intended task from ambiguous instructions and
examples. We do not use these datasets, or AmbiEnt and DUST, because their
items are general-domain sentences, questions or instructions whose
interpretations were enumerated or categorized in advance, by annotators or
by construction. The ambiguity we study comes from domain details
that a scientific claim omits, and no dataset we know of lists the
interpretations of such a claim. Measures built on a list of interpretations, for example
whether a model recognizes the ambiguity or recovers each interpretation,
therefore do not apply. A card instead records the interpretation the
agent adopts, which shapes the evidence it seeks, and our measures relate
that interpretation to a verdict an expert has labeled. Disagreement among annotators of
inference items persists with more ratings and context
\citep{pavlick-kwiatkowski-2019-inherent} and partly traces to
underspecified meaning \citep{jiang-marneffe-2022-investigating}. By
analogy, the run-to-run variation in interpretation that we report may
reflect the claims themselves and not only sampling noise.

\paragraph{Ambiguous claims in fact-checking:} In AmbiFC
\citep{glockner-etal-2024-ambifc}, annotators' disagreement about the
veracity of its roughly 10K claims
reflects ambiguity, which the authors trace partly to underspecification, and they train veracity
models on soft labels. Its claims are general-domain claims annotated
against evidence passages supplied with them, and its labels record how
annotators split rather than which interpretation a system adopted. Our
study needs claims that an expert labeled for feasibility and on which the
agent finds its own evidence, and AmbiFC's claims offer neither. Whether the
fields a card marks as default predict where AmbiFC's annotators disagree
is an open question. Dis2Dis
\citep{staliunaite-vlachos-2025-dis2dis} rewrites an ambiguous claim so
that the given evidence supports one interpretation unequivocally. A Dis2Dis rewrite could serve as the claim that \system{} assesses, though
we have not tested this. LIAR-New \citep{pelrine-etal-2023-towards} labels whether
a political statement has enough context to be judged, and
\citet{orlovskiy-etal-2024-uncertainty} ask the user for the information
an ambiguous statement leaves out. Benchmark claims have no author to
ask, so a card records the interpretation instead, and the fields it
marks as inferred or default show what such a question would ask about.
Claimify \citep{metropolitansky-larson-2025-towards} extracts a claim only when it
is confident of the intended interpretation and otherwise flags the
sentence as one it cannot disambiguate, and decontextualization
\citep{choi-etal-2021-decontextualization} recovers missing details from a
source document. Neither applies here, because a feasibility assessment
must return a verdict on the claim as given and our claims come without a
source document. News reports and LLM summaries of scientific findings
often state them more generally than the original papers do
\citep{wuehrl-etal-2024-understanding, peters-chinyee-2025-generalization},
a broadening that a card's quantifier field could record.
\citet{joseph-etal-2026-decide}, drawing on how clinical experts check
health claims, identify underspecified claims and mismatched intentions as
obstacles to end-to-end fact-checking, and they argue that fact-checking
should communicate with the user rather than end in a verdict.

\paragraph{How agents handle underspecified inputs:} On underspecified
software tasks \citep{vijayvargiya-etal-2026-ambigswe}, tool calls
\citep{wang-etal-2025-learning} and user requests
\citep{qian-etal-2024-tell}, agents usually fill the gaps with assumptions
rather than ask for the missing details. We do not use these benchmarks,
because none of their tasks is a claim to assess. Several methods decide
when and what to ask \citep{rao-daume-iii-2018-learning,
zhang-choi-2025-clarify}. Other
methods answer under each interpretation and then combine or list the
answers \citep{hou-etal-2024-decomposing,
saparina-lapata-2026-reasoning}, or use agreement among sampled answers to
decide when to abstain on an ambiguous question
\citep{cole-etal-2023-selectively}. \citet{wen-etal-2025-know} survey
abstention, which in our setting means scoring 0. A card run without
computational libraries took 3.8 minutes and cost \$0.27 on average, counted in GPT-5.4
tokens only and at list price (Appendix~\ref{app:faithful-more}), and
running the agent once per interpretation would multiply that by the
number of interpretations, which grows quickly across nine fields.
Running an edited card next to the card it was edited from is a targeted
version of such an ensemble. Our five runs per claim resemble that
sampling, but we use them to measure how the interpretation varies, not to
decide when to abstain. Self-consistency \citep{wang-etal-2023-self}
samples several reasoning paths and keeps the most common answer, and a
vote over a claim's runs would likewise hide which interpretation each
run used. Our consistency judge, which groups values that
express the same interpretation, resembles clustering sampled outputs by
meaning \citep{kuhn-etal-2023-semantic}.
Analysts given the same data and hypothesis reach divergent conclusions
through different analytic choices, whether they are people
\citep{silberzahn-etal-2018-many, schweinsberg-etal-2021-same} or LLM
agents \citep{bertran-etal-2026-many}, and a multiverse analysis reports a
result across the defensible choices instead of one
\citep{steegen-etal-2016-increasing}. Reporting a verdict with the
interpretation it rests on applies the same idea to one choice at a
time. BLADE \citep{gu-etal-2024-blade} grades an
agent's analytic choices against expert analyses, and DiscoveryBench
\citep{majumder-etal-2025-discoverybench} builds each goal so that it
leads to one answer, removing the ambiguity by construction. These
benchmarks pose questions about datasets, and our claims come without
data. They also assess an agent's choices after the analysis, and a card
records the interpretation at the start of a run.

\paragraph{Science agents with tools:} LLM agents now plan and run
chemistry experiments \citep{bran-etal-2024-augmenting,
boiko-etal-2023-autonomous}, synthesize literature
\citep{skarlinski-etal-2024-language}, retrieve materials data and run
atomistic simulations \citep{chiang-etal-2025-llamp}, answer
materials-science questions from a curated knowledge base and general
tools \citep{zhang-etal-2024-honeycomb}, and test hypotheses by sequential
falsification \citep{huang-etal-2025-automated}. In our runs with the
computational tools, the agent can call comparable tools, in which a coding agent
computes phase equilibria or runs a machine-learned interatomic potential
(Section~\ref{sec:tools-setup}). There we ask how the interpretation
adopted at the start of a run affects whether the agent calculates and what it
calculates (Section~\ref{sec:tooluse}). Across many runs of science
agents in several domains, the base model explained far more of the
variation than the agent scaffold, and failures in how agents used
evidence were invisible to evaluation of outcomes alone
\citep{rios-garcia-etal-2026-ai}. Tool-calling agents are inconsistent
across repeated trials \citep{yao-etal-2025-tau}, and agent evaluations
seldom account for cost \citep{kapoor-etal-2025-ai}. Comparing \system{}
with these agents would confound agent design with interpretation, since
this study keeps one agent and varies how its interpretation is recorded
or set. A card could be added to any of them.

\paragraph{Making an interpretation visible:} A model's reasoning need not
reflect what drives its answer \citep{turpin-etal-2023-language,
chen-etal-2025-reasoning}, and an explanation can be plausible without
being faithful \citep{jacovi-goldberg-2020-towards}. A card is extracted
from the agent's first reasoning step by another model, so the same
caution applies to it. In card mode the card reaches a run only through
the scorer and the report step, whereas in frozen mode it is a structured
instruction in the agent's prompt. Changing a field can change the score,
and the edited value often reached the questions the agent sent to its
tools, the context it passed to the search agent and its final answer
(Section~\ref{sec:control}). Such an edit is like editing a concept of a
concept bottleneck model at test time \citep{koh-etal-2020-concept}. Interfaces that show a model's interpretation
of a request, before it acts or as an explanation of its output, and let
users edit it have helped users steer or correct systems for text-to-SQL
\citep{tian-etal-2023-interactive}, end-user programming
\citep{liu-etal-2023-wants}, data analysis
\citep{kazemitabaar-etal-2024-improving}, code generation
\citep{zhang-etal-2025-neurosync} and agent planning
\citep{feng-etal-2026-cocoa}. A Claim Card is shown before
the agent's first tool call runs (Section~\ref{sec:visibility}).
Explanations in automated fact-checking justify a verdict after the
evidence is gathered \citep{atanasova-etal-2020-generating}, while the
fact-checkers interviewed by \citet{warren-etal-2025-show} asked for
explanations that show uncertainty and information gaps. Explanations can
also raise reliance on a system whether or not it is right
\citep{bansal-etal-2021-whole}, a risk we return to in the Ethics
Statement.

\paragraph{Claim structure and documentation:} Structured representations record what a claim states. PICO frames
clinical evidence by population, intervention, comparator and outcome, the
elements that EBM-NLP \citep{nye-etal-2018-corpus} annotates in trial
abstracts, and SciClaim
\citep{magnusson-friedman-2021-extracting} annotates the factors,
associations, magnitudes and qualifiers of scientific claims. Neither
records the threshold a claim must meet or where each value came from,
and neither carries feasibility labels, so their annotations could not
serve as cards or as gold labels here. FActScore
\citep{min-etal-2023-factscore} decomposes generated text into atomic
facts and scores each, whereas a card decomposes a claim
into the fields a feasibility assessment needs, including those the claim
leaves out. Model cards \citep{mitchell-etal-2019-model}, datasheets
\citep{gebru-etal-2021-datasheets} and data statements
\citep{bender-friedman-2018-data} record the assumptions behind a model or
a dataset for others to inspect, and preregistration
\citep{nosek-etal-2018-preregistration} commits to hypotheses and analysis
plans before data are seen. A Claim Card records the assumptions behind a
single verdict. Like a preregistration, it exists before any evidence is
gathered, and it is saved with the report.

\section{Further Discussion}
\label{app:discussion-more}

Some disagreements between a run and a gold label may be disagreements
about what the claim means (Section~\ref{sec:conclusion}), and the
experts' own calculations show one way this can arise. The experts
assessed two claims through proxies computed with CHGNet. For
the Fe-doped Co$_3$O$_4$ claim, the expert used the total magnetic moment
as a proxy for saturation magnetization, whereas the frozen card named
saturation magnetization itself. The expert's proxy for stiffness on the
Mg-in-Al stiffening claim was the bulk modulus, and the frozen card named
Young's modulus. On these
claims a run and the expert may have assessed different observables, so a
run that disagrees with the gold label has not necessarily reached a wrong
verdict.

Runs with the computational tools that obtained the same calculation still
had to decide whether the result met the claim. On the Li--AlCu reaction
claim, six card and no-card runs and all five frozen runs found the same
three-phase field and reached opposite verdicts, depending on whether a two-phase
reaction meant two coexisting phases or AlCu coexisting with a
lithium-containing phase (Appendices~\ref{app:samecalc}
and~\ref{app:gold}). Since the two groups of runs come from different
conditions, the contrast may also reflect their other differences. Part of
the decision can be made by the coding agent, because each playbook asks
it for its own score under tolerances that no card field sets
(Appendix~\ref{app:tooluse-more}). We think the computational tools move
part of the ambiguity from what evidence the agent can find to how it
judges a computed value against the claim.

Some of what decided the verdicts is not on the card. On the Fe--Al
Al$_2$Fe-onset claim, all five frozen runs computed the same B2 limit
under one card whose quantifier was typical, and their scores still ranged
from $+2$ to $-1$. Two of them had committed in their first reasoning step
to scoring it negative on a single matched counterexample. The card holds the
quantifier from which that rule follows, but not which result counts as a
counterexample. In one card run on the Al--Mg--Zn $\tau$-phase claim, the
report and the card disagreed on the quantifier
(Appendix~\ref{app:faithful-more}).
A decision-rule field could state, before any evidence is gathered, what
result would refute the claim, and a tolerance field could give the coding
agent the card's threshold in place of the playbook's defaults. We would
test these two fields. Other choices that can decide a computational
verdict could also become fields, among them an equilibrium or a kinetic
(Scheil) treatment of solidification, the composition basis (atomic or
weight percent), the basis of a phase fraction (mole or volume), the
thermodynamic database and the numerical tolerances.

Where a claim has an author to ask, the
fields a card marks as inferred or default could become clarification
questions put to that author before the run.

\section{Prompts}
\label{app:prompts}

This appendix gives the instructions that define the conditions, the
edits and the judges. Words in braces are filled in for each claim, and
listings are line-wrapped. The guidelines, playbooks and scoring rubrics
are too long to print and will be released with the code, together with
these prompts.

\paragraph{The card section of the guideline:} Card runs receive the
guideline's card section, given here in its general-science version. In
the materials-science version, the first item names a route or
architecture (with the interfaces the route implies) and an environment
(temperature, pressure, atmosphere, humidity, radiation, field) in place of
the method or protocol and the conditions, and the sufficiency set covers
system, route or deployment claims, with material viability, integration
budgets closed with margin, controls and operations, and manufacturability
as its elements. The rest differs only in typography. The section's ``lock before tools'' means that the agent records the
card before it uses a tool. It does not refer to the frozen condition.
The guideline tells the agent elsewhere to ``prefer the weakest faithful
interpretation of a claim.''

\begin{lstlisting}
### 1.1 Claim Card (lock before tools)

Record, verbatim where possible:

- **System/entity**; **state/phase**;
  **method/protocol** (any procedures implied);
  **conditions** (temperature, pressure,
  concentrations, durations, environmental
  variables).
- **Observable/FoM + units** (precise metric the
  claim is about).
- **Quantifier**: existential ("can/there
  exists"), typical-case universal
  ("generally/typically"), strict universal
  ("for all/always").
- **theta: threshold/significance rule**
  (numeric or standard). If none is given,
  choose the weakest defensible threshold (e.g.,
  "> method uncertainty," ">= detection limit").
- **Comparator logic** (vs what? AND/OR
  semantics; disjunctive parts interpreted as
  **OR** unless stated otherwise).
- **Sufficiency set** for
  *system/method/deployment* claims: the minimal
  elements that must simultaneously hold (method
  validity, boundary conditions closed with
  margin, controls, reproducibility).
- **Assumed defaults** when the text is silent
  (see 1.6), and **near-match tolerances** you
  will accept later (see 4.2).
- **What would decisively change the score**
  (pre-register the minimal decisive evidence).
- **Provenance**: for each entry above, mark
  whether it is *stated* in the claim text,
  *inferred* from the text with domain
  knowledge, or a *default* you assumed because
  the text is silent.
\end{lstlisting}

\paragraph{The frozen-mode instruction:} In frozen mode, \system{} appends
a short section to the agent's query, headed ``Locked Claim Card.'' Its
text reads ``The Claim Card for this claim has already been completed and
locked. Use it exactly as given. Do not write a new Claim Card, and do not
change the interpretation of the claim, its quantifier, its threshold, or
its assumed defaults.'' The card follows as JSON. The instruction's
``locked'' card is what this paper calls the frozen card.

\paragraph{The scorer:} The scorer reasons step by step before it
answers. Its inputs and outputs are listed below, and the description of
\field{claim} is replaced by one for the claim's domain. In no-card runs
the \field{claim\_card} input is removed, and so are the card items of
the scoring guideline, whose other pre-flight checks stay.

\begin{lstlisting}
LikertEstimator
  claim: <domain-specific description>
  claim_card: the claim card for the claim that
    describes how the claim was interpreted.
    Represented either as a stringified JSON object
    or as a string.
  agent_reasoning: agent's reasoning during the
    claim assessment
  claim_assessment_report: the report from the claim
    assessment
  -> likert_score: likert score in [-2, 2], with -2
    being extremely infeasible, -1 being partially
    infeasible, 0 being undecidable, 1 being
    partially feasible, and 2 being extremely
    feasible.
  -> confidence: value between 0 and 1.

Instruction: Given the fields `claim`,
`claim_card`, `agent_reasoning`,
`claim_assessment_report`, produce the fields
`likert_score` and `confidence` by following the
likert scoring guidelines:
<the domain's scoring guideline>
\end{lstlisting}

\paragraph{The threshold edit:} Claude Opus 5.5, with Gemini 3.1 Pro as
a fallback for the 60 claims run without computational libraries, received the following prompt for each
frozen card, with the card's threshold, observable and units.

\begin{lstlisting}
You are editing one field of a structured
interpretation of a scientific claim.

Claim: {claim}

Current threshold for deeming the claim
supported: {threshold}
Observable: {observable}
Units: {units}

Rewrite the threshold so that it is strictly
more demanding than the current one, and still a
threshold a domain expert could plausibly ask
for. If the current threshold is qualitative
(for example "any measurable increase"), replace
it with a concrete quantitative one. If it is
quantitative, make the required effect larger.
Keep the same observable and units.

Return only the new threshold text, in one
sentence, with no preamble.
\end{lstlisting}

\paragraph{The field audit:} The audit judge received the first prompt
for each first-run card, with each field shown together with its
provenance mark, and the second for each materials-science card whose
claim has an expert rationale.

\begin{lstlisting}
A system wrote a structured interpretation (a
"Claim Card") of a scientific claim before
assessing it. Audit the card against the claim.

Claim: {claim}

Claim Card:
{card}

For each field, choose one label:
- faithful: the value restates what the claim
  says, explicitly or by direct implication.
- reasonable_default: the claim is silent or
  vague on this point, and the value is a
  default that a domain expert would accept as
  the natural reading.
- unsupported: the value adds a constraint,
  number, or detail that the claim does not
  support and that an expert would not assume by
  default.
- contradicts: the value conflicts with what the
  claim says.

Return a JSON object with one key per field
({fields}). Each value is an object with the
keys "label" and "reason" (one sentence).
\end{lstlisting}

\begin{lstlisting}
A system wrote a structured interpretation (a
"Claim Card") of a scientific claim.
Independently, a domain expert judged the claim
and wrote a rationale.

Claim: {claim}

Claim Card:
{card}

Expert rationale:
{rationale}

Does the expert rationale rely on a reading of
the claim that the Claim Card excludes? Answer
"compatible" if the expert's reading fits within
the card, "excluded" if the expert reads the
claim in a way the card rules out (for example a
different system, scope, quantifier, threshold,
or comparator), or "unclear" if the rationale
does not show how the expert read the claim.

Return a JSON object with the keys "verdict" and
"reason" (one sentence), and "field" (the card
field at issue, or null).
\end{lstlisting}

\paragraph{The consistency judge:} The consistency judge received the
five runs' values of every field of a claim at once.

\begin{lstlisting}
Below are {k} interpretations of the same
scientific claim, produced by {k} independent
runs of a system. Each interpretation has the
same fields.

Claim: {claim}

{interpretations}

For each field, decide which runs express the
same interpretation. Two values are the same
interpretation if an expert testing the claim
would do the same thing under both; differences
in wording, order, or level of detail that would
not change the test do not count. Two values
differ if they would lead to a different test, a
different pass criterion, or a different scope.

Return a JSON object. For each field, give a
list of {k} group labels (integers starting at
1), one per run in the order shown, where runs
with the same label express the same
interpretation. Use exactly these keys:
{fields}.
\end{lstlisting}

\end{document}